\documentclass[Afour,sageh,times]{sagej}

\usepackage{moreverb,url}

\usepackage[colorlinks,bookmarksopen,bookmarksnumbered,citecolor=red,urlcolor=red]{hyperref}
\usepackage{soul}
\usepackage{graphicx}
\usepackage{tabularx}
\usepackage[textsize=small]{todonotes}
\usepackage{algorithm}
\usepackage{algorithmic}
\usepackage{amsmath,amssymb,amsfonts}
\usepackage{mathtools}
\usepackage{relsize}
\usepackage{subfig} % putting multiple figures into one figure
\usepackage{fancyhdr}
\usepackage{balance}
\usepackage{enumitem}

\newcolumntype{C}[1]{>{\centering}m{#1}}
\newcolumntype{L}[1]{>{\raggedright}p{#1}}

\newtheorem{mydef}{Definition}

\newtheorem{hypo}{Hypothesis}

\newcommand\BibTeX{{\rmfamily B\kern-.05em \textsc{i\kern-.025em b}\kern-.08em
T\kern-.1667em\lower.7ex\hbox{E}\kern-.125emX}}

\begin{document}

\title{FABRIC: A Framework for the Design and Evaluation of Collaborative Robots with Extended Human Adaptation}
%\shorttitle{A Framework for Cobots with Extended Human Adaptation}
% TODO: \title{A Framework for Collaborative Robots and A Pipeline to Train, Deploy and Evaluate Its Extended Human Adaptation}

\author{O.~Can~G\"{o}r\"{u}r\affilnum{1}, Benjamin~Rosman\affilnum{2}, Fikret Sivrikaya\affilnum{3}, and Sahin Albayrak\affilnum{1}} 

\affiliation{\affilnum{1}DAI-Labor, Technische Universit\"{a}t Berlin, Germany\\
\affilnum{2}University of the Witwatersrand, South Africa\\
\affilnum{3}GT-ARC gemeinn\"utzige GmbH, Germany
}

\corrauth{O.~Can~G\"{o}r\"{u}r, DAI-Labor,
Technische Universit\"{a}t Berlin
Ernst-Reuter-Platz 7,
10587, Berlin,
Germany.
}

\pagestyle{fancy}
\lhead{G\"{o}r\"{u}r et al.}
%\runninghead{G\"{o}r\"{u}r et al.}
\rhead{Cobots with Extended Human Adaptation}

\email{goeruer@tu-berlin.de}

\begin{abstract}
A limitation for collaborative robots (cobots) is their lack of ability to adapt to human partners, who typically exhibit an immense diversity of behaviors. We present an autonomous framework as a cobot’s real-time decision-making mechanism to anticipate a variety of human characteristics and behaviors, including human errors, toward a personalized collaboration. 
Our framework handles such behaviors in two levels: 1) short-term human behaviors are adapted through our novel Anticipatory Partially Observable Markov Decision Process (A-POMDP) models, covering a human's changing intent (motivation), availability, and capability; 2) long-term changing human characteristics are adapted by our novel Adaptive Bayesian Policy Selection (ABPS) mechanism that selects a short-term decision model, e.g., an A-POMDP, according to an estimate of a human's workplace characteristics, such as her expertise and collaboration preferences. To design and evaluate our framework over a diversity of human behaviors, we propose a pipeline where we first train and rigorously test the framework in simulation over novel human models. Then, we deploy and evaluate it on our novel physical experiment setup that induces cognitive load on humans to observe their dynamic behaviors, including their mistakes, and their changing characteristics such as their expertise. We conduct user studies and show that our framework effectively collaborates non-stop for hours and adapts to various changing human behaviors and characteristics in real-time. That increases the efficiency and naturalness of the collaboration with a higher perceived collaboration, positive teammate traits, and human trust. We believe that such an extended human adaptation is a key to the long-term use of cobots.

\end{abstract}


\maketitle

\section{Introduction}
\label{sec:intro}

%In the fourth industrial revolution (I4.0), mass customization takes an important role for the future of production. A greater individualization is expected for the consumer products leading to higher variability in produced parts \cite{oecd_next_2017}. This requires more flexibility and automation in production lines for an efficient operation \cite{satoglu_lean_2018}. For this purpose, I4.0 envisions the spatial and temporal overlapping of human and robot workstations in the production process \cite{koppenborg_effects_2017}. Future factories should efficiently cooperate the productivity, accuracy, strength and repeatability of robots with the flexibility, creativity and cognitive capabilities of humans in mass customization \cite{pfeiffer_robots_2016, rojko_industry_2017, villani_survey_2018}. A resulting challenge is the need for robots to handle a wider range of production processes and more dynamic human collaborations \cite{mavridis_nikolaos_review_2015, rojko_industry_2017}.
%Research into human-robot collaboration (HRC) has been focusing lately on altering a robot's plans dynamically, often without verbal communication, to maintain reliable and efficient collaboration with a human \cite{Hoffman2004}.
%TODO: remove the I4.0
An efficient human-robot collaboration (HRC) on production lines involves robots monitoring a human partner's actions and processing them to anticipate the human's plans and goals in a collaboration task \citep{koppenborg_effects_2017, KukaI40}. This anticipation requires nonverbal communication to increase the efficiency in industrial environments, e.g., due to hearing limitations \citep{Calisgan2012, Hoffman2007, Lasota2017}.
The use of such anticipated knowledge in a robot's decision-making mechanism is dependent on the adaptation of these collaborative robots (cobots) to various types of humans, their dynamic behaviors, needs, and preferences \citep{Huang2016, Hiatt2011, Nikolaidis2015}. Our motivation is to ensure such autonomous anticipatory adaptation of robots and their fluent coordination, i.e., well-synchronized coordination of robot behaviors with their human partners through mutual adaptation \citep{hoffman_evaluating_2019}. We refer to such robots as \textit{social cobots}.
%We are capable of altering our plans and actions dynamically, showing a great level of adaptation to the environment and to the collaborated partners
%%The majority of HRC is following command and response patterns, and not giving importance to fluent coordination, which is stated to evoke appreciation and confidence.
% TODO: RESEARCH QUESTION: ``How can a robot anticipate and adapt to (i.e., model) unanticipated human behaviors, including a human's short-term changing intent (motivation), availability, capability and willingness to collaborate, which could lead to erroneous human behaviors?''
%TODO: RESEARCH QUESTION: ``How can a robot anticipate and adapt to a diversity of long-term human characteristics, trust, personal habits and preferences in its decision-making, which may eventually affect a human's short-term behaviors?''

From the existing studies on interactive robots, we deduce that to ensure the long-term usability of collaborative robots, a robot should adapt to both short-term changes for individual differences (e.g., a tough day at work may cause a dynamic human performance) and long-term personal habits, preferences and trust \citep{Tapus2007,Leite2013,Irfan2019}.
Hence, we propose a framework that can handle both short-term and long-term robot adaptation.
For fluent coordination, we require that short-term adaptation should handle the following settings: \textit{i)} when the tasks are short or come with changing requirements, e.g., flexible task allocation for mass customization; \textit{ii)} when the collaboration partner exhibits dynamic behaviors due to, e.g., their changing mental states like intentions and emotions that may cause dynamic human performance and preferences \citep{Valerio2017, hoffman_evaluating_2019, Andriella2019}.
We refer to the collection of such short-term changing human behaviors that may result in a human error, a task failure, or dropped efficiency in a collaboration task as ``unanticipated'' human behaviors, as they are unreasonably uncommon to observe compared to expected human performance. These are largely overlooked by the existing studies. For example, a human may lose attention or may not want the robot's assistance due to, e.g. mistrust.
To address this, we propose a partially observable Markov decision process (POMDP) that models such unanticipated human behaviors as a latent variable. With that, it anticipates and effectively handles the conditions when a human intention is irrelevant to an assigned task (possibly unknown to the robot) and when a human does not want robot assistance. The model, then, further reasons on such human states to estimate if the human really needs assistance and whether the robot should intervene. Hence, we do not enforce a turn-taking collaboration but rather allow for flexible collaboration. A robot deploying this POMDP optimizes the plan for improved efficiency and naturalness of the collaboration, ensuring the safety and the autonomy of the human partner. We name our novel approach Anticipatory Partially-Observable Markov Decision Process (A-POMDP).
%It is stated that although it is crucial, robots have not yet reached a level of design that allows effective management of such human errors or their unexpected behaviors \autocite{Honig2018}. The premise of this thesis regarding a robot's short-term adaptation during a human collaboration is, for the first time to the best of our knowledge, to devise a robot decision-making mechanism that aims to remove these assumptions and model the \gls{HRC} over the course of repeated interactions to be able to reason about such ``unanticipated cases'' that could affect the performance of the collaboration. For this purpose, we ask the following research question to be able to extend the short-term adaptation of a \gls{cobot}:

%Long-term adaptation, on the other hand, focuses on behavioral changes on the robot over long periods of time based on the characteristics of the environment; therefore, it requires a memory of previous interactions (e.g., in repeated interactions) \cite{Andriella2019}.
Most HRC studies focus on short-term collaboration due to practical reasons, leaving long-term HRC that sustains a rapport and efficient collaboration as a largely unexplored area. Moving from the relevant literature in \citep{Leite2013, Leite2014, Valerio2017, Andriella2019, Irfan2019}, we consider the following target settings for long-term adaptation: \textit{i)} when collaboration is required for a repetitive task; \textit{ii)} when long-term collaborations with different people are expected (e.g., HRC in a factory environment with many work shifts). Our interpretation of long-term adaptation is that a robot needs to distinguish and adapt to various characteristics of the collaborated humans, i.e., personalization, that eventually affect their short-term behaviors, e.g., a human with a lower expertise is more likely to need a robot's assistance in a task. We call each different combination of such characteristics a unique \textit{human type}. 
We believe that it is already very difficult to design or learn a single intention-aware model for a person that a robot is collaborating with, let alone for different human types, due to the growing complexity of such models. Our intuition is that rather than a single adaptive model, sometimes a robot may need to follow completely different decision-making strategies, i.e., policies, to enable fast and reliable online adaptation to various human types.
%INFO: removed our previous study, but proposed it as if it is new.
For this purpose, we present a novel Bayesian policy selection mechanism built on top of existing intention- and situation-aware models (i.e., our A-POMDP models) for an extended adaptation of robots to various and possibly unknown human types.
The selection is based on an estimate of a human's long-term workplace characteristics that correlate to the policy performance; hence, we name this mechanism Adaptive Bayesian Policy Selection (ABPS).
% Instead of modeling known human types as a latent variable, we estimate human types that are unknown to the robot from the observed human behaviors using Bayesian belief estimation.
%The selection is based on the observed human behaviors that correlate to the policy performance. The observations are used for Bayesian belief estimation of an unknown human type, instead of modeling types as a latent variable. The classification is based on a human's long-term workplace characteristics and preferences such as level of expertise, stamina (or fatigue), attention and collaborativeness\footnote{We use this term to indicate the level of a human's will to collaborate that may change due to, for example, task-relevant distrust of the human to the robot.}.

In this paper, our first goal is to extend and further improve our previous works on short- and long-term adaptation models and mechanisms, i.e., Görür et al. \citeyearpar{Gorur2018} and \citeyearpar{Gorur2019}, respectively, and the second goal is to deploy and evaluate these approaches through user studies on real world scenarios.
Our first contribution is a novel framework for collaborative robots that integrates our strategies for short- and long-term extended human adaptation, i.e., A-POMDP models and ABPS mechanism, respectively, into a holistic approach.
This is an anticipatory architecture that ensures a robot's fully autonomous human adaptation by integrating the short-term adaptation goals and deploying long-term strategies to regulate and personalize short-term robot behaviors accordingly. That is, a decisional level categorizes a human's characteristics as her level of \textit{expertise, stamina (or fatigue), attention} and \textit{collaborativeness}\footnote{We use this term to indicate the willingness of a human to collaborate that may change due to, for example, task-relevant distrust of the human to the robot.}, and selects the best short-term strategy, e.g. a A-POMDP model, toward a personalized collaboration. Then, the selected strategy is executed to recognize the human's changing \textit{availability, intent (motivation)}, and \textit{capability} as observations, and to estimate further if the human wants/needs help to guide robot behaviors accordingly. The real-time execution of the selected strategy and sensing and actuating skills of a robot take place in the functional level of the architecture.

%TODO: try to give that the pipeline was needed in our case as the field needs a systematic way to go for beyond existing adaptation goals/problems.
Our second goal is to train and validate this framework; however, benchmarking interactive robots, in general, is a difficult task due to the lack of availability of the whole range of human behavior dynamics during their training and evaluation process \citep{Webster2020}. As a result, there exists almost no HRC research that considers a vast range of human intentions and actions to the best of our knowledge.
Subsequently, such robots show very limited adaptation capabilities as they face a greater diversity of previously unanticipated human behaviors when they are deployed in the wild \citep{Tulli2019, Hoffman2019_web}. As our second contribution, we address this issue by devising a pipeline for systematically benchmarking collaborative robots. The pipeline offers a way of training any such system toward various adaptation goals and to test them under these and many more diverse conditions.
Similar to the development cycles of autonomous systems, our pipeline rigorously tests the framework first in a simulation with accurate human models, then deploys and validates it in a real environment through user studies. Since the pipeline follows a continuous development and integration practice, both of these processes are repeated and continuously provide feedback while the collaborative robot encounters a vast range of short- and long-term dynamics of human behaviors.
%This ensures our framework's effective and extended human adaptation.

%TODO: see if I can fit Hypotheses below !!!!
%We follow the proposed pipeline in training and validating our framework over our collaboration environments that emulate real world conditions.
We follow the pipeline to evaluate our framework's extended adaptation goals. In order to exemplify our framework and the pipeline, we focus on a nonverbal HRC scenario at a conveyor belt for the task of inspecting and storing various products.
We devise a novel collaboration setup, that provides a rather unconstrained human intention space by running a cognitively exhaustive task and by not enforcing a turn-taking collaboration, unlike most HRC experiments \citep{hoffman_evaluating_2019, Chao2012}.
The task allows the robot to observe various human characteristics, e.g., a competitive person with bad skills or a defeatist person, and ``unanticipated'' human behaviors, e.g., a human constantly rejects the robot's assistance, and tires easily. Moreover, we develop the same scenario in a factory simulation with our novel human decision models as Markov Decision Processes (MDPs). The models provide reliable responses with a greater diversity through sampling than the ones observed in a real setup. This allows the framework to train on large scale data and to be rigorously tested under greater uncertainty. To the best of our knowledge, this is the first time an anticipatory robot decision-making mechanism has been tested on such a greater diversity of human behaviors and characteristics for a more realistic evaluation of its adaptation skills.
%without constraining human intentions, which eventually allows a cobot to observe ``unanticipated" human behaviors, different human characteristics and their various combinations in a lab environment.

%TODO: this para. will be updated based on the structure of the paper !!!
In the remainder of the paper, we first give details of our framework in Section~\ref{sec:framework} and how it incorporates our A-POMDP models for short-term adaptation (in Section~\ref{ssec:apomdp}) and ABPS mechanism for long-term adaptation (in Section~\ref{ssec:abps}). Then, we describe our continuous development and integration pipeline for human-aware systems and show how we use it to evaluate our framework in Section~\ref{sec:pipeline}. This is followed by our novel experiment setup and the collaboration scenario, detailed along with our system's real-time human interaction, in Section~\ref{sec:deployment}.
Finally in Section~\ref{sec:evaluation}, we provide our results by first demonstrating that our collaboration experiments are able to create a cognitive load on humans and that we observe ``unanticipated'' human behaviors. Then, we validate that our A-POMDP model design provides a more efficient and natural collaboration compared to anticipatory models that do not cover ``unanticipated'' human behaviors. In the final experiments, we show that the complete framework, handling such human variability, provides a fast and reliable adaptation to both short- and long-term changing human behaviors, while being perceived to have high collaboration skills, positive teammate traits and trust.
To the best of our knowledge, this is the first time a systematic approach has been applied to developing and testing social collaborative robots, that takes into account such a large diversity of human behaviors and characteristics.
%TODO: Most important below. If the contribution is the framework, then we test it in parts, and show its applicability at the end (not comparing it with other frameworks).

\section{Related Work}

Existing intention-aware planning approaches mostly introduce human intentions as a latent variable in a decision-making model, such as in POMDPs \citep{Broz2013,Nikolaidis2017,Chen2018}. Such a modeling scheme has proven to be efficient in HRC scenarios; however, this scales poorly with an increasing number of human intentions modeled. Hence, such a design conventionally has to limit the human intention space and systemic errors a human can make \citep{Hiatt2017}. Therefore, the studies implicitly make the assumption that either a human's intention (or goal) is constant or it is changing in a limited intention space known to the robot \citep{ Baker2014, Holtzen2016, Devin2016, Milliez2016, Huang2016,Penkov2016}. 
Moreover, they further assume that humans accept the robot's assistance when offered. 
It has been stated that such assumptions limit a robot's anticipation of a human's changing behaviors and goals that are mostly observed over the course of a repeated collaboration \citep{Leite2013,Albrecht2018}.
 
Contingencies in human actions have been partly considered \citep{Koppula2016,Hiatt2011}; however, all observed actions are still assumed to be toward fulfilling a task, possibly in a way that differs from the expected plan. In a repeated HRC over some tedious tasks, it is more likely that the human performs behaviors that are not even related to the task itself but implicitly affects her performance, e.g., due to fatigue \citep{Ji2006}. The robots should be aware of and adapt to such unanticipated behaviors of humans, which to the best of our knowledge, represents a largely unexplored area of research. For this purpose, we propose our A-POMDP mechanism as an alternative to the existing high-level human intention-aware decision models. It removes the assumptions mentioned here, and additionally incorporates such unanticipated human behaviors, e.g., getting tired, not wanting the robot's assistance. With such an extended short-term adaptation, it is also able to coordinate with humans on changing plans and roles for the collaboration task.

Long-term adaptation of robots is still largely unexplored in the HRI domain due to practical reasons \citep{Irfan2019}. In long-term, humans exert even more variety of behaviors that require further strategies for the robots to adapt. In \citet{Nikolaidis2015}, humans are clustered from observations during a training phase into a finite number of human types. The estimated human type is again used as a latent variable in a mixed observability MDP (MOMDP) model; hence, the authors limit the number of different types due to the increasing complexity with the models.
It has been recently stated that POMDPs require accurate system models that are often unavailable or fail to adapt to various conditions in long-term missions \citep{Mcguire2018}. Bandyopadhyay et al. \citeyearpar{Bandyo2013} build several such MDPs with varying reward and transition functions to handle different tasks. In other words, the robots are given the ability to explore different policies and trade-off between interaction and task quality. However, the study is limited to analyzing different policies to govern such varieties in humans, leaving out the autonomous selection of an optimal one. 
Our approach to long-term adaptation brings together the idea of generating many such reliable Markov models in \citet{Bandyo2013} to construct a policy library, and the idea of estimating human types on a meta-level as a complementary solution to the intention-aware models in \citet{Nikolaidis2015}, and goes beyond them to offer a fast and reliable policy selection mechanism as part of a closed-loop robot system.

For policy selection in the context of social agents, the exploration factor would be very dangerous and frustrating for a human collaborator in the real world; for instance, when using a contextual multi-arm bandit (CMAB) in an assistant selection for collaborated humans \citep{Mcguire2018}. In addition, in policy or reward learning algorithms, the learning rate is very difficult to tune and the response time is relatively high for any interaction in real-time \citep{Ramakrishnan2017}.
Particularly, when human workers have their shift changes or when a human drastically exerts different behaviors (e.g. loss of attention), learning a new policy would take time which is very costly in collaboration scenarios. Moreover, we still need to have an accurate reward and transition model, which, in the end, needs to be applicable to all humans being interacted with. In such cases, it is better to reuse an already trained reliable model rather than training a new one.
In our solution to the long-term adaptation, the ABPS mechanism, we have incorporated Bayesian Policy Reuse (BPR) in \citet{Rosman2016}, which has been shown to perform faster and more reliable in online adaptation tasks with greater uncertainty.
Instead of modeling human types as a latent variable, ABPS complements existing intention-aware planning solutions, e.g., our A-POMDP, by selecting a policy from a library of such models, each of which already handles the short-term adaptation, based on an estimate of a human's long-term workplace characteristics.
This allows us to handle more human diversity in a more computationally efficient way, and to handle unknown human types, while mitigating the need to learn response policies on the fly.
Even though the ABPS is agnostic to any labels of human types and robot policies, we generalize some characteristic features of humans in workplaces, inspired from previous studies \citep{Gombolay2017, Mcguire2018, Ji2006}, that are crucial for a collaborative robot to know. These are a human's \textit{expertise, attention, stamina-level} and \textit{collaborativeness} and they are used to describe a human type.

Regarding the evaluation of adaptation, benchmarking interactive robots is very difficult due to the lack of availability of the complete dynamics of human behaviors. It is hard to expect humans in real user studies to convey diverse behaviors, for example, due to the novelty effect they face or due to the constrained environments \citep{Leite2013}. That said, we believe that there is a lack of human simulations that the anticipatory robots can use as a ground truth to train and test on human diversity. Hence, we devise a simulation environment with crafted human models sampling this diversity. The tested solutions can then be brought to the real user-studies, providing more effective adaptation to humans. However, it is an open task to transfer the results from such a simulation into a real-world experiment and validate it by means of user studies. Our intuition is that an evaluation setup should emulate real conditions instead of constraining human intentions so that we observe more dynamics, including unanticipated human behaviors. Additionally, most of the existing HRC solutions are structured around command and response patterns or turn-taking with previously set roles, which limits the fluency, i.e., a key to a satisfying collaboration \citep{hoffman_evaluating_2019}. We believe that a collaboration setup also needs to enable flexible planning instead of preassigned plans and roles so that the collaborating partners can compensate for possible unexpected situations. For that, following the existing research on human work environments, e.g., \citep{Hiatt2017}, we design an experiment setup that induces a cognitive load on humans to invoke a larger diversity of human intentions and behaviors. Then, our robot is faced with greater uncertainty and should naturally coordinate with the human to compensate for possible human errors and to contribute positively to the task.

%In addition, throughout the thesis work we focus on nonverbal communication due to the following reasons: \textit{i)} It has been argued that the majority of human-human communication is nonverbal \cite{Calisgan2012}; \textit{ii)} A fluent coordination is often reached through nonverbal communication \cite{hoffman_evaluating_2019}; \textit{iii)} In a factory environment verbal communication is often not efficient due to the noise pollution and that the workers usually need to wear hearing protection \cite{Calisgan2012}.

%\section{Anticipatory Robot Decision-Making Framework}
\section{Anticipatory Decision-Making Framework}
\label{sec:framework}
\begin{figure*}
	\centering
	\includegraphics[width=0.80\textwidth]{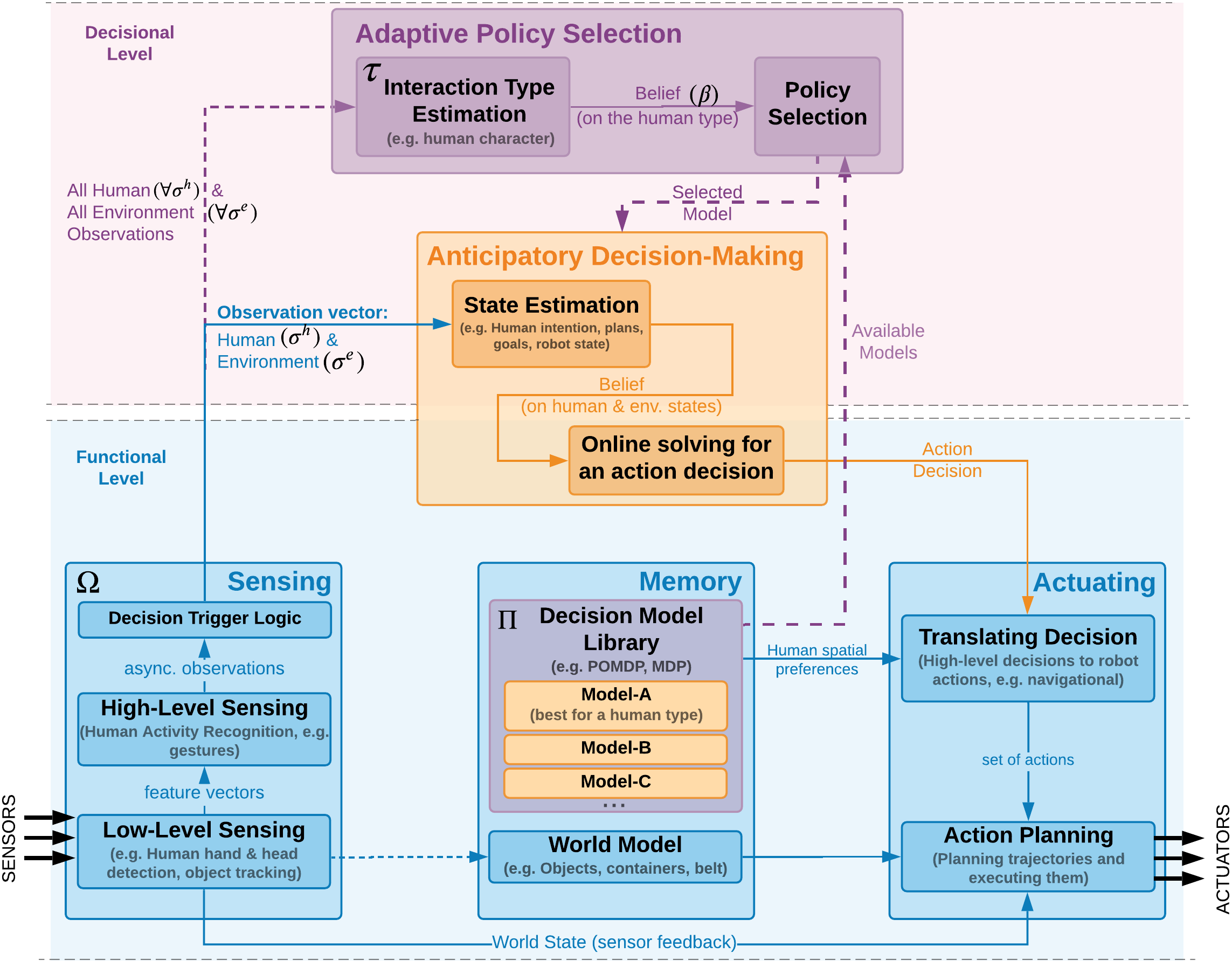}
	\caption{Our Framework for Anticipatory Behaviors in Robots toward Interactive human Collaborations (FABRIC) provides improved short- and long-term adaptation to humans for a robot's personalized and safe interaction with humans.}
	\label{fig:framework}
\end{figure*}

Our framework is designed as a human-aware system to anticipate human behaviors and respond with appropriate robot decisions. It models humans with their long-term characteristics and short-term changing behaviors and utilizes these models in intention-aware high-level decision-making to regulate the low-level functional processes of a robot. We deploy this framework to improve a collaborative robot's long-term assistance during an extended collaboration with humans. We name the architecture FABRIC, as a ``Framework for Anticipatory Behaviors in Robots toward Interactive human Collaborations'' (shown in Figure~\ref{fig:framework}). It is a fully autonomous lightweight system with human-in-the-loop decision-making in real-time. A cobot integrating FABRIC selects and executes the best intention-aware decision model, i.e., a policy, for the interacted human. Our application domain is HRC; however, broader use of FABRIC is possible for personalized robot assistants adapting to cared-for individuals, for example, social robots as home assistants.

A common practice in developing robotic system architectures is to consider autonomy in two levels, i.e., functional and decisional autonomy \citep{Alami1998}. It is also stated that self-regulated learning is crucial in open-ended tasks, which is the case for cobots in their long-term operations, and it requires both the functional process for sensing, acting, and constructing knowledge and the decisional process for monitoring and regulating the learned knowledge, i.e., high-level reasoning \citep{Gorur2016, Sun2006}. Moving from this, we design FABRIC with two main parts: \textit{Decisional level} and \textit{functional level}. The \textit{functional level} is mostly reactive and it integrates sensory-motor skills of a robot that is still able to function without the \textit{decisional level}. It comprises of the \textit{sensing, memory}, and \textit{actuating} components. The framework recognizes external stimuli (mostly human observations) and actuates action decisions for executing plans in the form of robot actions. The \textit{decisional level}, on the other hand, further processes the observations to infer the characteristics of the human (long-term behaviors), current human mental state (short-term behaviors), and the state of the environment, and generates high-level decisions in the form of strategies, goals, and plans for the functional components. It continuously evaluates the success of the current strategy and, if necessary, interrupts to change it; however, in general, it is a non-blocking process for the \textit{functional level} that runs in real-time. Our short- and long-term adaptation goals for the cobot is realized in the \textit{decisional level} through the \textit{anticipatory decision-making} (in Section~\ref{ssec:apomdp}) and the \textit{adaptive policy selection} (in Section~\ref{ssec:abps}) components, where the former acts as a mediator between the two levels as shown in Figure~\ref{fig:framework}.

\subsection{Functional Level}
\label{ssec:functional_level}
The functionality of this level starts with the \textit{sensing} component that consists of \textit{low- and high-level sensing} and \textit{decision trigger logic}. \textit{Low-level sensing} maps the environment, detects the objects and the presence of a human, and extracts signals from the human (e.g., in our case a human's hands and head motions). Afterward, the \textit{high-level sensing} block takes the detected observables as a feature vector, and with the help of recognition algorithms, e.g., a human activity recognition (HAR) algorithm to recognize human gestures, it semantically describes the interaction environment. For instance, it concludes that a human grasped the red-colored object and put it onto the green container as part of a pick-and-place task or a human is looking around even though the task is awaiting. Finally, an observation vector is generated and forwarded to the \textit{decisional level} through a \textit{decision trigger logic} to update the decision-making (in Figure~\ref{fig:framework}). The logic synchronizes various observations and integrates rules to dynamically call for a decision update (instead of a constant frequency). This is to be able to respond reliably to irregular timings of human behaviors (our applied logic is detailed in Section~\ref{ssec:obsUpdate}). After a decision is generated from the \textit{decisional level}, the \textit{actuating} component processes it to generate motion commands for the robot actuators. This component semantically describes the abstracted decisions, e.g., the robot arm should navigate to the green-colored cube and grasp it, as a related action or a set of actions and goals. Finally, the \textit{action planning} mechanism buffers the semantically described actions coming after each new decision and prioritizes them to execute. It plans trajectories and generates motion commands for the robot actuators. It runs a control loop that receives the sensor feedback from the \textit{sensing} component until the actuators reach the goal stated by the current action command. As the design of sensing and actuating components are system- and environment-specific, we detail their implementation in Section~\ref{sec:deployment}.

The \textit{memory} component stores the learned models, i.e., the experience, of different contexts for a cobot's future operations. The \textit{world model} component of the \textit{memory}, as shown in Figure~\ref{fig:framework}, contains information on the environment the interaction takes place. The \textit{action planner} retrieves the knowledge of an interaction space for a collision-free trajectory. For example, in our context, the environment model contains the information on the location of all the detected static objects like the containers and the conveyor belt in our collaboration setup. Most importantly, the \textit{memory} block stores our novel intention-aware cobot decision-making models in the \textit{decision model library} as different decision strategies, which are encoded as POMDPs and MDPs. The reason for having multiple decision models follows our discussions that rather than a single adaptive decision model, sometimes a cobot may need to follow completely different decision-making strategies, i.e., policies, for a more reliable and faster adaptation, especially when adapting to humans. For that, we design several A-POMDP models with distinct intrinsic parameters (in Section~\ref{ssec:apomdp}). Each of these models generates a unique set of policies that are stored in the library and may be selected as the best strategy for a unique interaction type, which is assumed initially unknown to the cobot but to be estimated (in Section~\ref{ssec:abps}). In conclusion, the \textit{sensing, actuating, and memory} blocks are crucial for the deployment of our framework to satisfy a cobot's autonomous, reliable, human-in-the-loop, and long-term collaboration.

\subsection{Decisional Level: Anticipatory HRC}
\label{ssec:decisional_level}

Our primary novelty lies in the \textit{decisional level} of our framework. The decision-making solutions applied at this level are detailed in the following subsections. Here, we provide the entire anticipatory decision-making process of FABRIC summarized in Algorithm \ref{algo:abps}. The algorithm implements a nested loop at its core: The outer loop (starting at \textit{line 3}) iterates between the tasks and runs the \textit{adaptive policy selection} process of FABRIC before a new task starts; the inner loop (starting at \textit{line 6}) iterates during a task and runs the selected policy over the task (i.e., the \textit{decision-making} component of FABRIC).
In our application, we select a new policy once at the beginning of each collaboration task since our decision models generating these policies, i.e., A-POMDPs, are designed with a state machine that starts and terminates with a task (in Figure~\ref{fig:apomdp}). That is, a policy terminates after the task ends and another policy is to be selected at the outer loop for the next task. Hence, we refer to the outer loop as \textit{long-term adaptation} since it assumes the context might change from one task to another when compared to changes during a task, i.e., the inner loop, which is rather short-term in our case.

\begin{algorithm}
	\caption{Anticipatory Decision-Making of FABRIC}\label{algo:abps}
	\algsetup{linenosize=\small}
	\small
	\begin{algorithmic}[1]
		%\REQUIRE (Refering to FABRIC in Figure \ref{fig:framework}) Interaction type space ${\mathlarger{\mathlarger{\mathlarger{\tau}}}}$ (i.e., human types, in our case), the \textit{decision model library} of FABRIC with policies $\Pi$, an observation vector for the observed human behaviors $\sigma^h$ and the observed environment $\sigma^e$ in an observation space $\Omega$, an observation model to match observables to known interaction types $P(\Omega|\mathlarger{\mathlarger{\mathlarger{\tau}}},\Pi)$, utility as accumulated discounted reward obtained from running a policy $U$, a performance model $P(U|\mathlarger{\mathlarger{\mathlarger{\tau}}},\Pi)$, number of collaboration tasks $K$, exploration heuristics $\mathlarger{\mathlarger{\mathlarger{\upsilon}}}$.
		\REQUIRE (Refering to FABRIC in Figure \ref{fig:framework}) 
		\begin{itemize}[leftmargin=*]
		    \item Interaction type space ${\mathlarger{\mathlarger{\mathlarger{\tau}}}}$ (i.e., human types, in our case),
		    \item \textit{Decision model library} of FABRIC with policies $\Pi$,
		    \item Observation vector for the observed human behaviors $\sigma^h$ and the observed environment $\sigma^e$ in an observation space $\Omega$,
		    \item Observation model to match observables to known interaction types $P(\Omega|\mathlarger{\mathlarger{\mathlarger{\tau}}},\Pi)$,
		    \item Utility as accumulated discounted reward obtained from running a policy $U$,
		    \item Performance model, $P(U|\mathlarger{\mathlarger{\mathlarger{\tau}}},\Pi)$
		    \item Exploration heuristics $\mathlarger{\mathlarger{\mathlarger{\upsilon}}}$,
		    \item Number of collaboration tasks $K$.
		\end{itemize}
		\STATE Train offline for performance and observation models. For that, we run several collaboration tasks between all types in ${\mathlarger{\mathlarger{\mathlarger{\tau}}}}$ and all robot policies in $\Pi$.
		\STATE Initialize a belief: $\beta^0$ uniform distribution from the prior ${\mathlarger{\mathlarger{\mathlarger{\tau}}}}$. ($\beta$ is the belief on the current interaction type, i.e., the output of the \textit{interaction type estimation} component of FABRIC.)
		\FOR{task IDs $t = 1 ... K$}
		\STATE The \textit{policy selection} component selects a policy $\pi^{t}\in\Pi$ using $\beta^{t-1}$ and performance model, $P(U|\mathlarger{\mathlarger{\mathlarger{\tau}}},\Pi)$, using $\mathlarger{\mathlarger{\mathlarger{\upsilon}}}$ in Equation \eqref{eq:abps_ei}.
		%$\arg\max_{\pi \in \Pi} \upsilon_{\pi} = \mathlarger{\mathlarger{\mathlarger{\upsilon}}}(\pi,\beta^{t-1})$
		\STATE Forward the decision model generated the selected policy, $\pi^{t}$, to the \textit{anticipatory decision-making} component of FABRIC to solve and execute it online over the task.
		%\STATE \textbf{\textit{wait}}(until the task $t$ is completed. In the meantime, $\pi^{t}$ is executed that follows our A-POMDP state machine in Figure /ref{fig:apomdp})
		\WHILE{task $t$ continues}
			\STATE \textbf{wait} for the new $\sigma^h$ and $\sigma^e$ from the \textit{sensing} component.
			%\STATE Collect the new $\sigma^h$ and $\sigma^e$ vectors from the \textit{sensing} component.
			\STATE The \textit{state estimation} component updates the belief state, i.e., a state is $s \in S$ of our A-POMDP model design in Figure \ref{fig:apomdp}, using $\sigma^h$ and $\sigma^e$.
			\STATE Solve the A-POMDP for an action decision. In our case, an action is $a \in A$ of our A-POMDP model design in Figure \ref{fig:apomdp}.
			\STATE Forward the action decision to the \textit{actuating} component and execute it on the world.
		\ENDWHILE
		\STATE Obtain all of the collected observation vectors, $\sigma^{t} = (\forall \sigma^h) + (\forall \sigma^e)$, emitted during the task $t$.
		\STATE The \textit{interaction type estimation} component updates belief $\beta^{t}$ using $\sigma^{t}$ by belief update function in Equation \eqref{eq:abps_belief}.
		\ENDFOR
	\end{algorithmic}
\end{algorithm}

%TODO: how to describe that we forward the model not the policy to allow it solve for another in case different may be required for the short-term adaptation.

Starting with the outer loop, a policy is selected according to the current interaction type estimated (\textit{line 4} of Algorithm~\ref{algo:abps}). In FABRIC, the type is an abstract term to represent the long-term dynamics of collaboration. For instance, it can identify the changing requirements of a collaboration task, which may require a different strategy for the cobot.
In our case, the interaction type refers to the unknown type of the human collaborator, i.e., her characteristics, such as the level of her expertise or stamina. For the type estimation, an interaction type space ${\mathlarger{\mathlarger{\mathlarger{\tau}}}}$, an observation model to match observables to known types, and a belief distribution $\beta^0$ from the prior ${\mathlarger{\mathlarger{\mathlarger{\tau}}}}$ are initialized. The current belief on the type ($\beta^t$, where $t$ defines the current task ID) is always updated using the observation model and the collected sets of observations from the previous interactions (in \textit{lines~12 and~13} of Algorithm~\ref{algo:abps}). The most suitable policy is then selected for the current task from the \textit{decision model library} $\Pi$ according to the current belief on the type $\beta^t-1$ and a performance model that matches types to policies (at \textit{line~4}). For example, the cobot may select a policy that more likely assists the collaborating human during a task if her type is estimated to have beginner-level expertise for the task. The entire policy selection process is detailed in Section~\ref{ssec:abps}.
%The interaction type estimation is based on an observation model and the policy selection is based on an performance model trained offline.

%TODO: SHORT THE PARAGRAPH BELOW ! 
After the policy selection process at the outer loop, the selected decision model, e.g., an A-POMDP model in Figure~\ref{fig:apomdp}, is forwarded to the \textit{anticipatory decision-making} component of FABRIC to solve and execute it online during the task.
Then, the cobot is ready for the new collaboration task, i.e., the inner loop starting at \textit{line~6} of Algorithm~\ref{algo:abps}. We note that we use an online POMDP solver on the selected A-POMDP model to find the best robot action to execute at each time step (detailed in Section~\ref{ssec:apomdp}). This provides a faster and more reliable adaptation to the dynamics during the task, e.g., to the human collaborator. The online planning starts at \textit{line~7}. After the \textit{anticipatory decision-making} component receives the observations, at \textit{line~8} it estimates the current state of the collaborating human, the task, and the environment (i.e., the belief on the current A-POMDP state in Figure~\ref{fig:apomdp}). According to this estimation, the online solver runs the lookahead search on the A-POMDP model to find the best action decision to execute at each time (at \textit{line~9}). For example, the human may be estimated to have lost her attention during the task, which may lead to the cobot decision of acting on the task to assist the human (see the possible action decisions in Figure~\ref{fig:apomdp}). After an action decision is generated, the \textit{actuating} component of FABRIC processes it to generate motion commands for the cobot actuators to realize it in the environment (at \textit{line~10}), which ends one iteration of the inner loop. 
The next iteration starts at \textit{line~7} after receiving the new observations that are generated in the \textit{sensing} component of FABRIC. Since the human responses are dynamic and their timing is arbitrary, we do not follow a constant frequency to iterate the inner loop. Instead, we design a logic that triggers a new decision for the cobot in response to human actions. This logic (in Figure~\ref{fig:decisionTrigger}) is detailed in Section~\ref{ssec:expcognition}. After the task ends, the inner loop ends, another policy selection takes place, and the algorithm waits until a new task starts. The entire decision-making process at the inner loop, i.e., the short-term adaptation, is detailed next. 
%Due to the dynamic nature of human responses, the cobot needs to generate action decisions with a dynamic response time to respond reliably to humans. For that, the inner loop does not iterate with a constant frequency but a new loop is triggered whenever a new observation is received at the \textit{line~10}. The observations are handled in the \textit{sensing} component and we provide a human-aware observation triggers that is detailed in Section~\ref{ssec:expcognition}. The entire decision-making process at the inner loop, i.e., the short-term adaptation, is detailed next. 

\subsubsection{Anticipatory Decision-Making Component}
\label{ssec:apomdp}
\begin{figure}[!htbp]
 \centering
 \includegraphics[width=1\columnwidth]{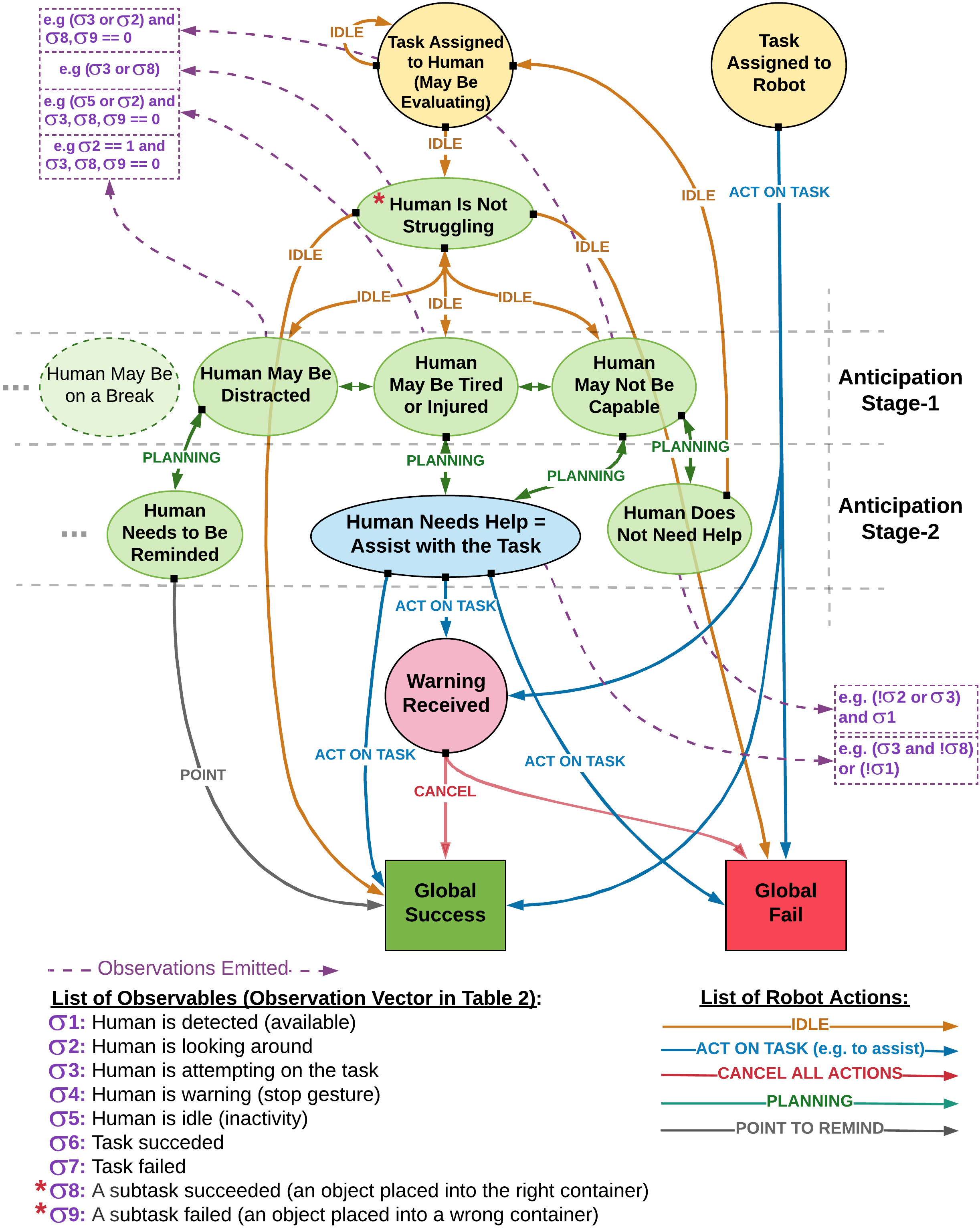}
 \caption{Anticipatory robot decision-making model, A-POMDP, adapted from \citep{Gorur2018} for the real user studies and the collaboration environment. \textcolor{red}{\textbf{*}} marks the new additions; a new state added to track if human is not struggling with a task and the new observables, i.e., $\mathlarger{\mathlarger{\sigma}}8$ and $\mathlarger{\mathlarger{\sigma}}9$, are introduced to inform about the current subtask status.}
 \label{fig:apomdp}
\end{figure}

\hfill \break
This component implements the short-term adaptation of a cobot, i.e., the inner loop between the \textit{lines 6-11} of the Algorithm~\ref{algo:abps}. The adaptation focuses on unanticipated behaviors of humans and their changing preferences during a collaboration task in real-time. We propose a novel stochastic robot decision-making model, a partially observable Markov decision process (POMDP), which we call an Anticipatory POMDP (A-POMDP). This anticipates a human's state of mind in two-stages (see in Figure~\ref{fig:apomdp}). In the first stage, it anticipates the human's task-related changing \textit{availability, intent (motivation),} and \textit{capability} during the collaboration. This includes, but not limited to, anticipating the human's changing tiredness, attention, and task success (capability). In the second, it further reasons about these states to anticipate the human's true need for assistance. Our contribution lies in the ability of our model to handle these unanticipated conditions: 1) when the human's intention is estimated to be irrelevant to the assigned task and may be unknown to the cobot, e.g., motivation is lost, another assignment is received, the onset of tiredness; 2) when the human's intention is relevant but the human does not want the cobot's assistance in the given context, e.g., because of the human's changing emotional states or the human's task-relevant distrust for the cobot. 

We have previously implemented a basic version of the model and evaluated it in simulation to show that integrating this model into a cobot's decision-making process to handle unanticipated human behaviors increases the efficiency and naturalness of the collaboration \citep{Gorur2018}.
In this paper, we improve the model design to ensure its real-time interaction capability with real humans in a more realistic collaboration scenario. These improvements are mostly through the insights we obtain and the data collected from the previous simulation tests (see our real-world integration pipeline in Section~\ref{sec:pipeline}).
Our scenarios are not limited to turn-taking collaboration and the environment is rather unconstrained, leaving a human collaborator with flexible strategy-making and arbitrary behaviors. Hence, the biggest challenge in designing such a decision model is the greater uncertainty in the human states, their changes (i.e., state transitions), and the timing of these transitions. In a real-world setting, the time to finish executing an action for humans, including thinking, resting, observing, and picking and placing, would greatly differ from one person to another, and even for an individual during multiple tasks running consecutively for a long-time. As a result, our decision models cannot follow a constant update frequency for the decision-making and it should be aware that the transition probabilities may greatly differ from one collaboration to another.

Our A-POMDP model is defined as a tuple $\big\{S, A, T, R, \Omega, O, \gamma \big\}$. $S$ comprises the mental states of the human collaborator from the cobot's perspective, the global success and failure states that define the result of a task (terminal states), the states of a new task assigned to the human or to the cobot (initial states), and a state when the cobot receives a warning from the human for any reason; $A$ is the set of cobot actions and $\Omega$ is the set of cobot observations with the observation vector $\sigma \in \Omega$ from the human and the environment (listed in Figure~\ref{fig:apomdp}); $T$ is the state transition probabilities where $T(s'|s,a)$ gives the probability of a transition from state $s \in S$ to $s' \in S$ for a given robot action $a \in A$; $O$ is the observation probabilities where $O(\sigma|s',a)$ gives the probability of observing $\sigma$ in a new state $s'$ after taking action $a$; $R$ is the immediate reward the robot receives and $R(s,a)$ gives the expected reward for taking action $a$ while in state $s$; $\gamma$ is the discount factor for delayed rewards.  We solve the POMDP model for an optimal robot policy, $\pi$.
We generalize the states and the robot actions to comply with several collaboration tasks. The cobot actions are to wait for the human (\textit{idle}), plan for assisting action (\textit{planning}) and \textit{assist} the human, e.g., by directly acting on the task. Further details of our design strategies are given in \citep{Gorur2018}. In the previous work, our goal was to prove that whether our models could cover the unanticipated human states, and show the importance of doing so. In this paper, our goal is to extend the approach to more realistic real-world scenarios.

For that, we consider a collaboration task with multiple steps (in Section~\ref{ssec:exp_task}), which we call subtasks, as opposed to each successful attempt completing a task in our previous work. This suggests longer iteration in our POMDP state machines before reaching a terminal state (e.g., \textit{Global Success}).
To better track the state of a human collaborator during such a task, we introduce new states and observations to our A-POMDP tuple (see the red marks in Figure~\ref{fig:apomdp}). A partially observable state called \textit{Human Is Not Struggling} is added to indicate that the human is successfully achieving the subtasks (i.e., each successful placement of an object in our case), and the cobot most likely estimates and stays at this belief state unless the human repeatedly executes any unanticipated behaviors. In this case, the cobot may transition to any state in the \textit{anticipation stage-1}. The new model also favors a transition from this stage back to the \textit{Human Is Not Struggling} state if the human could keep up with the task again. Finally, the newly added observables reflect the success status of a subtask, as detailed in Section~\ref{ssec:obsVector}.

The A-POMDP receives a negative reward $R$ if the model reaches the \textit{Warning Received} state, that is when the cobot receives a warning from its human partner for any reason. This mostly happens when the cobot misjudges the human's need for assistance and wrongly takes over a task. Other than that, the immediate rewards are assigned for the terminal states of \textit{Global Success} and \textit{Global Fail}. Finally, we set a discount factor to avoid longer wait times before an object placement takes place. This encourages a cobot takeover if the human delays. We set a cooperative reward for the human-robot team to evaluate the performance of the collaboration during a task. Whenever the system detects a subtask success or a subtask failure, an immediate cooperative reward is assigned. It should be noted that the same rewards are assigned after a subtask success or failure regardless of who completes it.
%The discounted reward function given below is the final value a robot model collects after a subtask.
% \begin{equation} \label{eq:meth_subtask_reward}
% R_T = \sum\limits_{t=0}^{T} \gamma^t r_{t}
% \end{equation}
%\noindent where $R_T$ denotes the total discounted reward in a subtask, $r_t$ is an immediate reward between time $t=0$ and $t=T$ discounted by $\gamma^t$. Also, $t$ is the time at which a new observation is received. It is reset to $0$ at the end of each subtask (i.e., an object has been detected on a container). Finally, the total discounted reward in a task, given below in Equation \ref{eq:meth_task_reward}, is the sum of all subtask rewards and it is a direct indicator of the performance of a robot model during an episode of a collaboration.

% TODO: see what is written about the online solving already in the algorithm description above (!)
We use an updated version of DESPOT solver in \citep{Ye2017} to solve for and generate a policy online from our A-POMDP models. Our version is also capable of executing the policies in real-time against real humans. The online solver constructs a belief tree with the belief on the current state at the root of the tree and performs a lookahead search on the tree for a policy \citep{Ye2017}. This alleviates the computational complexity by computing good local policies at each decision step, which also allows for an online adaptation to the uncertainties in the environment or in the collaborating human's behaviors. The setting of the transition and observation probabilities are realized in multiple steps that involve simulation runs and training with real humans (following our continuous integration pipeline in Figure~\ref{fig:pipeline}, detailed in Section~\ref{ssec:training}).

% \begin{equation} \label{eq:meth_task_reward}
% R_{task} = \sum\limits_{subtask=1}^{n+1}(\sum\limits_{t=0}^{T} \gamma^t r_{t})
% \end{equation}
%
% \begin{table}[!hbp]
% 	\centering
% 	\caption{Immediate rewards assigned during the experiment}
% 	\begin{tabular}{l|c}
% 		\textbf{Observation} & \textbf{Immediate reward}   \\
% 		\hline
% 		warning received                & -3             \\\hline
% 		sub task success                & +6             \\\hline
% 		sub task failure                & -6              \\\hline
% 		all others                      & 0              \\
% 	\end{tabular}
% 	\label{tbl:immediate_rewards}
% \end{table}

\subsubsection{Adaptive Policy Selection Component}
\label{ssec:abps}
\hfill \break

Our goal is to extend the adaptation of a single intention-aware robot decision-making model, such as A-POMDP, to various and changing human characteristics in long-term for personalized collaboration. Here, we present a novel policy selection mechanism, which we call adaptive Bayesian policy selection (ABPS), that builds on top of existing intention- and situation-aware robot decision-making models for an extended adaptation of cobots to various human types. It handles the long-term adaptation of FABRIC, satisfying a personalized collaboration for a cobot. We have previously implemented a basic version of ABPS and evaluated it in simulation to show that such a mechanism extends a cobot's human adaptation and provides a more efficient and natural collaboration in long-term \citep{Gorur2019}. In this paper, we improve ABPS to ensure its real human collaboration and its system integration in a more realistic scenario.
ABPS is equipped with a policy library (i.e., the \textit{decision model library} of FABRIC in Figure~\ref{fig:framework}) to act appropriately in the context of some human types and tasks in the HRC domain. The selection process is handled in the \textit{adaptive policy selection} component of FABRIC. It is presented with an unknown interaction type, which must be solved within a limited time and small number of trials. As mentioned, we define an interaction type in our case as a human collaborator having an unknown type (i.e., characteristics) in a known task. The goal of ABPS is to select policies from the library for the new and possibly unknown human type, over which it has a belief distribution, while minimizing the total regret in a limited time. Minimizing the regret in our domain is defined as increasing the task success rate and decreasing the number of warnings received from the human collaborator (i.e., naturalness), relative to the best alternative from the library in hindsight.

%In Algorithm \ref{algo:abps}, we provide the entire anticipatory decision-making process of FABRIC.
At the core lies the \textit{decision model library} $\Pi$ that holds all robot decision models. Our goal is to limit the arbitrary generation of the policies to avoid overloading the library with unreliable candidates \citep{Albrecht2018}. For that, we generate new models moving from a base model design, which is an A-POMDP model with the connection scheme as in Figure~\ref{fig:apomdp}. The intrinsic parameters of the base model is trained after the calibration experiments described in Section~\ref{ssec:training}.
The offline generation of different robot models is done by adjusting $T$ and $O$ of the base model: the state and observation probabilities of the A-POMDP model corresponding to different human types (see Section~\ref{ssec:apomdp}). Changes in $T$ correspond to different transitions of a human's internal states, e.g. a model assumes the human tires faster, or the human needs more assistance when she is not capable. Changes in $O$ then define the observations emitted by the human as a function of her internal states. For example, a human not being able to handle the task could indicate that she is tired, or she is an inexperienced, both of which should be handled differently by the cobot. Additionally, by adjusting $O$, we are able to make the model a partially, mixed or fully observable Markov decision process (POMDP, MOMDP or MDP, respectively). We randomly adjust the probabilities to generate various models, each of which handles a unique human type, and solve for their optimal policies to construct our policy library $\Pi$. Through this random generation, we are agnostic to specific human types and behaviors, which also allows ABPS to integrate into any existing intention-aware model. We are aware that this random generation may still produce unreliable policies for the cobot. Nonetheless, reliability and usability of such policies become prominent after the training process (in Section~\ref{ssec:training}).

%TODO: Examine the exploration heuristic and decide if to put it here or not!!
ABPS measures the similarity between an unknown interaction type and previously known types, to identify which policies may be the best to reuse (in the \textit{interaction type estimation} component in Figure~\ref{fig:framework}. In this case, a collaborated human's type is latent and the human type space is not fully known. Therefore, a correlation between policies and a bounded set of human types is not possible.
%The similarity of types is extracted from offline training with some known types and by utilizing this trained model online, constructing $\beta(.)$.
In fact, the space of human types is in general infinite, but we limit this to control complexity. Therefore, the construction of a type space $\mathlarger{\mathlarger{\tau}}$ is a crucial process. For this purpose, we train an estimation model from a set of known types and use it online to estimate a new unknown type as a belief distribution over the known ones, $\beta(.)$ (see Section~\ref{ssec:training}). In order to train such a model, we generalize some characteristic human features to approximate a human type. These are a human's changing levels of expertise, attention, stamina, and collaborativeness. These features are not exhaustive but they are inspired by \citep{Gombolay2017, Mcguire2018, Ji2006} and are stated to be crucial to be known by a cobot.
%The last term is a more general description of a human's acceptance rate of a robot's offer for assistance.
The type space consists of many human types by adjusting the level of these features, e.g. a human with beginner skills, pensive, low stamina and non-collaborative behaviors (e.g. always rejecting a robot's assistance due to distrust). We argue that any human worker can be represented as a distribution of such features in our experiments.
%More details on the simulated human types in type space $\mathlarger{\mathlarger{\tau}}$  are given in Section~\ref{sec:abps_human}.
The interaction type estimation model is used by ABPS as \emph{a priori} information, which we call the \textit{observation model}.

{\setlength{\parindent}{0cm}
	\begin{mydef}[\textbf{Observation model}]\label{def:abps_obs}
		For a robot policy $\pi$, an interaction (i.e., human) type $\tau$ and an observation vector $\sigma$ obtained from the human actions and the environment, the observation model $P(\sigma|\tau,\pi)$ is a probability distribution over the observation signals $\sigma\in\Omega$ that results by applying the policy $\pi$ to the type $\tau$.
\end{mydef}}

All combinations of known human types in $\tau$ and the robot policies in the library are run against each other offline several times to generate our \textit{observation model} (in \textit{line 1} of Algorithm~\ref{algo:abps}, detailed in Section \ref{ssec:training}).
The observation signals are emitted by the collaborated human and the environment, reflecting a human's actions and their impact on the task and the environment. In our experiments, an observation vector, $\sigma\in\Omega$, is a 9-D boolean vector as listed in Figure~\ref{fig:apomdp} and detailed in Table~\ref{tbl:observationVector}. The ABPS agent receives these observables at every episode of a task and accumulates them to update its belief on the human type after a task terminates (see \textit{line 12, 13} of Algorithm \ref{algo:abps}).
Finally, the type belief update is Bayesian, given by

\begin{equation}\label{eq:abps_belief}
\beta^t(\tau)=\dfrac{P(\sigma^{t}|\tau,\pi^{t})\beta^{t-1}(\tau)}{\sum_{\tau' \in \mathlarger{\mathlarger{\tau}}} P(\sigma^{t}|\tau',\pi^{t})\beta^{t-1}(\tau')},      \enspace\enspace \forall \tau \in \mathlarger{\mathlarger{\tau}}
\end{equation}

{\setlength{\parindent}{0cm}
	where $\beta^{t-1}$ is the previous belief and $P(\sigma^{t}|\tau,\pi^{t})$ is the probability of observing $\sigma^{t}$ after applying $\pi^{t}$ in an interaction with any human type $\tau$. This distribution is directly retrieved from the \textit{observation model} for each requested type and policy.
}
%%%%%%%

The policy selection process of the robot is based on an exploration heuristic called \textit{expected improvement (EI)} \citep{Rosman2016}. As stated in \textit{line 4} of Algorithm \ref{algo:abps}, this algorithm runs on another trained \emph{a priori} model called the \textit{performance model}.

{\setlength{\parindent}{0cm}
	\begin{mydef}[\textbf{Performance model}]\label{def:abps_perf}
		The performance model, $P(U|\tau,\pi)$, is a probability distribution over the utility, $U$, of a policy $\pi$ when applied to interaction (i.e., human) type $\tau\in\mathlarger{\mathlarger{\mathlarger{\tau}}}$.
\end{mydef}}

The system utility, $U$, is the accumulated discounted reward received after a policy is run (see Section~\ref{ssec:apomdp} for the immediate rewards a robot obtains during a task). All the combinations of known human types $\tau \in \mathlarger{\mathlarger{\tau}}$ and the robot policies $\pi \in \Pi$ are repeatedly run against each other offline to generate our \textit{performance model} (see Section~\ref{ssec:training}). Then, this model is used by the policy selection heuristic. The heuristic assumes that there is a $U^+$ in reward space which is larger than the best estimate under the current type belief, $U^\beta$. A probability improvement algorithm can be defined to choose the policy that maximizes Equation \eqref{eq:abps_pi} and achieves the utility $U^+$.
\begin{equation}\label{eq:abps_pi}
\pi' = \arg\max_{\pi\in\Pi}\sum_{\tau\in\mathlarger{\mathlarger{\tau}}}\beta(\tau)P(U^+|\tau,\pi)
\end{equation}
Because the choice of $U^+$ directly affects the performance of the exploration, its selection is crucial to the performance of this exploration. The \textit{expected improvement} approach instead addresses this nontrivial selection of $U^+$. The algorithm iterates through all the possible improvements on an existing $U^\beta$ of the current belief, which satisfies $U^\beta < U^+ < U^{max}$. The policy with the best potential is then chosen, as given in Equation \eqref{eq:abps_ei}.

\begin{equation}\label{eq:abps_ei}
\pi' = \arg\max_{\pi\in\Pi}\int_{U^\beta}^{U^{max}}\sum_{\tau\in\mathlarger{\mathlarger{\tau}}}\beta(\tau)P(U^+|\tau,\pi)dU^+
\end{equation}
\begin{equation}\label{eq:abps_ei2}
= \arg\max_{\pi\in\Pi}\sum_{\tau\in\mathlarger{\mathlarger{\tau}}}\beta(\tau)(1 - F(U^\beta|\tau,\pi))
\end{equation}

{\setlength{\parindent}{0cm}
	where $F(U^\beta|\tau,\pi) = \int_{-\infty}^{U^\beta}P(u|\tau,\pi)du $ is the cumulative distribution function of $U^\beta$ for a $\tau$ and $\pi$. The algorithm, therefore, selects the robot policy with the most likely improvement on the expected utility.
} Finally, once a policy is selected, the A-POMDP model that has generated it is forwarded to the \textit{anticipatory decision-making} component of FABRIC to be executed online (see \textit{line 5} of Algorithm~\ref{algo:abps}). The process is repeated until all the scheduled tasks end.  

%\section{A Pipeline for Continuous Development and Integration of Interactive Robots}
\section{A Pipeline for Continuous Development and Integration of Cobots}
\label{sec:pipeline}

\begin{figure*}[!htbp]
	\centering
	\includegraphics[width=0.7\textwidth]{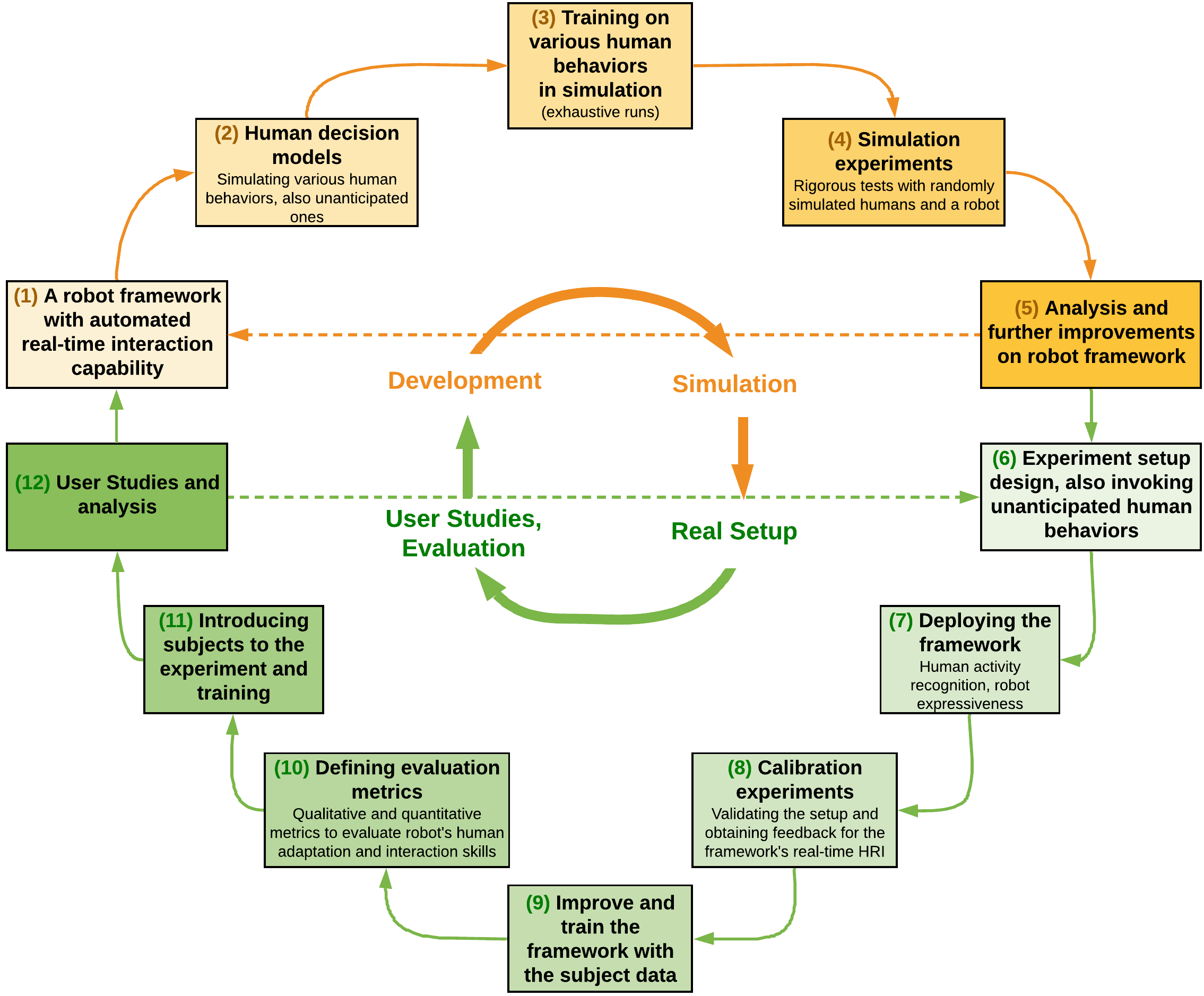}
	\caption[Evaluation Pipeline]{Our pipeline for integration, deployment and  evaluation of an anticipatory collaborative robot with extended human adaptation. The development, deployment, simulation experiments, and user studies are iterated while the framework trains on and is tested against a greater diversity of human behaviors.}
	\label{fig:pipeline}
\end{figure*}

When deployed for interaction with humans, robots face a great deal of uncertainty in the long-term. This is majorly due to the lack of diversity of human behaviors available during the training and validation processes of the interactive robots. To decrease the uncertainty, we follow the conventional way of developing autonomous systems, i.e., rigorously testing a system in simulation, and then deploying and validating it in a real environment. To do so with interactive robots, we propose a pipeline in Figure~\ref{fig:pipeline}, to systematically train and evaluate a robotic framework with a vast range of short- and long-term dynamics of humans. The pipeline takes an interactive robot framework, runs it through simulations, designs a dynamic and unconstrained environment, deploys it for user studies and follows a continuous development and integration practice by iteratively training on various human behaviors.

First, a simulation environment with simulated humans running realistic decision models is needed to start the training and testing processes of an interactive robotic framework (\textit{step 2} of the pipeline in Figure~\ref{fig:pipeline}). Interactive and collaborative robots mostly rely on the data collected from people participating in user studies, which are mostly in a confined space with limited observable human intentions, e.g., due to constrained environments. This makes it challenging to reuse and benchmark such robotic solutions \citep{Rajendran2020}. Through sampling methods, e.g., Monte Carlo, running on simulated human models, we can generate a diversity of human behaviors including the unanticipated ones that are hard to observe in lab environments (see our simulation in Section~\ref{ssec:humanSimu}). In \textit{step 3} of the pipeline, the simulations enable a robot framework to train on this large-scale human data. Then, in \textit{step 4}, the same simulation environment is used to rigorously test and evaluate the framework.

Even though we use the same simulation environment for training and testing, the randomly generated human models and the sampling of human behaviors at every run avoids overfitting of the trained models and provides enough diversity for their evaluation. Also, we highlight the importance of designing a 3D simulation environment with real-world physics to introduce changing dynamics of the manipulation tasks, e.g., unknown and dynamic timing and strategy for a human to complete an action. This greater uncertainty and diversity provide feedback during the evaluations, which is both qualitative with observations from the simulations and quantitative like collected rewards for a robot's decision-making. In \textit{step 5} of the pipeline, these are analyzed to identify and handle unnatural (e.g. unreliable or unexpected) or inefficient robot interactions before the framework's real-world deployment. We would then repeat \textit{steps 3-5} of the pipeline. The details of our simulation environment are in Section~\ref{ssec:simulation}.

For validation, a user study experiment should be realistic in the sense that human participants should not be confined to a structured environment with limited interactions. However, it is also difficult to have humans in real user studies convey a full range of possible diverse behaviors. \textit{Step 6} of the pipeline, therefore, focuses on the open task of transferring the results from the simulation targeting a wide range of human behaviors into a real-world experiment and validate it by means of user studies. Additionally, most of the existing cobot solutions are structured around command and response patterns or turn-taking with previously set roles, which limits the fluency of a collaboration \citep{hoffman_evaluating_2019}. Hence, our design goals in \textit{step 6} are twofold: 1) A setup should foster unanticipated and uncommon human behaviors; 2) It should let a human and the cobot freely act on the task and flexibly replan to compensate for such behaviors. For that, our experiment setup design must place a cognitive load on humans to invoke such human behaviors (detailed in Section~\ref{ssec:realenv}).

Physically interacting with humans in real-time in such an unconstrained environment brings about a number of challenges for an autonomous robot. In HRC, poor coordination between the human and the cobot is likely to result in a collaboration that is neither effective nor perceived as natural. In \textit{step 7}, the pipeline deals with these challenges to ensure a reliable coordination between the cobot and human in real-time. There are a variety of different human characteristics with various preferences; as a result, recognizing human actions and understanding human behaviors show a great diversity due to their unique reaction times and different ways of reflecting the same intentional and emotional behaviors, i.e., hidden intentions. For example, one person may idle too long when evaluating a subtask whereas for another this would mean they need assistance. This is especially the case when an interaction does not follow a turn-taking approach. Thus, different reaction times and strategies need to be considered in the decision models for reliable responses. In addition, effective collaboration means good coordination of intentions and actions \citep{villani_survey_2018}, and communication plays a decisive role in this. The cobot should then be equipped with intention expressive gestures that effectively communicate its decisions. Hence, in \textit{step 7}, we need to develop: 1) human activity recognition solutions, 2) adaptive decision update frequencies for the robot to respond timely to dynamic human behaviors in real-time, and 3) expressive robot motions to effectively communicate the robot's action decisions to a human.

The rest of the pipeline focuses on systematic user studies. In \textit{step 8}, we conduct calibration experiments with real humans to validate \textit{steps 6} and \textit{7} and to provide feedback for further improvements and training in the next step. In these studies, a first batch of participants interact with the cobot running the framework to evaluate the activities under \textit{step 7} and, in general, the reliability of the cobot's interaction. In the second batch, we evaluate the experiment setup and our design goals to see whether we could invoke the unanticipated human behaviors by comparatively testing several interaction scenarios through within-subject experiments. These experiments are briefly detailed in Section~\ref{ssec:eval_calib}. In \textit{step 9}, we first process the results of \textit{step 8} to improve the setup and the framework for its real-time interaction capability, then we train the decision-making models with the real human data obtained from \textit{step 8} along with the simulation data.

The next steps are conventional steps in most of the user studies. In \textit{step 10}, the performance metrics need to be defined to evaluate the framework's interaction and human adaptation goals. In general, the participants are asked to evaluate a robot's adaptation skills, whereas in the HRC context, this can also be objectively analyzed from the efficiency metrics defined for the collaboration. In our case, we have designed a warning gesture for the participants to reflect their dissatisfaction with the cobot responses, which also numerically reflects the reliability of the cobot adaptation. Finally, in \textit{step 11}, we introduce the participants to the experiment setup and do training runs to familiarize them with the environment. This way, we avoid the practice effect in our comparative analysis and partially let the novelty effect fade away. We believe that this step is crucial to more realistically evaluate human adaptation skills of a robotic framework. Then, we conduct user studies in \textit{step 12}. The user study and simulation loops are iterated with different participants and various human models until the robot framework performance converges to steady state.

\section{Applying Our Framework through the Pipeline}
\label{sec:deployment}
In this section, we train and deploy our framework given in Figure~\ref{fig:framework} following the pipeline in Figure~\ref{fig:pipeline} both to give an example of how the pipeline works and to evaluate and validate the framework and our discussions on short- and long-term human adaptation of cobots. As in the pipeline, we divide this section into simulation and real setup applications. A figure summarizing and listing all of our activities following the pipeline is given in the supplementary materials.

\subsection{Simulation Environment}
\label{ssec:simulation}
\begin{figure}[!htbp]
	\centering
	\includegraphics[width=\columnwidth]{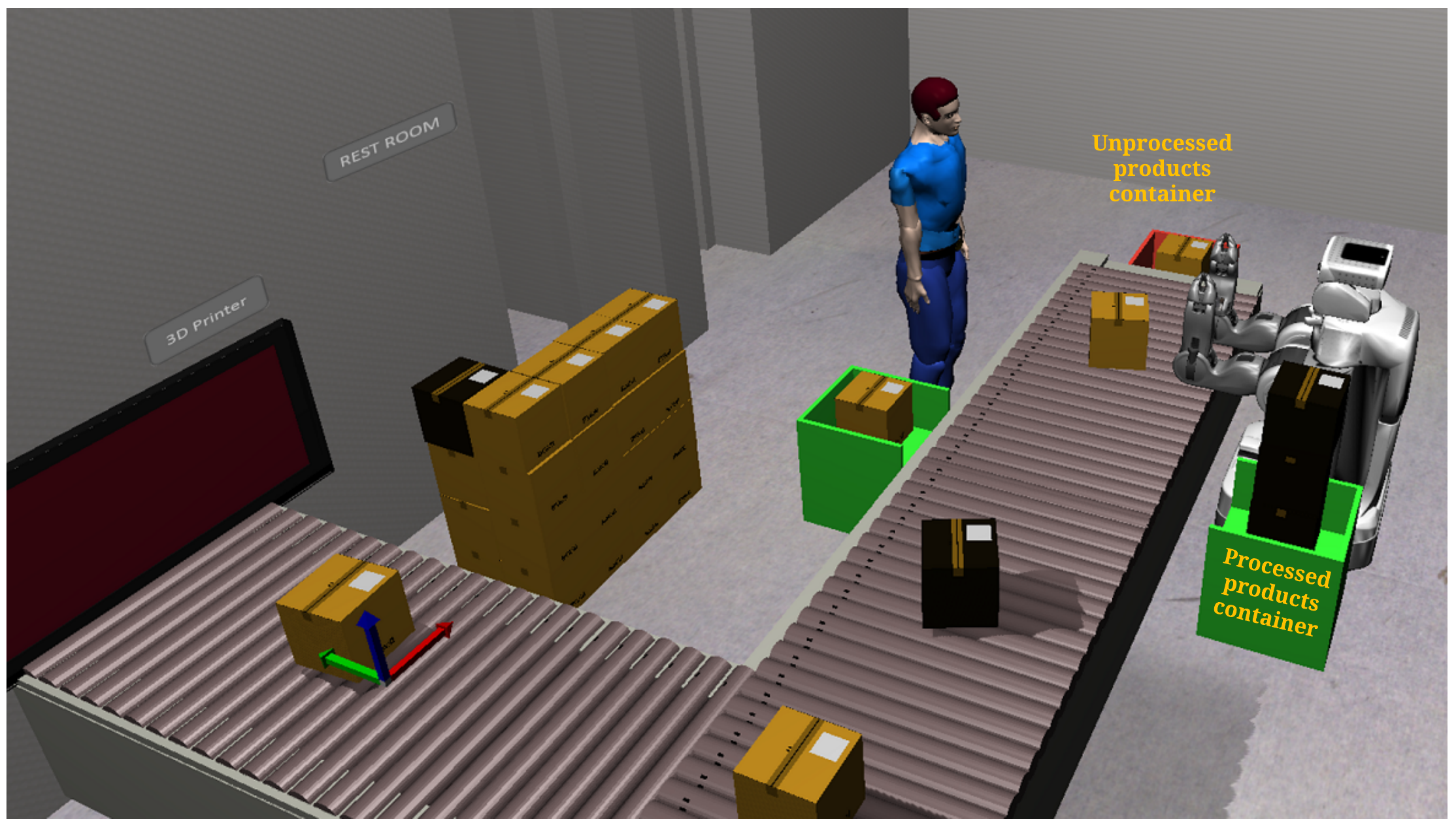}
	\caption{Simulation of an HRC at a conveyor belt for the task of product inspection and storing.}
	\label{fig:simu_env}
\end{figure}

This section follows \textit{steps 2-5} of the pipeline in Figure~\ref{fig:pipeline}. For this purpose, we devise a 3D simulation environment to simulate our HRC scenario in a factory environment on a conveyor belt. This environment has already been used to train and evaluate both \textit{anticipatory decision-making} (i.e., A-POMDP models) and \textit{adaptive policy selection} components (i.e., ABPS) in G\"or\"ur et al. \citeyearpar{Gorur2018} and in G\"or\"ur et al. \citeyearpar{Gorur2019}, respectively. In this section, we detail the environment and our novel simulated human models. The simulation, in Figure~\ref{fig:simu_env}, consists of existing human and robot models (PR2 is selected but we are indifferent to the robot hardware), a conveyor system, produced packages, two containers for processed products, a container for unprocessed products and a restroom. All of our scenarios consist of several sequential task assignments to simulate long-term collaboration. A task in the simulation is a product inspection and storing job. It starts with a user-defined task assignment. A task is successful when the product is inspected and put into the processed-product containers either by the human or by the robot. The conveyor belt waits for a certain time for a package to be processed, and then runs and the product falls into the uninspected-product container leading to a task failure.

Both the robot and the human are controlled by their decision models whereas the generated action decisions are executed in the MORSE simulator. The robot's decisions are generated by our anticipatory decision-making approaches. The observations the robot receives (in Section~\ref{ssec:obsVector}) are the 3D human body joints that are always available directly from the simulated human model and the proximity sensors placed inside the containers to monitor the task status as succeeded, failed, or ongoing.
A state-of-the-art HAR module, inspired by Roitberg et al. \citeyearpar{Roitberg2014}, has been implemented to recognize the constrained and distinct simulated human gestures from the body joints available, constructing the \textit{sensing} component of FABRIC.
We have designed certain human actions that are required for our scenario, that are also observations for the robot. Based on the available list of actions, we have designed human models that simulate a variety of human decisions. Just like the robot, the human decisions are executed in MORSE, and the human decision model receives observations for the current state of the environment (including the robot motions) as feedback. Hence, simulated humans are also automated agents. The timing and the duration of human actions are dynamic with random factors, in order to contribute to the uncertainty of the environment. In summary, this environment allows for a fully automated long-term HRC to train and test robot decision-making solutions under various conditions.

\subsubsection{Human Simulation: Human Decision Models}
\label{ssec:humanSimu}
\begin{figure}[!htbp]
	\centering
	\includegraphics[width=\columnwidth]{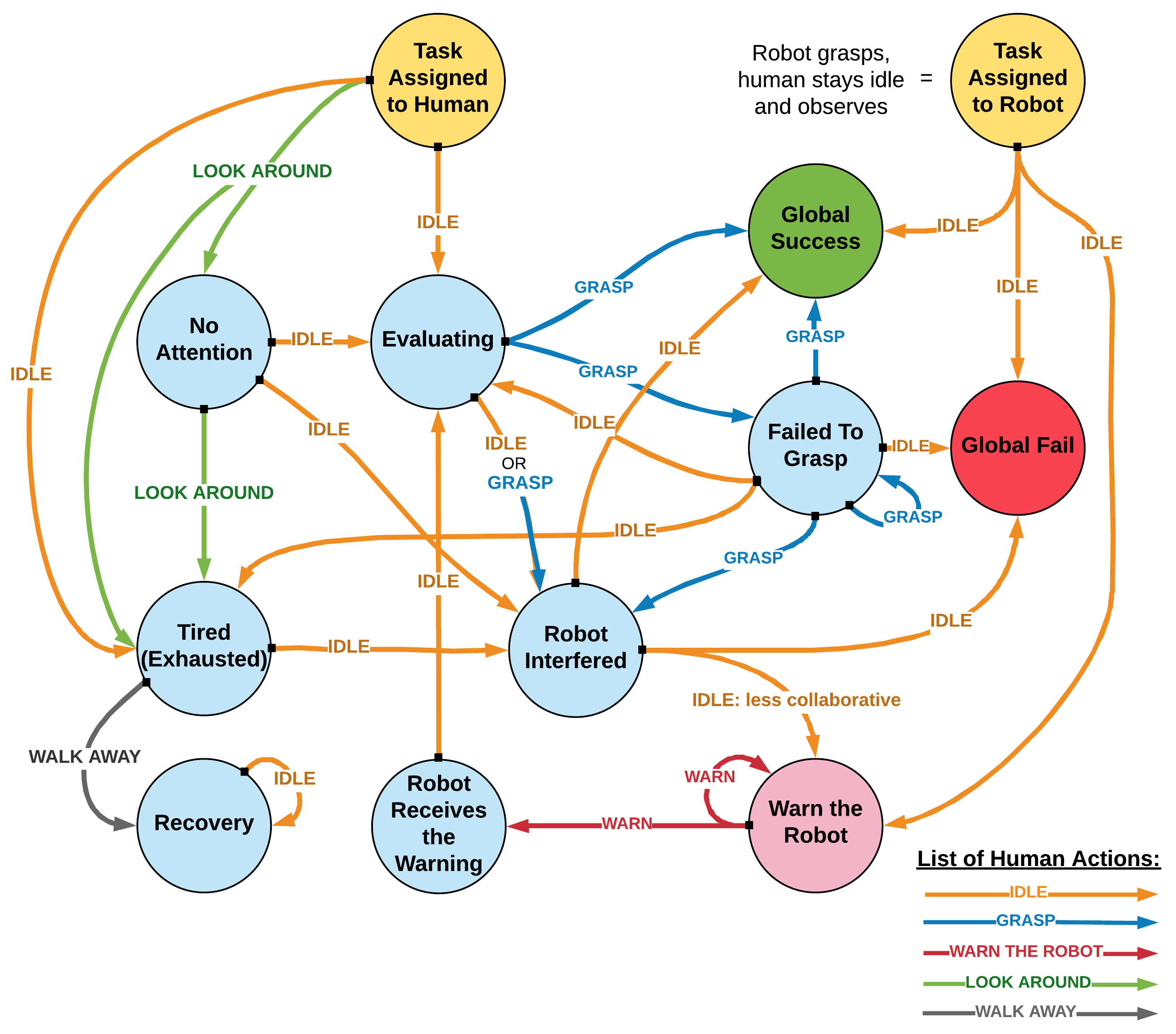}
	\caption{Human MDP model in detail, showing state-action connections for the most probable transitions.}
	\label{fig:human_model}
\end{figure}

Simulating humans allows us to scale the experiments to emulate many different combinations of human behaviors, including unanticipated ones. Our goal is to create use-cases where a human worker follows the aforementioned unanticipated conditions and occasionally performs behaviors like stubbornly rejecting the robot's help, tiring quickly, being easily distracted, and being distrustful of the robot. A representative proof of concept human decision model is built using an MDP as shown in Figure~\ref{fig:human_model}. We note that only the transitions with non-negligible probabilities are shown in the figure. Our model design is inspired by available studies analyzing human workers operating on repeated tedious tasks in a workplace \citep{Gombolay2017, Mcguire2018, Ji2006}. We assume that a human worker optimizes an objective function to reach her goal. However, following our statement, this may also be an internal goal irrelevant to the assigned task, e.g., leaving their place for a short break. We also assume that any human actions may be imperfect \citep{Hiatt2017}. Simulating such a human has been shown to be accurate using an MDP to generate a policy for a human agent \citep{Bandyo2013}.

Our human MDP is a tuple $\big\{S, A, T, R, \gamma \big\}$ where $S$ is the human states of mind, $A$ is the human actions, $T$ is the state transition probabilities, $\gamma$ is the discount factor, and $R$ is the immediate rewards received based on the result of a task and the type of the human to encourage that type of behavior, e.g., a distracted person receives positive rewards in the \textit{Global Success} and \textit{No Attention} states.
This model is inspired by our expectation that a human chooses an action based on the collaborated robot's action, the state of a task, the human's internal states, and her internal goals. Additionally, we govern a human's responsiveness to the interacted robot actions. Such responsiveness is handled through a transition function $T(s,a,s') = P(s'|s,a,n_{r},k_{t})$ for $s,s'\in S$, $a \in A$, the number of times the robot interfered in a task $n_{r}$ and the number of tasks handled so far $k_{t}$. That means we have dynamic transition probabilities changing over the course of the interactions, leading to updates on human models after each task, and so updated human behaviors. An example of such responsive behaviors is that a human becomes less collaborative as her robot partner selects wrong policies, e.g., the robot takes over a task (depicted as $n_{r}$) when the human was already planning to handle it. A decrease in collaborativeness is handled with an increased transition probability of the human to \textit{Warn the Robot} when a robot interferes with a task. Another example is that a transition to the state of being tired depends on the number of tasks already handled, $k_{t}$.

The model samples random but goal-oriented dynamics for the human collaborator using Markov Chain Monte Carlo (MCMC). In the end, the MCMC sampling and the responsive transition function lead to human simulations exerting dynamic behaviors changing in response to the robot decisions and with a small random factor. Through changing $T$ and $R$ and solving the model for a policy, $\pi$, we create various human types with changing characteristics (i.e., long-term behaviors). This modeling scheme is then used to automate the training and testing of our cobot's reasoning under various hard-to-predict conditions. We discuss our training process in Section~\ref{ssec:training}. Our simulation experiments, following \textit{steps 3-5} of the pipeline, for A-POMDP model (i.e., short-term adaptation) are in \citep{Gorur2018} and for the integrated anticipatory decision-making with ABPS (i.e., long-term adaptation) are in \citep{Gorur2019}. We show our analysis on the reliability of our human modeling scheme and the diversity of behaviors toward large scale training and tests in the supplementary materials.

\subsection{Real Environment and Setup Design}
\label{ssec:realenv}

\begin{figure}[!htbp]
	\centering
	\includegraphics[width=\columnwidth]{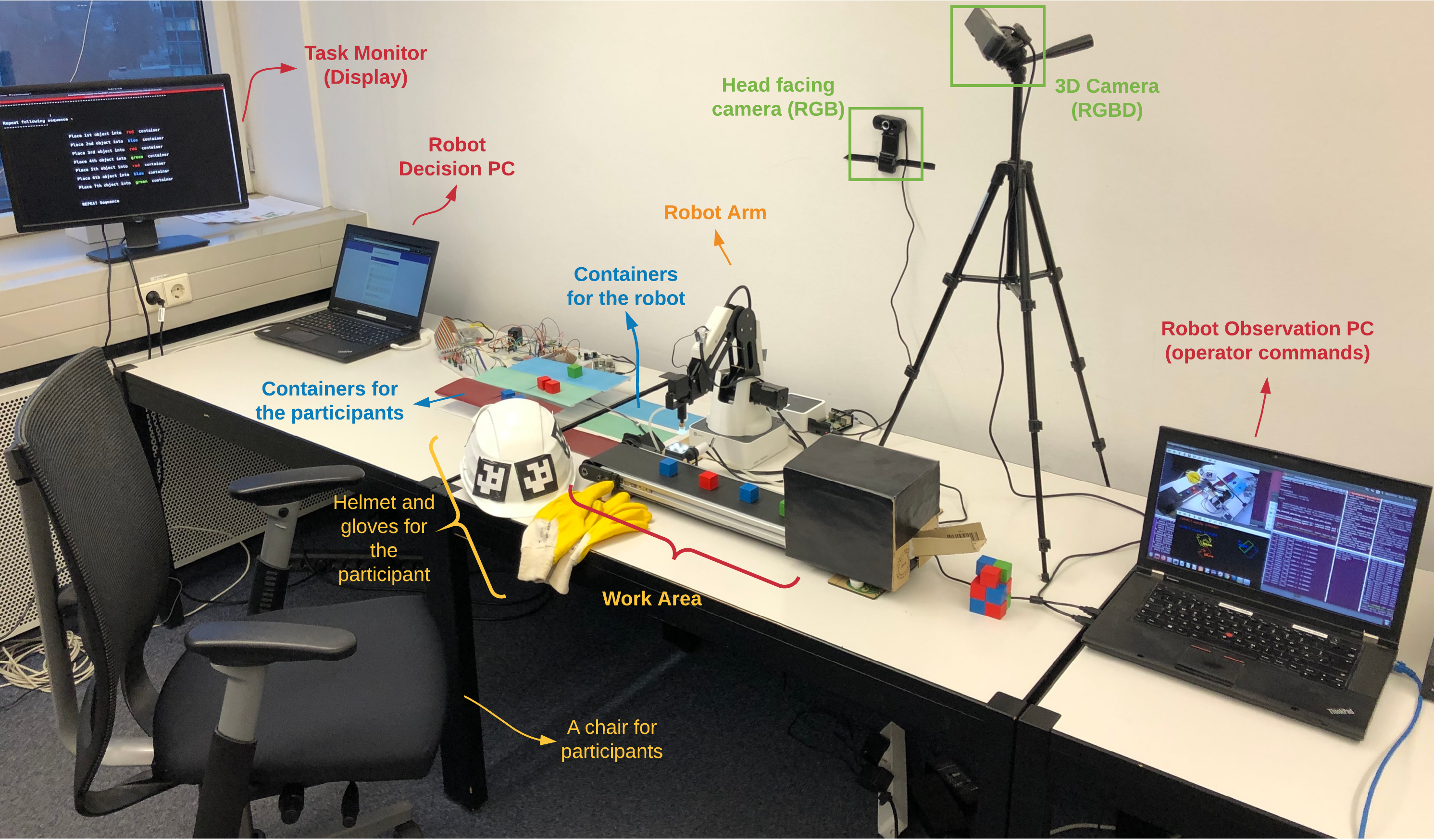}
	\caption{Real-world setup: Human-robot collaboration on a conveyor belt.}
	\label{fig:realsetup}
\end{figure}

This section details our activities on \textit{step 6} and \textit{step 7} of the pipeline (see Figure~\ref{fig:pipeline}).
%Our design goals, as mentioned, to invoke unanticipated human behaviors that may also lead to human errors and to let the human and the robot freely take turns and replan a task flexibly. For that, we first detail the collaboration setup (Section~\ref{ssec:exp_env}), then the use-case scenario and the collaboration task (Section~\ref{ssec:exp_task}), the integration of our robot framework to the setup and its real-time human-in-the-loop interaction capability to satsify a fluent collaboration (Section~\ref{ssec:exp_cognition}).
%We also briefly mention our design decisions for the motions (e.g., gestures) of our robot platform for explainability of the robot decisions (see in Section~\ref{ssec:exp_motions}).\todo{see that later}.
We devise a collaboration setup, shown in Figure~\ref{fig:realsetup}, that provides a rather unconstrained human intention space. Moreover, we design a cognitively exhaustive task of sorting colored cubes continuously flowing on a conveyor belt and placing them into relevant colored containers according to complex rules. Through inducing a cognitive challenge, our goal is to observe various human characteristics, e.g., a competitive person with a bad memory, and invoke unanticipated behaviors, e.g., constantly rejecting the robot's assistance, lost attention and lost motivation. Our goal is to show that the environment is realistic and able to induce a desired cognitive load to invoke such diverse human behaviors. Additionally, we do not enforce turn-taking collaboration in the task, allowing a human and the robot to flexibly replan task allocations and freely act based on their estimate of the partner's behaviors. As a result, this allows for evaluating a broader range of adaptation skills leading to the validation of our framework.

\subsubsection{Collaboration Scenario and the Task Design}
\label{ssec:exp_task}

Our collaboration scenario starts after a human participant sits on the chair and wears the yellow gloves and the safety helmet shown in Figure~\ref{fig:realsetup}. These are used to recognize human activities. The experimenter selects a task and the task rules are displayed on the \textit{task monitor} as shown in Figure~\ref{fig:realsetup}. Every task consists of several object sortings and placements. The task description is visible only for a certain amount of time and then it disappears as the participant's job is to memorize it (i.e., a another cognitive challenge).
The selected task starts right after the rules disappear. The conveyor belt then starts to move and transports a wooden cube towards the human and the robot. Our robot arm, we call it the cobot for simplicity, picks the cubes with suction yet for simplicity we also refer it as grasping. A cube is available to be grasped once it stops in front of an infrared sensor on the belt. The participant and the cobot might now decide to grasp the object and place it in one of the containers according to the task description. After each placement, either by a human or by the cobot, the status of which cube is placed on which container is displayed live on the \textit{task monitor} without the information on whether the placement is correct or not. This is only for the participants to keep a track of the task with many cube placements yet introduces a distraction on a continuous task flow. Once a maximum number of cubes are placed in a task, the cobot's decision-making terminates (see \textit{Global Success/Fail} terminal states that are fully observable in A-POMDPs in Figure~\ref{fig:apomdp}). The task results are then shown on the monitor, which are the success rate, task duration and a score. A score is to motivate the participants to achieve the goals during an experiment, which is detailed later.

\begin{figure}[!htbp]
	\centering
	\includegraphics[width=0.8\columnwidth]{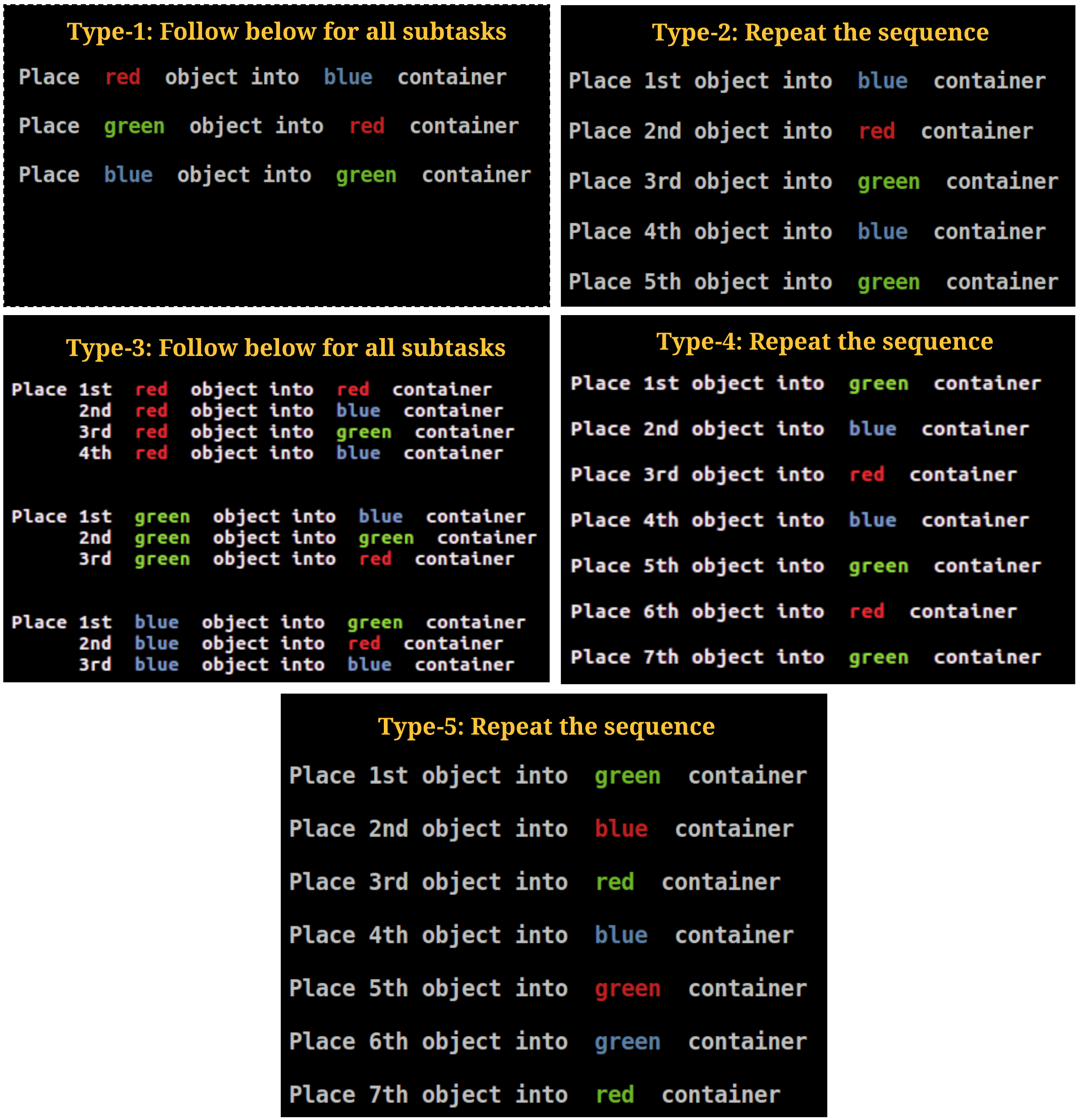}
	\caption[Task rules]{Displaying task rules for 5 different task types. The cubes flow on the conveyor with a random order, unknown to the participants.}
	\label{fig:taskRules}
\end{figure}

\paragraph{Cognitively challenging task design:} A task in our experiments allow for a fluent collaboration, where a human and a cobot should take the initiative to change and adapt to task allocations on-the-fly. In simulation, we model a task to be physically demanding, and a cobot would track a human's physical abilities and conditions. In user studies with real people, such a setup would be difficult to prepare in a lab environment. For this purpose, we still focus on a task of pick and place but for the simplicity we create a cognitive load for the humans instead of a physical one. A cognitive load replaces physical exhaustion and incapability by introducing cognitively demanding memory and coordination exercises. We argue that human states like tiredness, distraction, lost motivation etc. can also result from a difficult cognitive task, which is an easier and a safer option in a lab environment.

A larger number of placement rules to remember yields a more difficult task since they are only displayed to a participant for a short amount of time. Therefore, a task is cognitively demanding to varying extents according to the complexity of the rules and the amount of time they are displayed. In total, 5 different task types have been implemented (shown in Figure~\ref{fig:taskRules}). Each type is inspired from mind games to sort colors with confusing rules, e.g., using Stroop effect \citep{stroop_studies_1935}.
Our goal is to select tasks that are both challenging but with a difficulty still achievable by the people, on average, when they receive no help from a cobot. In this way we ensure a collaboration. Hence, we also evaluate and compare the cognitive loads these tasks induce on people during the calibration experiments (in \textit{step 8} of the pipeline), which is summarized in Section~\ref{ssec:eval_CogLoad} and detailed in the supplementary materials. In addition to the color rules and their limited display time, after each cube placement (subtask), the participants are informed to wait for audible feedback before they move to the next one. In the scenario, this corresponds to a supervisor check. The timing of the sound is random (usually around 2 seconds); hence, it introduces another cognitive challenge to remembering the rules.

\paragraph{How to induce the unanticipated human behaviours?}

\begin{table*}[!htbp]
\centering
\caption{Our experiment design choices to invoke unanticipated human behaviors.} \label{tbl:mapExpToBehavs}
\resizebox{\textwidth}{!}{
\begin{tabular}{|p{.15\textwidth}|p{.77\textwidth}|}

\hline \multicolumn{1}{|c|}{\textbf{Human Behaviors}} & \multicolumn{1}{c|}{\textbf{Invoked through}} \\ \hline

Failures                 & cognitively demanding task rules, rules being visible to a participant for seconds, and audible feedback before moving to the next subtask. \\
\hline
Distraction						    & the \textit{task monitor} that participants sometimes check to see the current status of a task and do not pay attention to the task itself. \\
\hline
Tiredness                           & the cognitive load that accumulates through multiple tasks and a long experiment that takes approximately 1,5 hours. \\
\hline
Motivation to work          & scoring system that drives competitive behaviors and the training phase in which the participants practice tasks and the collaboration with the cobot before the experiments. \\
\hline
Lost motivation					    & the same task type repeated several times, the difficulty of a task, and the length of an experiment that all require a constant attention and memory. \\
\hline
Willingness to collaborate with the cobot  & the trust in the cobot's success in a task, the scoring system, and the difficulty of an experiment. \\
\hline
Not wanting the cobot to assist (non-collaborative behaviors)  	& decrease in trust due to unwanted behaviors of the cobot (e.g. misinterpreting a human's assistance need, when the cobot makes a mistake.), that the tasks are initially assigned to the participants where they get more rewards if the person achieves a task. The participants can also warn the cobot to stop, indicating competitive behaviours in humans. \\
\hline
\end{tabular}
}
\end{table*}

The collaboration between a human and the cobot is motivated by a scoring system that punishes wrong placements with negative rewards and assigns positive rewards for any correctly sorted objects (subtasks), whether it is by a participant or by the cobot. As the tasks are designed to require high cognitive load from the participants, such a rewarding system is expected to incentivize them to accept help from the cobot, especially when they are struggling with a task. We also have initial task assignments to be able to better evaluate the cobot's assessments over the human's progress when, e.g., the task is assigned to her. In our scenarios, more rewards are received when a subtask is achieved by whomever the task is assigned to. This is to favor the assignee to do the job. The same rewarding mechanism is also used in the cobot decision models, i.e., A-POMDP (in Section~\ref{ssec:apomdp}). Hence, it is a shared goal of the team to maximize the collaborative score. Since the color of the cubes can be misidentified due to the changing lighting conditions and other classification errors, the cobot may also make a mistake. This is informed to the participants without the success rates of the cobot, which is above $95\%$, to also evaluate trust. After a task ends, the final score is displayed to the participant to allow her to draw conclusions from it and create strategies for the next task. For instance, she may decide to trust the cobot more, which we categorize as the long-term changing human characteristics.

One goal of our cobot is to adapt to a human's changing willingness to collaborate. Therefore, we initially assign all of the subtasks to the participants, and the cobot is there to assist if needed. Our intention is to reduce the situations where a participant lets the cobot permanently take over a task and to create a more dynamic collaboration by giving her the autonomy to decide if the cobot should assist or not. For example, if a human is not sure about a subtask, she could let the cobot take over, thus avoiding a negative reward while sacrificing the larger reward. All of this is told to the participants before the experiments with an analogy from a real factory that an assigned task is better achieved by the assignee for an increased quality. To describe this analogy better, we give an example that the objects to be placed are fragile and so a human would handle the placement better than a cobot. In order to maximize the collaborative reward, we expect this to encourage the participants to take over as many of the placement tasks as possible. This helps us observe more human dynamics during an experiment. Finally, the goal of this environment setup and the scenario is to invoke the unanticipated human conditions in a work environment, such as tiring, losing motivation, failures, etc. that are expected to be observed during monotonous working conditions. In Table \ref{tbl:mapExpToBehavs}, we summarize all of our strategies to induce such conditions (referring to Figure~\ref{fig:realsetup}). The actual effect of these strategies depends on a participant's characteristics (e.g., memory, stubbornness, competitiveness), the time of the day in which the experiment is conducted and a participant's background (e.g., field of study, previous experience with robots). This gives us more variety and less control, which is a better test environment for evaluating adaptation.

\subsubsection{Real-Time Human Interaction}
%%%%%%%%%%%%%%%%  ROBOT COGNITION  %%%%%%%%%%%%%%%%%%
\label{ssec:expcognition}

This section details our activities on \textit{step (7)} of the pipeline (see Figure~\ref{fig:pipeline}) to successfully deploy the \textit{sensing and actuating} components of FABRIC in Figure~\ref{fig:framework}. In order to allow for a real-time human interaction, the cobot needs to constantly perceive its environment and the human actions. For this purpose, we gather visual information about the scene using 2 cameras. A human worker interfaces the system through a yellow glove and a safety helmet (as shown in Figure~\ref{fig:realsetup}). From an RGBD camera, we recognize hand gestures and the objects to interpret human actions and interactions with the objects. The other camera on the wall tracks human attention through the head gestures recognized. In addition, there are load sensors located under the container trays that detect the objects the participants put onto them. 
%Using the information of the currently grasped object and the container it is placed on, the system checks whether the human placed an object on the correct tray or not, indicating the success of a subtask. 
The human activities and environment state are processed under the \textit{sensing} component of our framework (in Figure~\ref{fig:framework} to generate the observation vector for the cobot's decision-making.

\paragraph{Observation Vector}
\label{ssec:obsVector}

\begin{table*}[!htbp]
	\caption{Observation vector with the descriptions and how they are derived}
	\centering
	\includegraphics[width=0.8\textwidth]{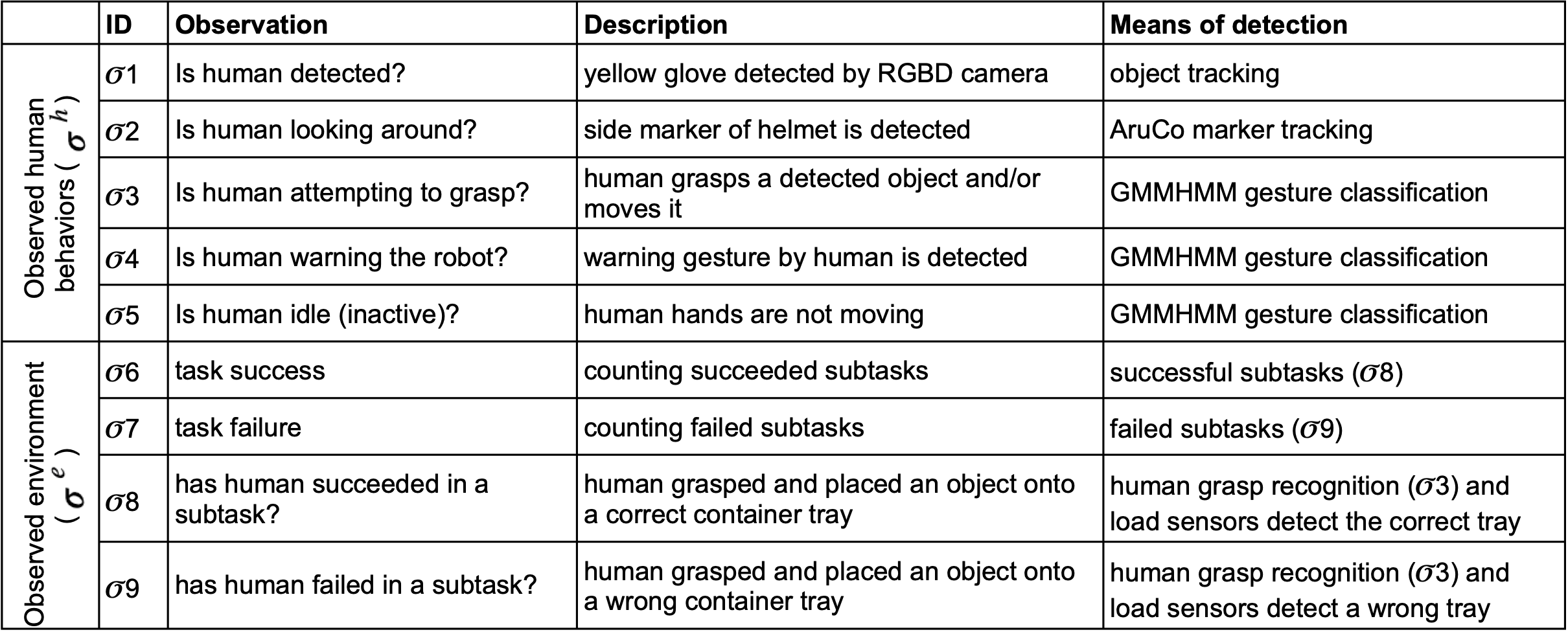}
	\label{tbl:observationVector}
\end{table*}

The observation vector is the feature vector for the cobot decision-making, i.e., observables for A-POMDP models during a task and collectively used to estimate the interaction type in ABPS before a new task (in Section~\ref{ssec:decisional_level}). We define the vector with nine different observables as shown in Figure~\ref{fig:apomdp}. In Table \ref{tbl:observationVector}, we list these observables and how they are generated by the \textit{sensing} component of FABRIC. Among the observables, there are direct observations, such as, a human is detected ($\mathlarger{\mathlarger{\sigma}}1$) when a hand glove is visible for the HAR system, the human attempts to grasp an object of interest ($\mathlarger{\mathlarger{\sigma}}3$), the human warns the cobot to stop ($\mathlarger{\mathlarger{\sigma}}4$), and the human stays idle ($\mathlarger{\mathlarger{\sigma}}5$). 

\begin{figure}[!htbp]
	\centering
	\includegraphics[width=0.8\columnwidth]{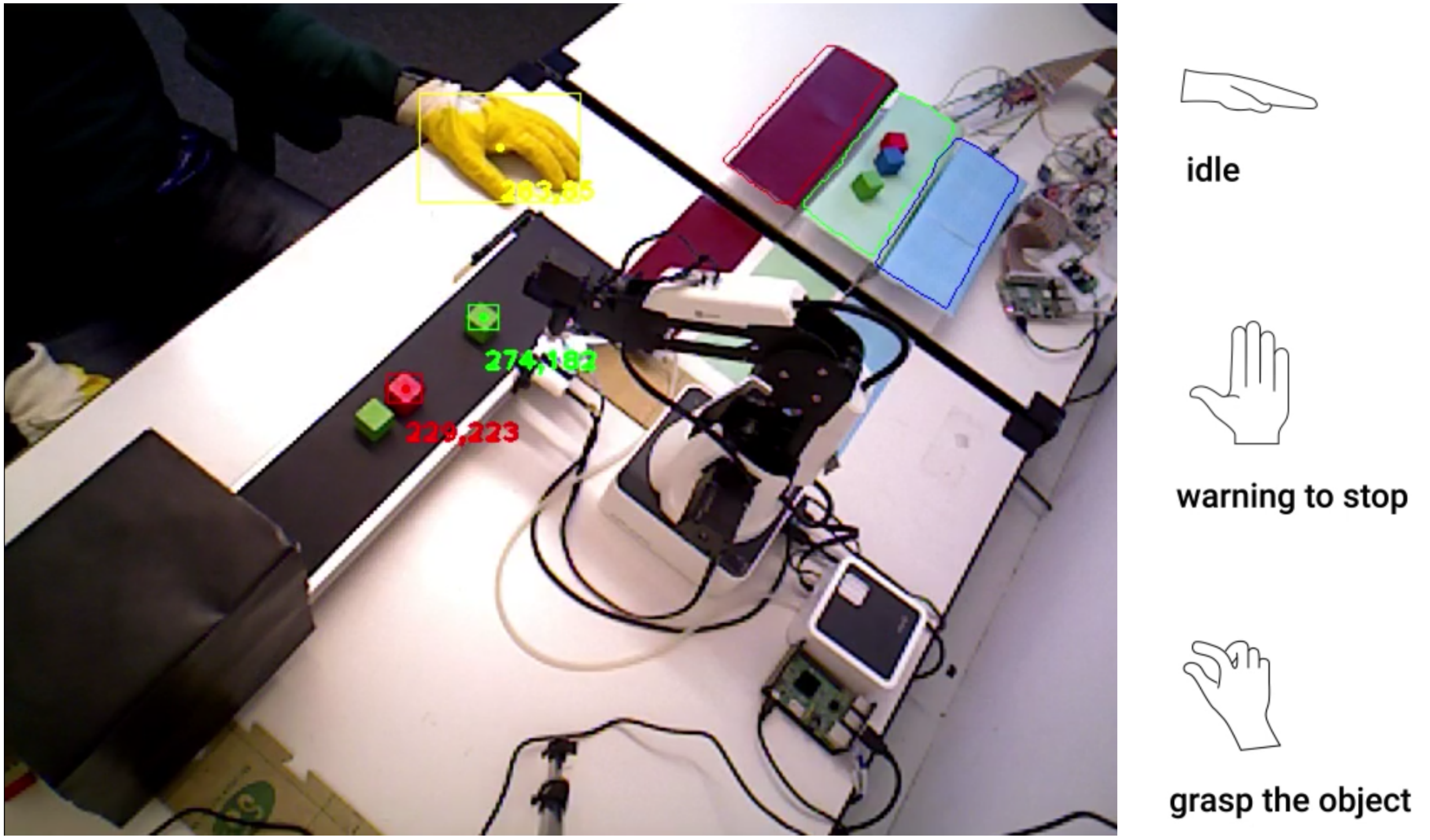}
	\caption{The cubes, the containers, and the glove are detected in the work space even when the cubes are grasped.}
	\label{fig:harRGBD}
\end{figure}

The last three, $\mathlarger{\mathlarger{\sigma}}3$, $\mathlarger{\mathlarger{\sigma}}4$, and $\mathlarger{\mathlarger{\sigma}}5$, are obtained by a state-of-the-art HAR module that recognize the constrained and distinct human gestures from the movements of the glove (see Figure~\ref{fig:harRGBD}).
Moving from the 3D tracked location of the glove, the hand's velocity, and the spatial relation between the tracked objects and the hand, human actions of \textit{idling (inactive), attempting to grasp}, and \textit{warning sign} are recognized by a Gaussian Mixture Model (GMM) Hidden Markov Model (HMM) classifier (inspired by \citep{Li2018, Roitberg2014}). An additional class of ``undefined action'' is also added with some random irrelevant gestures in order to avoid false positives on the three important gestures. A GMM operation creates discrete clusters of these continuous variables that are then classified by HMM. We create a separate GMM-HMM for every action of interest. Then, every time a feature vector arrives, all models generate a confidence value and the one with the highest confidence is taken as the current most probable action. After training the system with various people, we reached over 90\% accuracy with $~10Hz (fps)$ on a continuous video stream.

The observable $\mathlarger{\mathlarger{\sigma}}2$, looking around action, is detected by tracking the markers on the safety helmet (in Figure~\ref{fig:realsetup}). The posture of the participant's head is used to detect if the person is looking around but not gazing somewhere in the work area. For this purpose, we ran several tests during the calibration experiments to define margins of tracked markers to interpret if a human is looking around. The accuracy of this observable is very high; however, a misdetection is compensated through the need to sequentially observe this gesture to estimate the state of a human has lost her attention in our A-POMDP (in Figure~\ref{fig:apomdp}). Additionally, there are some observables that are derived from the others. According to the assigned task rules, we check the status of all of the load sensors under the container trays and the grasped object to output a subtask success or failure ($\mathlarger{\mathlarger{\sigma}}8$ and $\mathlarger{\mathlarger{\sigma}}9$). Finally, a task success and failure are calculated by counting the amount of successful and failed subtasks. When the total number reaches a value defined by a task definition (mostly 10 in our experiments), either $\mathlarger{\mathlarger{\sigma}}6$ or $\mathlarger{\mathlarger{\sigma}}7$ becomes true so that a cobot decision-making model (i.e., A-POMDP) terminates. In our experiments, the task status flags are just for the cobot models to terminate. They are not used to measure the final task performance, which is calculated using the subtask results and who achieved it.

\begin{figure}[!htbp]
	\centering
	\includegraphics[width=1\columnwidth]{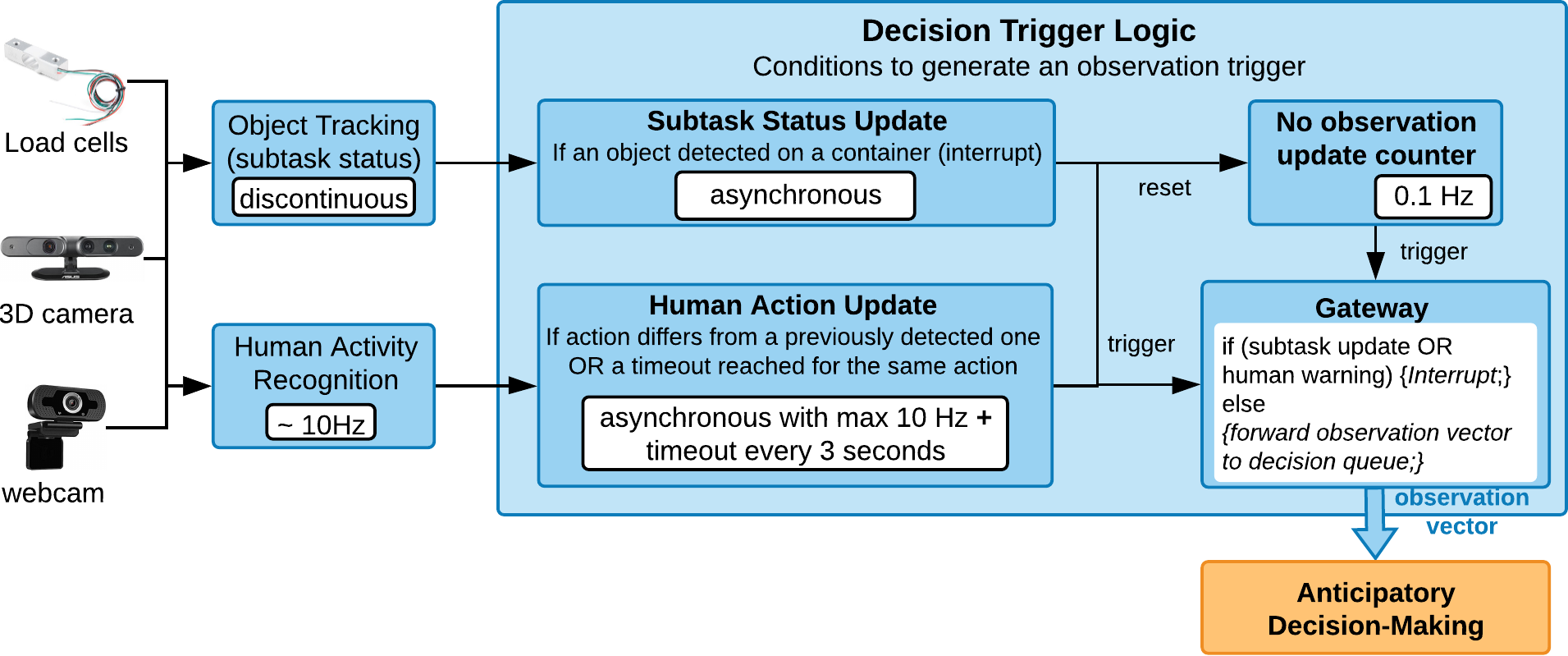}
	\caption[Decision trigger logic]{Decision trigger logic of our framework in Figure~\ref{fig:framework}.}
	\label{fig:decisionTrigger}
\end{figure}

\paragraph{Decision Trigger Logic}
\label{ssec:obsUpdate}
The stochastic nature of human behaviors makes it difficult for a robot to follow the decision-making with a constant update frequency, especially when the scenario does not follow turn-taking collaboration. For this reason, our cobot should balance timely decision-making with natural interaction speeds, catching human actions and the important changes in the environment to respond reliably. For fluent coordination, our cobot implements a decision trigger logic as represented in Figure~\ref{fig:decisionTrigger}. In our setup, a cobot decision can be triggered in three different conditions: 1) when a container update is recognized (a subtask is completed), 2) when a change in human actions is recognized, 3) when a timeout is reached in case of no change in human actions or a long-time absence of an observation update. The \textit{decision trigger logic} component of FABRIC handles these conditions to synchronize the sensory inputs to create the observation vector, analyze it in case a decision should be triggered, and forward it to the \textit{anticipatory decision-making} component accordingly (at the \textit{line~7} of Algorithm~\ref{algo:abps}). Starting with the discontinuous events, as shown in the figure, both a subtask update (i.e., when a cube is detected on a container) and a new human action can happen at any time. We generate interrupts for an immediate decision-making whenever a cube is detected on a container, and when a warning action is recognized. If a decision of cancelling an ongoing action is made from the warning gesture, this also interrupts the existing cobot action under the \textit{actuating} component of FABRIC.

For all of the other cases, we catch and respond to every change in recognized human actions. A currently recognized human action is compared with the previously detected one and a decision is triggered whenever they differ. Nevertheless, a new decision is triggered regardless of this comparison after a timeout of 3 seconds. Through the calibration experiments with the participants, we have calculated the time it takes for a human to start and finish an action to be approximately 3 seconds, taking into account the actions of pick and place, warning the robot, and idling (in the supplementary materials). Therefore, an update frequency of $10~Hz$ would be unnecessarily fast for the cobot decision-making and would even be unreliable as the cobot may think a human is taking the same action repeatedly even if it is still one action in progress. Even though we use an online POMDP solver with a fast real-time response, generating a decision is the limiting factor for the update frequency of the cobot. Our experiments have shown that generating an optimal decision can take up to around 1 second on an average PC, whereas we may generate and forward a new observation vector with a frequency of about $10~Hz$. For that, we queue the observation vectors, except for an interrupt case, not to overload and crash the \textit{decision-making} component with many requests (see the gateway in Figure~\ref{fig:decisionTrigger}). A new observation from the queue is forwarded as soon as a currently running decision process ends. Thanks to the multi-agent architecture of FABRIC, a decision-making process runs in parallel with an ongoing cobot action being executed.

\subsection{Calibration Experiments}
\label{ssec:eval_calib}

This section summarizes our activities under \textit{step (8)} of the pipeline in Figure~\ref{fig:pipeline}. Since the experiments conducted here are not the main focus of the paper, their details are provided in the supplementary materials. Calibration experiments involve an initial case study with real humans to validate and fine-tune our novel interaction setup and the cobot system. This experiment acts as prototyping tests on the setup to check and analyze if our setup and the cobot's collaboration is reliable, and whether the setup can induce sufficient cognitive load on the participants to observe unanticipated human behaviors. By doing so, we can better evaluate the adaptation goals of our anticipatory decision-making approaches and the contributions of our framework to the fluency of an HRC. We also utilize the results of these experiments to fine-tune the decision trigger logic, real-time human collaboration capability, and the cobot's reliability in its responses (i.e., the decision-making models).

The first round of experiments evaluated whether the cobot effectively communicates its decisions and internal states to its human partner, i.e., robot expressiveness. We invited 12 people and showed them the cobot actions defined in our A-POMDP design (in Figure~\ref{fig:apomdp}). Especially the \textit{planning} and \textit{pointing to remind} actions are needed to effectively communicate the cobot's intention as the planning is a precursor for the human to be aware that the cobot may take over the task soon, and the pointing action is to remind the human about the task. In the end, we experimented with different motion designs and chose the most expressive ones, which were correctly understood by the majority of the participants who have never interacted with a robot before.

Our collaboration setup aims to place a cognitive load on the humans and the degree of cognitive load may differ between the task types in Figure~\ref{fig:taskRules}. Our goal is to compare the types and choose suitable ones for the final experiments. Our criteria are: 1) The task is cognitively demanding enough to observe unanticipated human behaviors, 2) the task is easy and motivating enough to keep the human engaged in the collaboration. We let 8 participants and the cobot collaborate over an extended period (approximately 1.5 hours each) executing our tasks in Figure~\ref{fig:taskRules}. We compare objective and subjective measures from each task to find out how well the human-cobot team achieved the tasks, the levels of cognitive load induced on the participants (through NASA-TLX metrics \citep{hart_development_1988}), how frequently they exerted unanticipated behaviors, such as, lost attention and failures. More details on the experiments, the metrics used, and the results are given in the supplementary materials.

In summary, we conclude that the setup is effective in invoking unanticipated human behaviors. This is clear in task type-4 (in Figure~\ref{fig:taskRules}) through the participant responses on not remembering the rules, their lost attention over the course of the experiment, and from very high NASA-TLX scores. 
These analyses show that the higher the cognitive load is, the more unanticipated human behaviors, followed by more human errors. Since it has led to fluent collaboration with a significant amount of invoked unanticipated behaviors, we choose task type-4 as the main collaboration task to evaluate our cobot's short-term adaptation capabilities in Section~\ref{ssec:eval_shortTerm}. For the evaluation of long-term adaptation in Section~\ref{ssec:eval_longTerm}, we select to work with task type-5, which has the same configuration as type-4 along with an additional Stroop effect (in Figure~\ref{fig:taskRules}e). As mentioned in Section~\ref{ssec:exp_task}, the Stroop effect was very difficult to master for this experiment that repeated the task only 3 times for each participant. However, it has proven to be very suitable for a longer collaboration as it leads to a noticeable change in human characteristics, such as expertise through the learning effect.

\subsection{Training For Real Anticipatory Decision-Making}
\label{ssec:training}
In this section, we first detail how we improve our framework with the feedback from the calibration experiments then talk about our training process, following \textit{step (9)} of the pipeline in Figure~\ref{fig:pipeline}. The parameters of the A-POMDP models have been first tuned in simulation, in various collaboration scenarios. However, a transition to the real-world interaction requires a further adaptation of the models to a real human's decision-making frequency. We are aware that there cannot be one robot decision model optimized for all types of people and interactions. For this purpose, as we discuss in Section~\ref{ssec:abps}, our initial goal is to obtain an A-POMDP model, the ``base model'', that works reasonably in many interaction scenarios and obtains high rewards. This model is then used to generate several other models for ABPS to select the most suitable one during a specific interaction, as we did for the simulation experiments in \citep{Gorur2019}, toward a personalized human collaboration.

For the base model, we first manually tweaked the probabilities until the new cobot policy reaches an average high reward against many different interactions at the lab. Then, we collect real observations from the calibration experiments (in Section~\ref{ssec:eval_calib}) and use this data to further train the model's intrinsic parameters. In particular, from an observed sequence of $s_0, a_0, s_1, a_1, ..., s_n, a_n, s_{n+1}$ we update the transition probabilities, $T(s'|s,a)$, where $s \in S$ and $a \in A$ are the A-POMDP states and actions defined in the model (in Figure~\ref{fig:apomdp}). The current state information, i.e., $s$, is subjectively obtained from the participants if it is a human mental state, e.g., the human is tired. Similarly, the sequence of $s_0, a_0, \sigma_1, s_1, a_1, \sigma_2, ... , s_n, a_n, \sigma_n, s_{n+1}$ is used to update the observation probabilities, $O(\sigma|s',a)$, where $\sigma \in O$ is the observation vector the cobot receives from the environment after its action. 
For the purpose of systematically observing and training for distinct human behaviors, most common interaction cases are identified and examined. We also examine the cobot's reactions in the test runs to filter out the unreliable cases. We list below some of these common interaction scenarios we iteratively tested with different participants and collected the observations at different execution speeds and intervals.

\begin{itemize}
	\item Human continuously grasps: Human starts in the idle position, grasps the object and places it into a container to grasp the other object right afterward.
	\item Human idles for long: People idle for changing duration of time as a reflection of their states like evaluating the task or being tired. The cobot may take over or give more time to the human depending on the individual.
	\item Human warns the robot: The cobot should receive and process the warning on time no matter what the cobot was doing and in whatever way it decides to respond to this warning.
	\item Human looks around, not attending to the setup for a while: The cobot should estimate that the human may have lost her attention and take an action.
\end{itemize}

One of the qualitative findings was regarding the duration between multiple subtasks, i.e., the transition times. For example, sequentially recognized idling actions between two subtasks may be interpreted as ``the human is tired'' if the robot transitions to that state quickly, whereas in reality the human may be waiting for a new object to arrive on the conveyor. As a result, the state transition probabilities are updated to favor the probability of a state persisting rather than transiting to another one. In the training, we use the subject data to improve the model parameters as mentioned above, which were previously trained on the simulation. The final base model turned out to be a reliable model against various humans. However, we note again that there cannot be a single probability distribution defined for a robot in its interactions with multiple people. The base model is used as the ``proactive model'', our A-POMDP that handles unanticipated human behaviors, in our short-term adaptation experiments to compare it with a reactive model that discards such behaviors (see Section~\ref{ssec:eval_shortTerm}).

For the long-term experiments (in Section~\ref{ssec:eval_longTerm}, we generate a policy library, being the \textit{decision model library} of FABRIC. For that, we randomly adjust the transition and observation probabilities of the base A-POMDP model to generate various models, each of which handles a unique human type, and solve for their optimal policies to construct our policy library $\Pi$. This is to limit the arbitrary generation of the robot policies to avoid overloading the library with unreliable candidates. Then, our ABPS mechanism runs on top of the library to select a policy for an estimated human type (see Algorithm~\ref{algo:abps}). We first train ABPS for the observation and the performance models (in Definition~\ref{def:abps_obs} and \ref{def:abps_perf}, respectively) against simulated humans as in \citep{Gorur2019}. In the end, 20 policies were selected for their use in the experiments based on: 1) how well they performed overall against many different randomly generated human models after discarding the worst ones, 2) how distinct their performance models are from the other policies by grouping the similar ones. Some policies ignore a human's warnings and try to complete a task, whereas some pay more attention to a human's needs, taking the human as the leader of the collaboration. The trade-off between these two is clearer when it comes to non-collaborative human types. There are also some policies that prefer to encourage the human to complete the task, e.g. by pointing out to remind the human when distracted instead of directly taking over the task. Which policy is optimal depends on the interacted human type and the task definition.

We are agnostic to the exact type labels of humans in our experiments. As mentioned, we assume each human the robot is interacting with has an unknown type to the robot, which can only be estimated as a distribution over the known types. For that, we have crafted 16 different (known) human types using the modeling scheme in Figure~\ref{fig:human_model} with the goal of each generating as distinct set of human actions as possible in simulation. That means creating human types with the extremes of the four characteristics of our concern, namely the levels of \textit{expertise, stamina, attention}, and \textit{collaborativeness}. Our assumption is that an unknown human type can be approximated as a probability distribution over these extreme types. We note that since each of the 16 human models are stochastic, they still generate a diversity of behaviors after random sampling (see Section~\ref{ssec:humanSimu}). The collaboration of each of the 20 robot policies and each of the 16 human types is repeated for 90 sequential tasks in our simulation environment. In total, we accomplished 28800 interactions (28800 task instances and 288000 subtask instances), which is very difficult to manage in real-world scenarios.

Finally, the human type estimation of ABPS, the observation model, from the simulation experiments is also updated with the real human observations collected from the calibration and the short-term adaptation experiments, where in both ABPS, \textit{the adaptive policy selection} component of FABRIC, is not used. The labeling of the human types is done from the objective and subjective measures collected from the participants, such as their success rates, the warning amounts, exhaustion and attention levels, and their average idling times. The updated observation model provides a more reliable estimation thanks to the real observations and can accurately estimate a wider diversity of the human types than the interacted ones in our calibration experiments thanks to the simulated data (in Section~\ref{ssec:eval_longTerm}). Our goal is now to show that the framework is applicable in the real-world for its real-time autonomous human collaboration and to prove that it covers our extended human adaptation goals leading to a more efficient and natural collaboration.

\section{Evaluation and Results}
\label{sec:evaluation}
This section evaluates our framework and its extended human adaptation capabilities covering both our short- and long-term collaboration goals (i.e., the \textit{steps (10), (11) and (12)} of the pipeline).
In Section~\ref{ssec:eval_shortTerm}, our goal is to validate our findings in Section~\ref{ssec:apomdp}, which is the short-term adaptation capability of our robot handling unexpected human conditions. We repeat the same experiment done in simulation, comparing our A-POMDP robot decision model design with a conventional reactive model in G\"or\"ur et al.~\citeyearpar{Gorur2018}, this in a real-world setting. Then, in Section~\ref{ssec:eval_longTerm}, we integrate the full anticipatory decision-making system on the setup to validate the applicability of both our ABPS mechanism from Section~\ref{ssec:abps} and our overall framework introduced in Section~\ref{sec:framework}. All of the experiments use a within-subject design and we invited different people to each of the experiments to be able to independently evaluate different aspects without any practice effect or prior system knowledge.

\subsection{Evaluation of Short-Term Adaptation}
\label{ssec:eval_shortTerm}

With this experiment, our goal is to examine the hypotheses below:

\begin{hypo}\label{hypo_2}
	A cobot's fluent collaboration with a human contributes to increased performance in a cognitively challenging task when compared to a human working alone.
\end{hypo}
\begin{hypo}\label{hypo_3}
	Our A-POMDP model adapting to a human's unanticipated behaviors (extended short-term adaptation) contributes to more efficient and natural collaboration when compared to a cobot model that does not handle such behaviors.
\end{hypo}
\begin{hypo}\label{hypo_4}
	Our A-POMDP model adapting to a human's unanticipated behaviors (short-term adaptation) shows better adaptation skills and it has a higher perceived collaboration, trust, and positive teammate traits, than a cobot model that does not handle such behaviors.
\end{hypo}

We let the participants interact autonomously with two different cobot planners. The first one is a \textit{proactive robot} that runs our A-POMDP model in Figure~\ref{fig:apomdp}. This cobot first anticipates a human's characteristic, e.g., lost attention, incapability, or tiredness, and then it estimates if the human needs assistance or not (extended short-term adaptation, in Section~\ref{ssec:apomdp}). On the contrary, the other cobot, the \textit{reactive robot}, does not handle the unanticipated behaviors of a human. It treats a human's need for help as a directly observable (deterministic) state. The reactive robot deterministically decides that a human needs help when (i) a certain time duration has passed without a cube placement (i.e., without a subtask completion), (ii) the human is not detected around the workplace, or (iii) the human fails in a subtask. We design the \textit{reactive robot} by removing the \textit{anticipation stage-1} of our A-POMDP model design in Figure~\ref{fig:apomdp}. With the \textit{anticipation stage-2} being deterministic, this model is designed as an MDP. Through this comparison, our intention is to show the importance of handling the unanticipated human behaviors (i.e., stochastic interpretation of such human states) for an improved short-term adaptation.

\subsubsection{Objective and Subjective Measures}
\label{ssec:exp1_measures}
\hfill \break
This section details our activities under \textit{step~10} of the pipeline in Figure~\ref{fig:pipeline}. Our goal is to measure the efficiency and naturalness of the collaboration and the adaptation skills, perceived collaboration, trust, and positive teammate traits of the cobot. In a work environment, in general, efficiency of a task is defined by the quality and the duration of the work done. Hence, work division is crucial for the efficient use of resources that takes into account the availability and the matching skill sets of the workers. For example, in a pick and place task, varying fragility of the products may require a work division, where a human should handle the fragile objects while the cobot should work on bulky and heavy ones. Thus, we consider the highest efficiency when \textit{a task is successfully accomplished by its assigned collaborator}. In our case, the tasks are initially assigned to the people as we focus on the cobot's human anticipation skills and its correct interpretation of a need for assistance. This is motivated to the participants through our scoring system (see Section~\ref{ssec:exp_task}) and is also known to the cobot by initializing A-POMDP models from the state of ``Task is Assigned to Human'' (in Figure~\ref{fig:apomdp}).

The term naturalness defines a natural interaction between the cobot and a human where they achieve a fluent collaboration. In particular, the level of naturalness is defined by the fluency of their communication that is often expected to be nonverbal in a collaboration \citep{hoffman_evaluating_2019}. A good indicator of the naturalness of collaboration is the level of intrusive behaviors from the cobot, which leads to frustration in the human partner. A cobot should reliably understand the collaborated human's needs and preferences to adapt to a situation and to avoid intrusive behaviors. In our setup, a warning gesture is provided to the human so that she communicates her displeasure with a cobot behavior. Therefore, the number of warnings hints at the naturalness of a collaboration. To conclude, our quantitative measures are the following:

\begin{itemize}
	\item \textbf{Rewards gathered:} We use the same robot reward mechanism as in Section~\ref{ssec:apomdp} for both of the reactive and the proactive robot, i.e., punishing each warning received from a human and a subtask failure and rewarding each subtask success.
	\item \textbf{Number of warnings:} The number of warning gestures a participant made during a task.
	\item \textbf{Task success rate ($S_{task}$):} The overall success rate of a task, calculated by $S_{task} = \frac{n_{s}}{n_{total}}$, where $n_{s}$ is the total amount of successful subtasks and $n_{total}$ is the total amount of subtasks.
	%is the total number of successfully completed subtasks in a task divided by the total number of subtasks.
	\item \textbf{Human success rate ($S_{human}$):} The rate of the successful placements of a human collaborator out of all of her attempts, calculated by $S_{human} = \frac{n_{s_{human}}}{n_{s_{human}} + n_{f_{human}}}$, where $n_{s_{human}}$ and $n_{f_{human}}$ are the amount of successful and failed subtasks by the human in a task, respectively.
	\item \textbf{Human contribution in a task ($C_{human}$):} The overall successful contribution of the human to a task is calculated by $C_{human} = \frac{n_{s_{human}}}{n_{total}}$
%as in Equation \ref{eq:userstudies_humanContr}.
	\item \textbf{Task efficiency ($\eta_{task}$):} The task efficiency depends on the overall task success and how much the assignee (human partner) has contributed. It is calculated by, $\eta_{task} = S_{task} \cdot C_{human}$.
\end{itemize}

We subjectively evaluate the fluency of the collaboration from the perspective of the participants. The subjective measures are collected by means of questionnaire responses, where the participants describe their agreement with a statement through a 5-step Likert scale. The statements are given in Table~\ref{tbl:exp1_subjStatements}. The category of questions concerning the cobot's effect on the cognitive load is to see if the cobot is perceived as a negative or positive influence on such a challenging task. The remainder of the statements are for measuring the metrics of \textit{perceived naturalness, reliability, trust, positive teammate traits,} and \textit{perceived collaboration of the cobots} that are mostly inspired by the relevant HRC research, in \citep{hoffman_evaluating_2019, Nikolaidis2013, Koppula2016, Nikolaidis2017_ijrr}. 
Even though we categorize the statements for simplicity, they are interchangeably evaluated under the relevant hypotheses. There are also similar statements that are shuffled in the questionnaires to check the consistency of a participant's ratings. The statements categorized as ``general'' are rated only twice for each participant, one is after the proactive robot experiment and the other is after the reactive one. The comparison statements are only asked once at the very end of the experiment to let the participants compare the two cobots. The other statements are all task-specific and asked after each task completion (see Figure~\ref{fig:exp1_procedure} for the experiment protocol).

\begin{table}[!htbp]
	\caption{Subjective statements asked during the short-term adaptation experiments}
	\centering
	\includegraphics[width=\columnwidth]{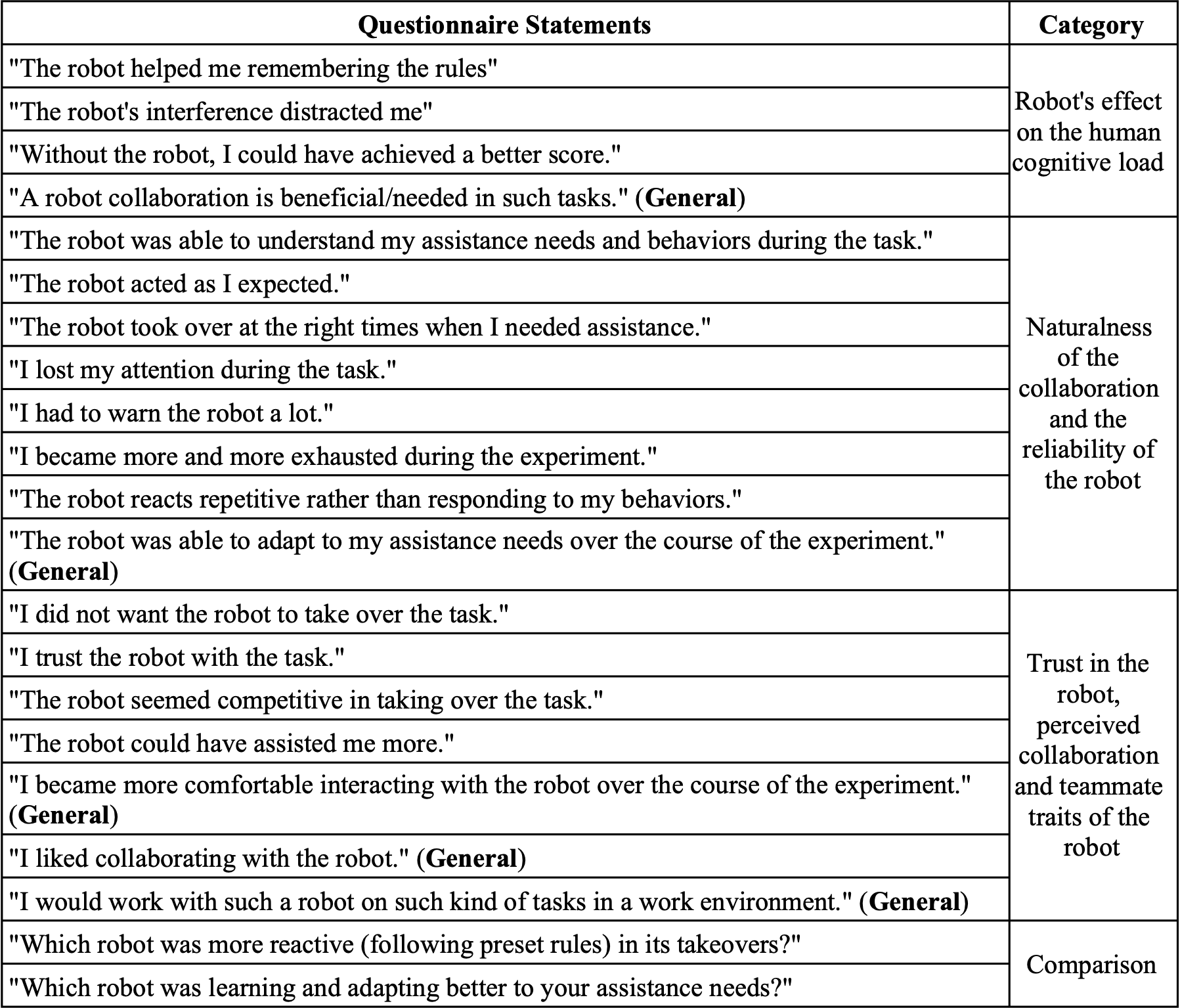}
	\label{tbl:exp1_subjStatements}
\end{table}

\subsubsection{Experiment Protocol}
\label{ssec:exp1_protocol}
\begin{figure}[!htbp]
	\centering
	\includegraphics[width=1\columnwidth]{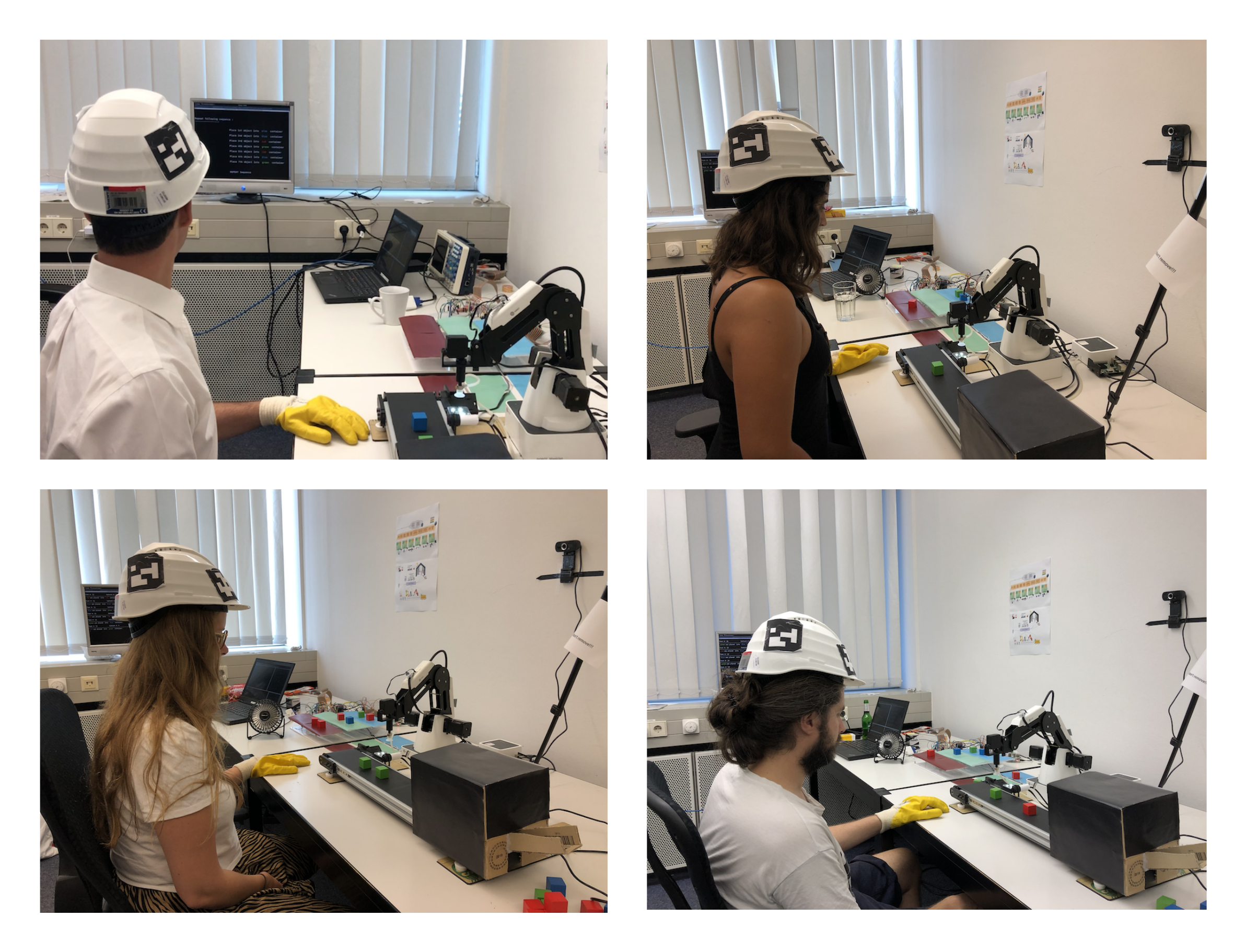}
	\caption[Participants during the short-term adaptation experiments]{Participants are collaborating with the cobot.}
	\label{fig:exp1_participants}
\end{figure}

\hfill \break
We invited 14 people (the ages range from 17 to 38) from different ethnic groups and from various backgrounds (computer science, social sciences, law, chemical engineering, tourism, public affairs, and business school) and let them interact with the two cobot models. We considered age and ethnicity as potential covariates but found no significant effect on the results. We designed a within-subject experiment to compare the two cobots. Thus, each of the participants interacted with both of the robot types. We note that the participants knew that there were two types of cobots and that they needed to evaluate them both. We only notified a participant when we switched the robot type; hence, they only knew them as robot type-1 and type-2 for anonymity. Once a participant takes the chair, the experiment starts (see Figure~\ref{fig:exp1_participants}).

\begin{figure}[!htbp]
	\centering
	\includegraphics[width=\columnwidth]{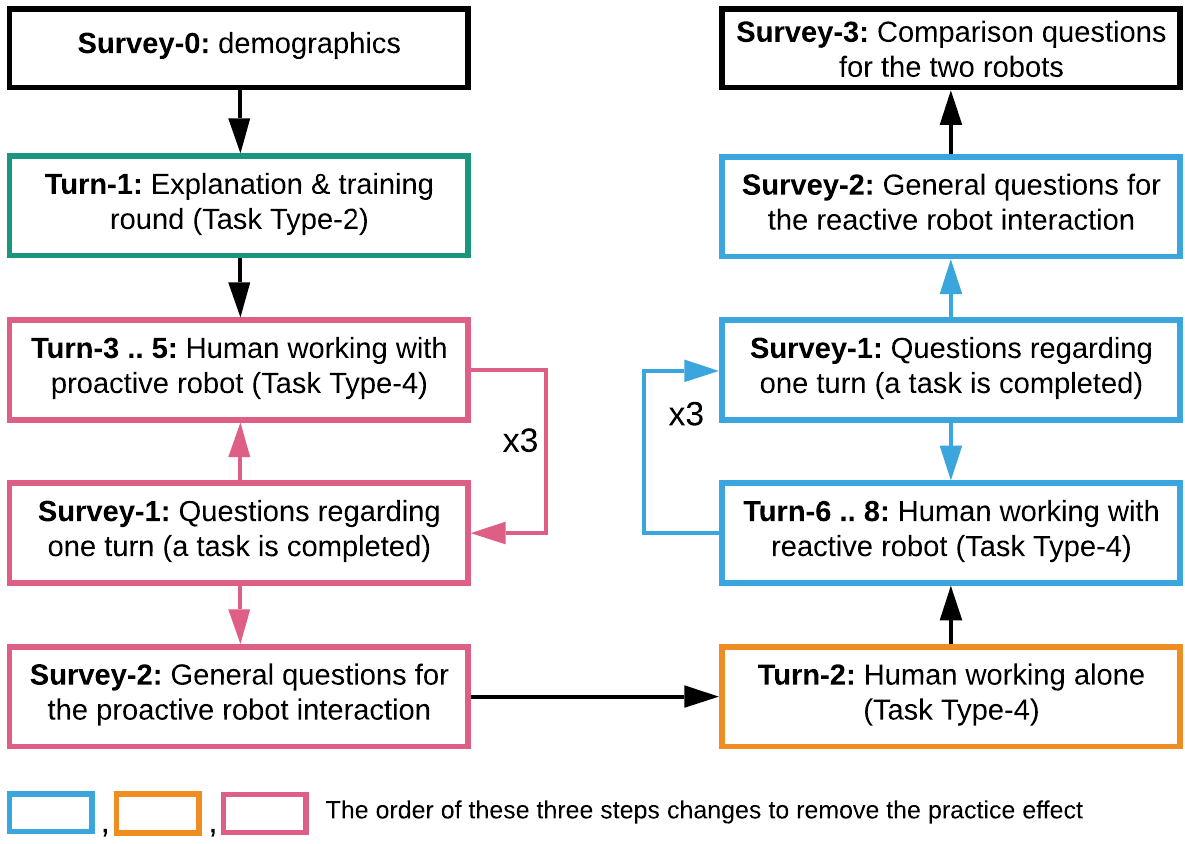}
	\caption[Experiment procedure - Experiment 1]{The protocol of the experiments for each participant}
	\label{fig:exp1_procedure}
\end{figure}

After the calibration experiments, we chose task type-4 as the most suitable task for collaboration (see Figure~\ref{fig:taskRules}). We kept the task type the same throughout the whole experiment to remove any effect from changing task difficulties. The experiment procedure is depicted in Figure~\ref{fig:exp1_procedure}. First, we do a training round with each participant on a simpler task to avoid practice-effect. The operator describes how to complete a task successfully (e.g., how to grasp and place the cube, how to warn the robot) and motivates the participant that the tasks are always assigned to them and for each subtask they achieve the team receives a higher score than the cobot completing it (see the scoring system in Section~\ref{ssec:exp_task}). The operator also mentions that there is no time limitation and they have to wait for audio feedback after each placement before moving to the new one. Then, the experiment starts.
In general, there are three main steps we follow during the experiments: 1) The participant completes one task alone without any robot interaction, 2) the participant collaborates with the reactive robot for 3 tasks, 3) the participant collaborates with the proactive robot for 3 tasks (in Figure~\ref{fig:exp1_procedure}). In total, each participant completes 7 tasks. In order to remove any practice-effect over different robot types, 7 randomly selected participants interacted first with the reactive robot and the other 7 with the proactive one. After each task, a participant needs to fill out a task-specific survey. At the end of each robot type interaction, the participant fills out the general survey questions (in Table \ref{tbl:exp1_subjStatements}). In total, an experiment with a participant lasts approximately 1,5 hours.

\subsubsection{Results and Discussions}
\label{ssec:exp1_results}
\begin{figure*}[!htbp]
	%\hspace{0.01\columnwidth}
	\begin{minipage}[c]{.5\textwidth}
		\centering
		\subfloat[Overall task success rates \label{fig:exp1_successRate}]{{\includegraphics[width=0.47\textwidth]{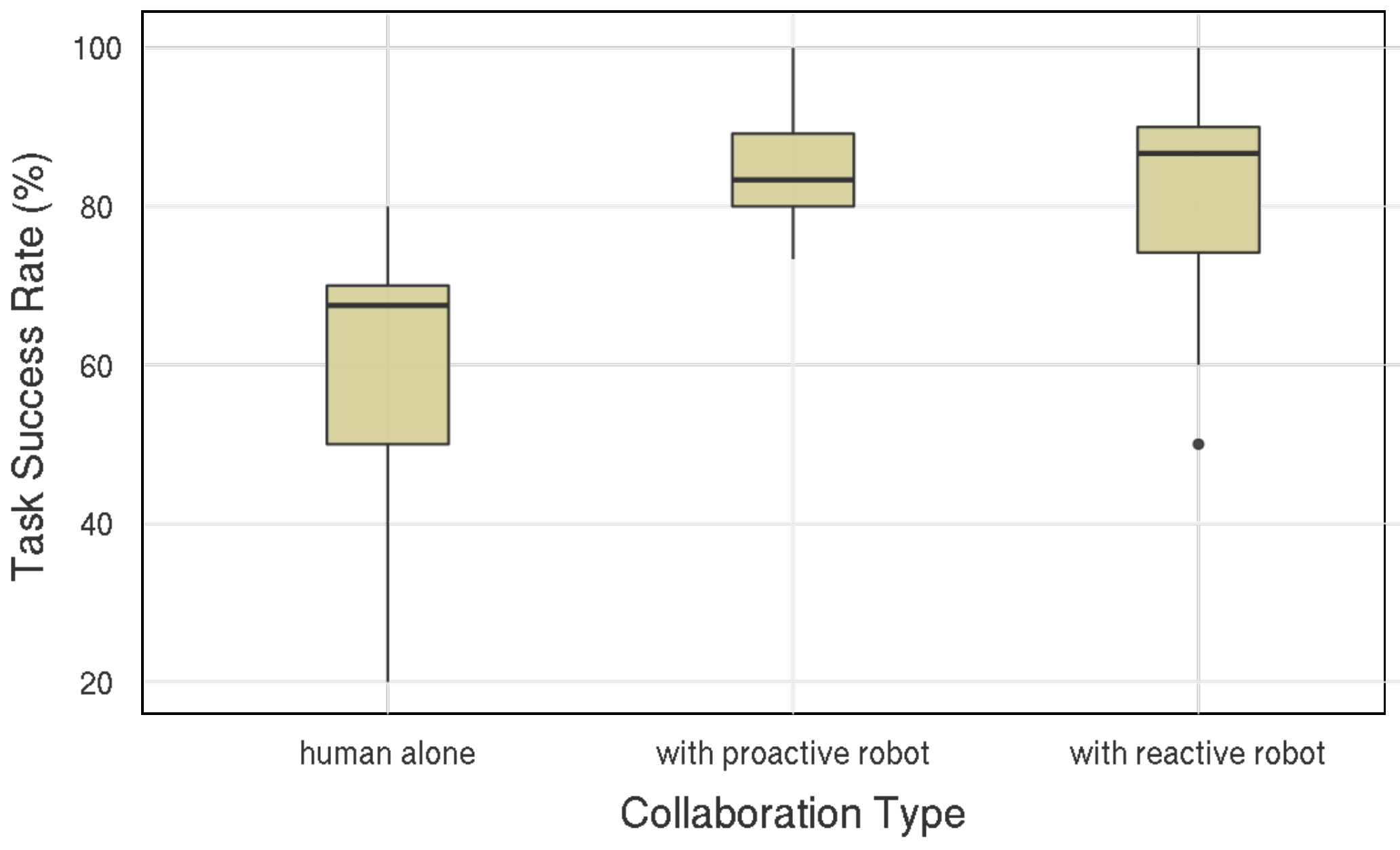} }}
	%\end{minipage}
	%\begin{minipage}[c]{0.4\columnwidth}
		\hspace{0.02\textwidth}
		\subfloat[Human success rate in a task \label{fig:exp1_humanSuccess}]{{\includegraphics[width=0.47\textwidth]{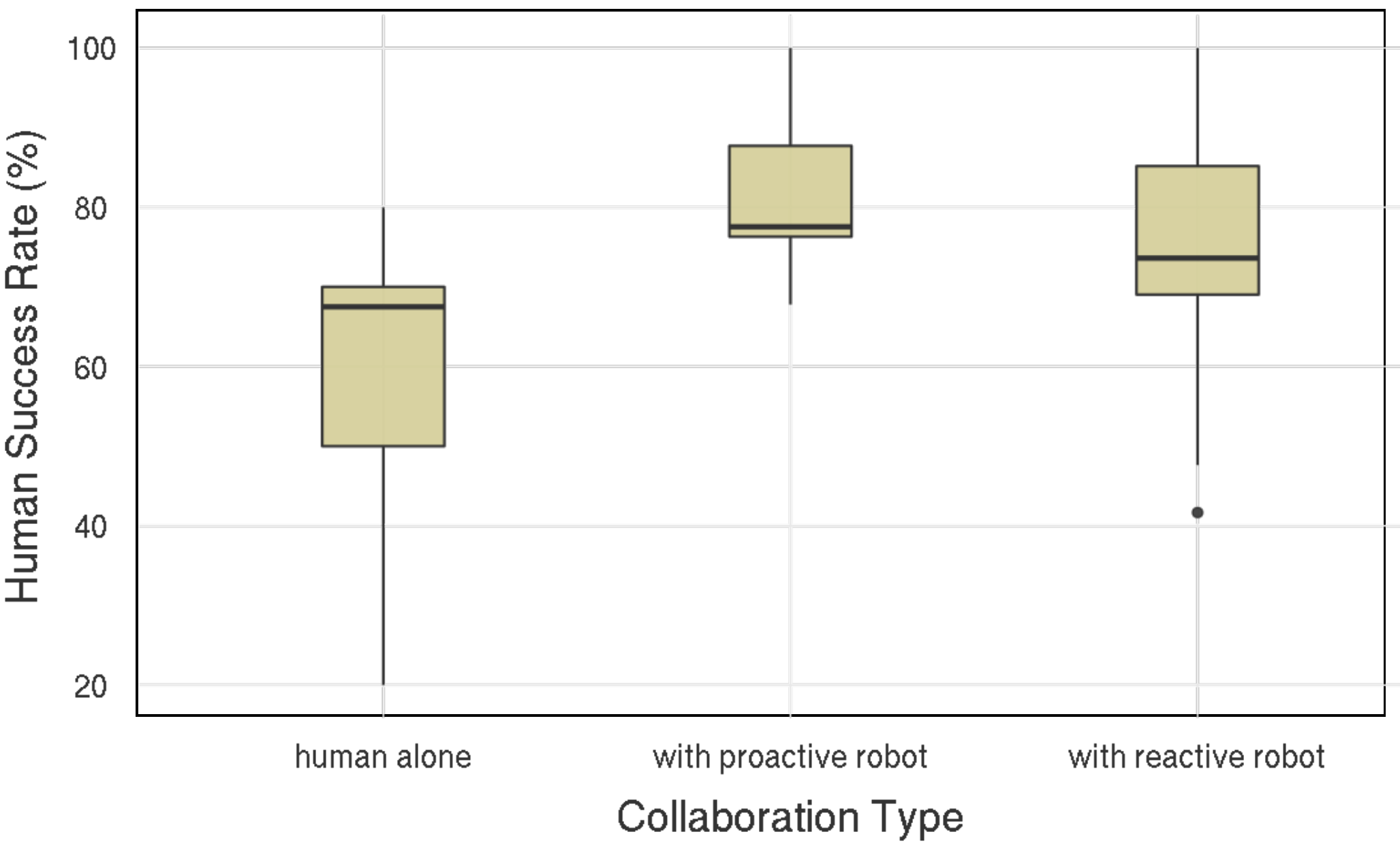} }}
	%\end{minipage}
	%\begin{minipage}[c]{\columnwidth}
		\hfill\allowbreak
		\centering
		\subfloat[Human's successful contribution to the task \label{fig:exp1_humanContr}]{{\includegraphics[width=0.47\textwidth]{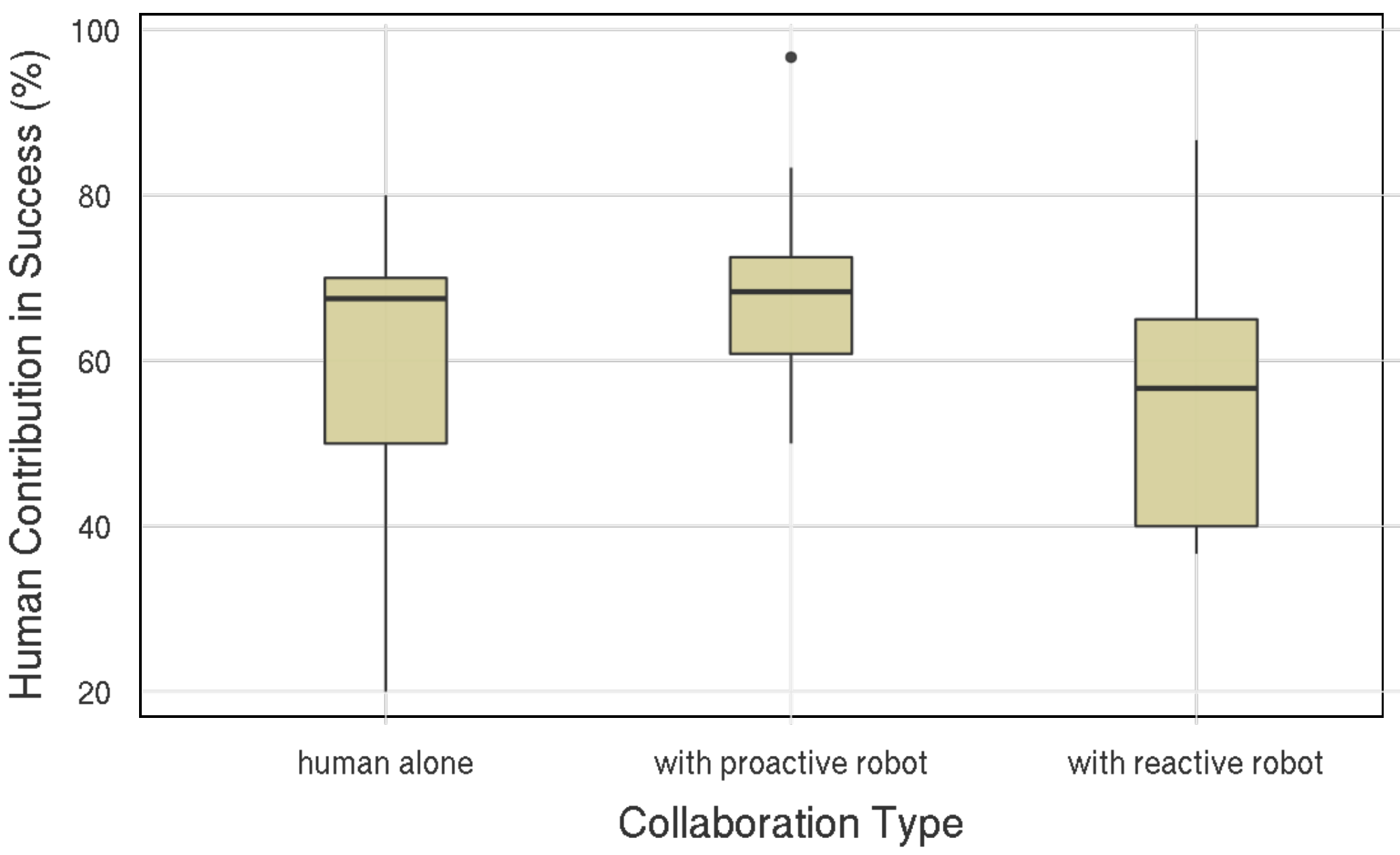} }}%
	%\end{minipage}
	%\begin{minipage}[c]{0.4\columnwidth}
		%\qquad
		\hspace{0.02\textwidth}
		\subfloat[Task Efficiencies \label{fig:exp1_efficiency}]{{\includegraphics[width=0.47\textwidth]{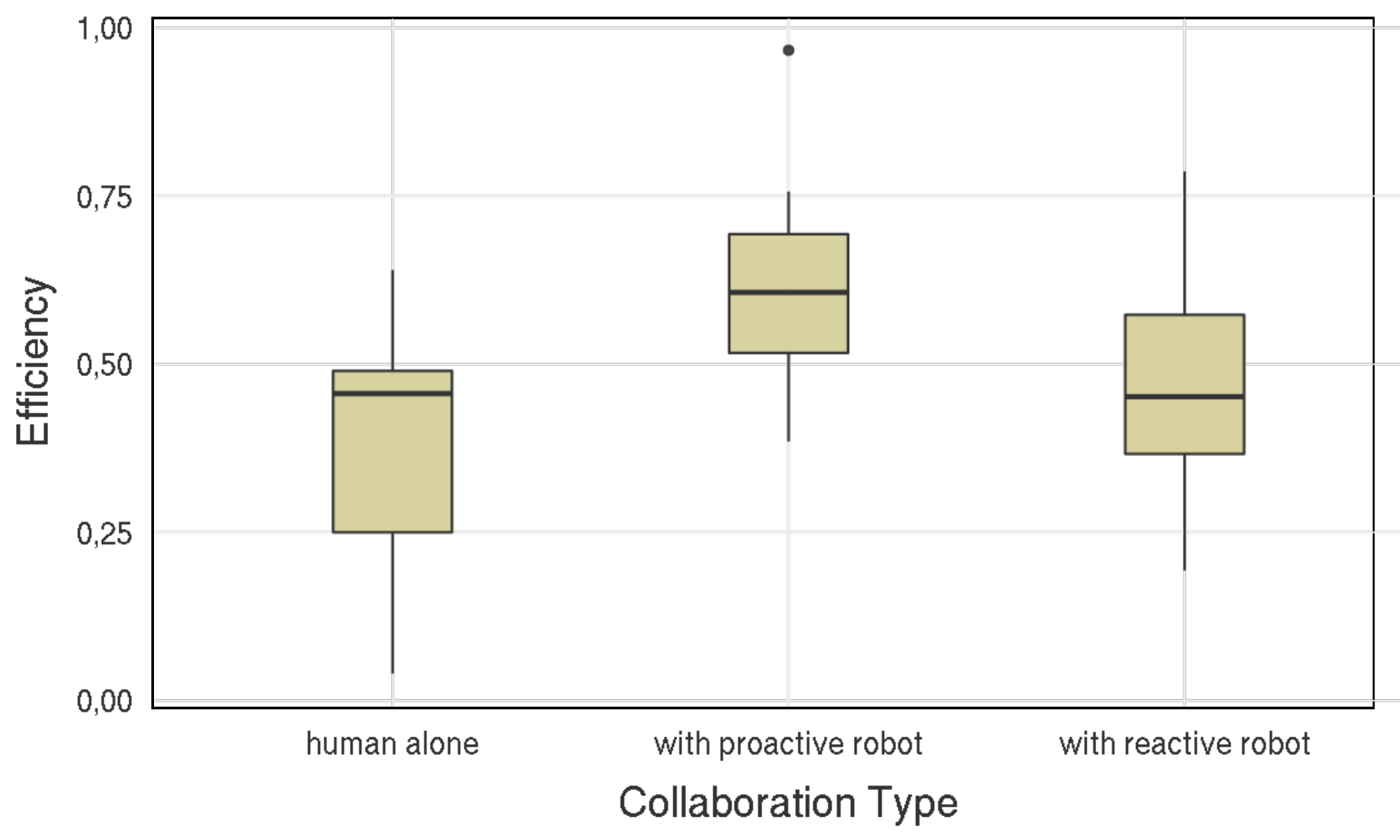} }}
	\end{minipage}%
	\begin{minipage}[c]{.5\textwidth}
		\centering
		\subfloat[Mean values and the ANOVA results. \label{fig:exp1_successTable}]{{\includegraphics[width=0.97\textwidth]{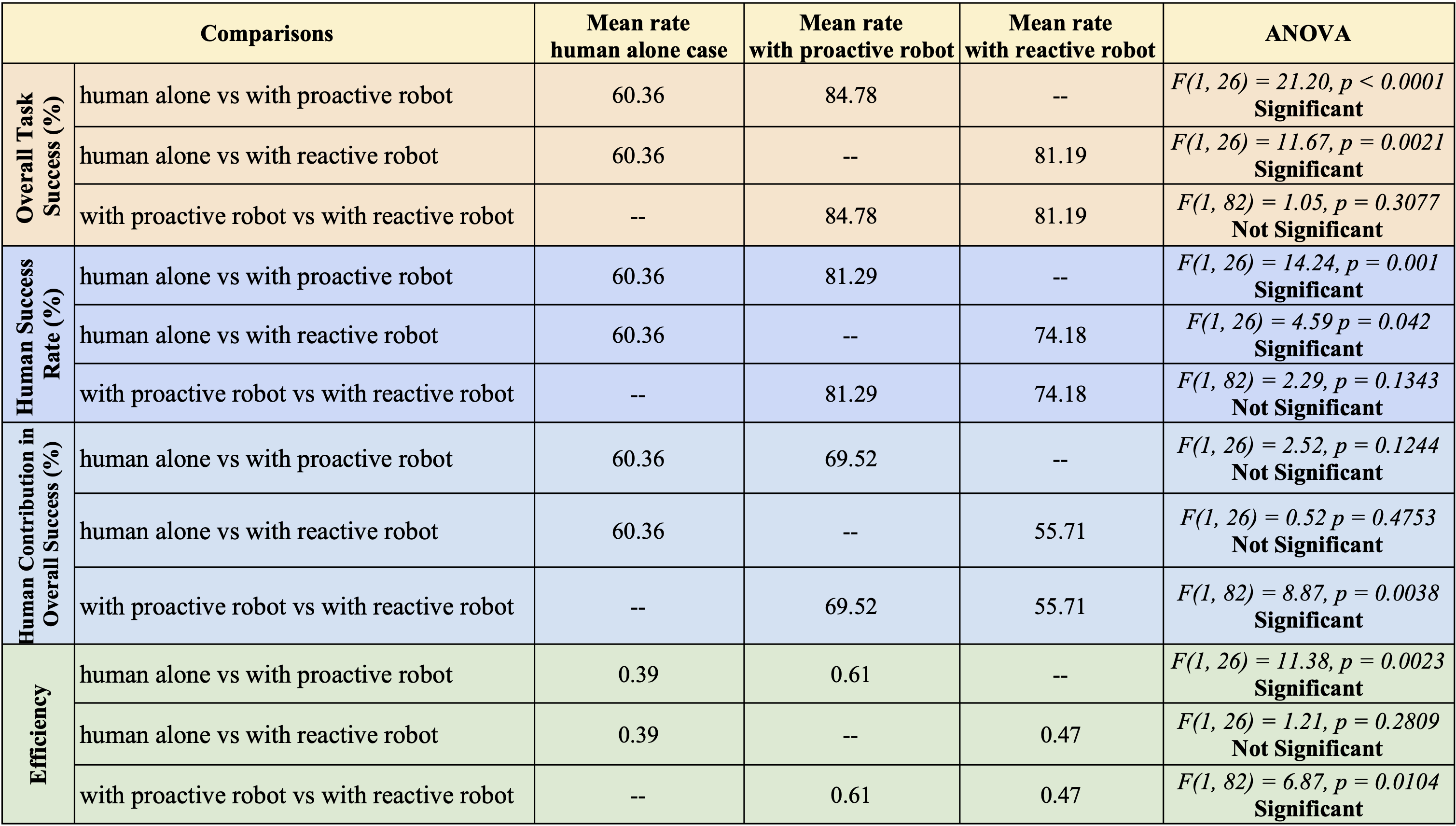} }}
	\end{minipage}
	\caption{(a)-(d) Box and whisker plots of the overall task success rates, a human's success rate, a human's successful contribution to an overall task, and the efficiency values averaged over 14 participants during their performance alone and their collaboration with the proactive robot and the reactive robot. (e) The ANOVA results of the plotted objective measures to compare the performance of the two cobots.}
	\label{fig:exp1_successPlots}
\end{figure*}

\hfill \break
In Figure~\ref{fig:exp1_successPlots}, we give the box and whisker plots of the success and efficiency analysis and in Figure~\ref{fig:exp1_successTable} are the numerical values and the ANOVA results. As seen in Figure~\ref{fig:exp1_successRate}, the task success rate when the participants worked without a cobot is significantly worse than a collaboration with any of the cobot models ($p < 0.05$). A collaboration with a proactive robot contributes positively to a task success by $~41\%$ and with the reactive robot, it is increased by $~35\%$ compared to a human working alone. Similarly, a human's success rate has also significantly increased when the human collaborated with either of the cobots (in Figure~\ref{fig:exp1_humanSuccess} with $p < 0.05$ for both cases), whereas with the proactive one this increase is slightly higher (by $~35\%$ increase on the human success rate).

In Table~\ref{tbl:exp1_subjective_results}, we give the Likert scale (1 to 5 with increasing agreement) results of the participant's statements throughout this experiment. Table~\ref{tbl:exp1_subjective_results}a shows the statements for analyzing a cobot’s impact on such a challenging task.
The participants state that both of the cobots helped them remember the task rules and that a robot collaboration is beneficial in such tasks (see high mean rates in the table). Finally, the average task efficiencies are shown in Figure~\ref{fig:exp1_efficiency}. Both the reactive and the proactive robot contributed significantly positively to the task efficiency ($p < 0.05$), compared to a human working alone (an increase of $~56\%$ for the proactive robot and $21\%$ for the reactive one to the overall task success rates in Figure~\ref{fig:exp1_successTable}).
%the efficiency (in Equation \ref{eq:userstudies_efficiency}).
%to point out the robot's positive contribution (see the mean values of the mentioned statements in Table \ref{tbl:userstudies_exp1_subRobotEffectsTable}.
Thereby, we underscore the importance of such cobot collaborating with humans in challenging tasks. The success rate analysis, the efficiency results, and the subjective ratings of the participants support \textbf{Hypothesis~\ref{hypo_2}}.

Even though the proactive robot on average achieved higher success rates than the reactive one (see Figure~\ref{fig:exp1_successTable}), there is no significant difference between them. The decision models only decide to the level of taking over a subtask. After that, both cobots perform achieve success in placing the cubes (over $95\%$). Therefore, some participants were comfortable leaving the task to the cobot once they realize its capability. For the reactive robot, this happened significantly more than the proactive one due to its deterministic rules. Still, the proactive robot provides a more stable success than the reactive one by keeping the variance low (see Figure~\ref{fig:exp1_successRate}). In both cases, our main concern is how much of this success actually comes from the human. Figure~\ref{fig:exp1_humanContr} shows that the proactive robot significantly increased the human's contribution to success when compared to the reactive robot ($p = 0.038$). The reactive robot also respects the initial task assignment; however, it favors taking over a task when, for example, a human idles too long, discarding the unanticipated human behaviors and preferences. This led to a decrease in a human's successful contribution in a task during her collaboration with the reactive robot when compared to her performance alone (as shown in Figure~\ref{fig:exp1_successTable}). Finally, since a higher efficiency is achieved when a task is successfully accomplished by its assigned collaborator, the proactive robot significantly increased the task efficiency of a human working alone by $~57\%$ ($p=0.0023$) and ruled out the task efficiency achieved with the reactive robot (with $p=0.0104$ in Figure~\ref{fig:exp1_successTable}).

\begin{table}[!t]
		\caption{The results of the subjective statements asked during the short-term adaptation experiments. The participants state their agreement over a 5-step Likert scale and the results are averaged over 14 participants. The ANOVA results compare the performance of the proactive and the reactive robot.}
		\centering
		\includegraphics[width=\columnwidth]{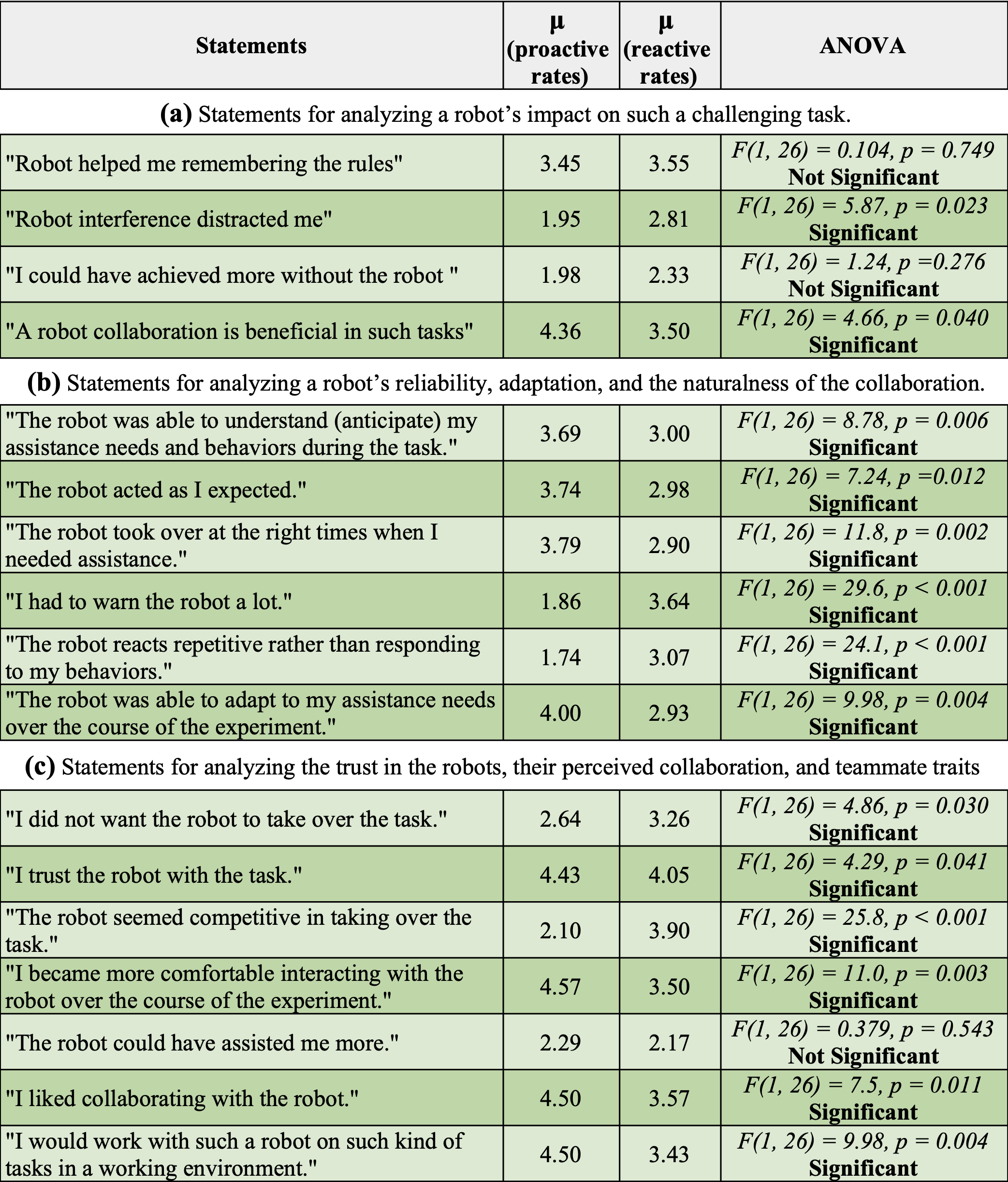}
		\label{tbl:exp1_subjective_results}
\end{table}

\begin{figure}[!t]
	%\hspace{0.01\columnwidth}
	\begin{minipage}[c]{\columnwidth}
		\centering
		\subfloat[Number of warnings received \label{fig:exp1_warnings}]{{\includegraphics[width=0.47\columnwidth]{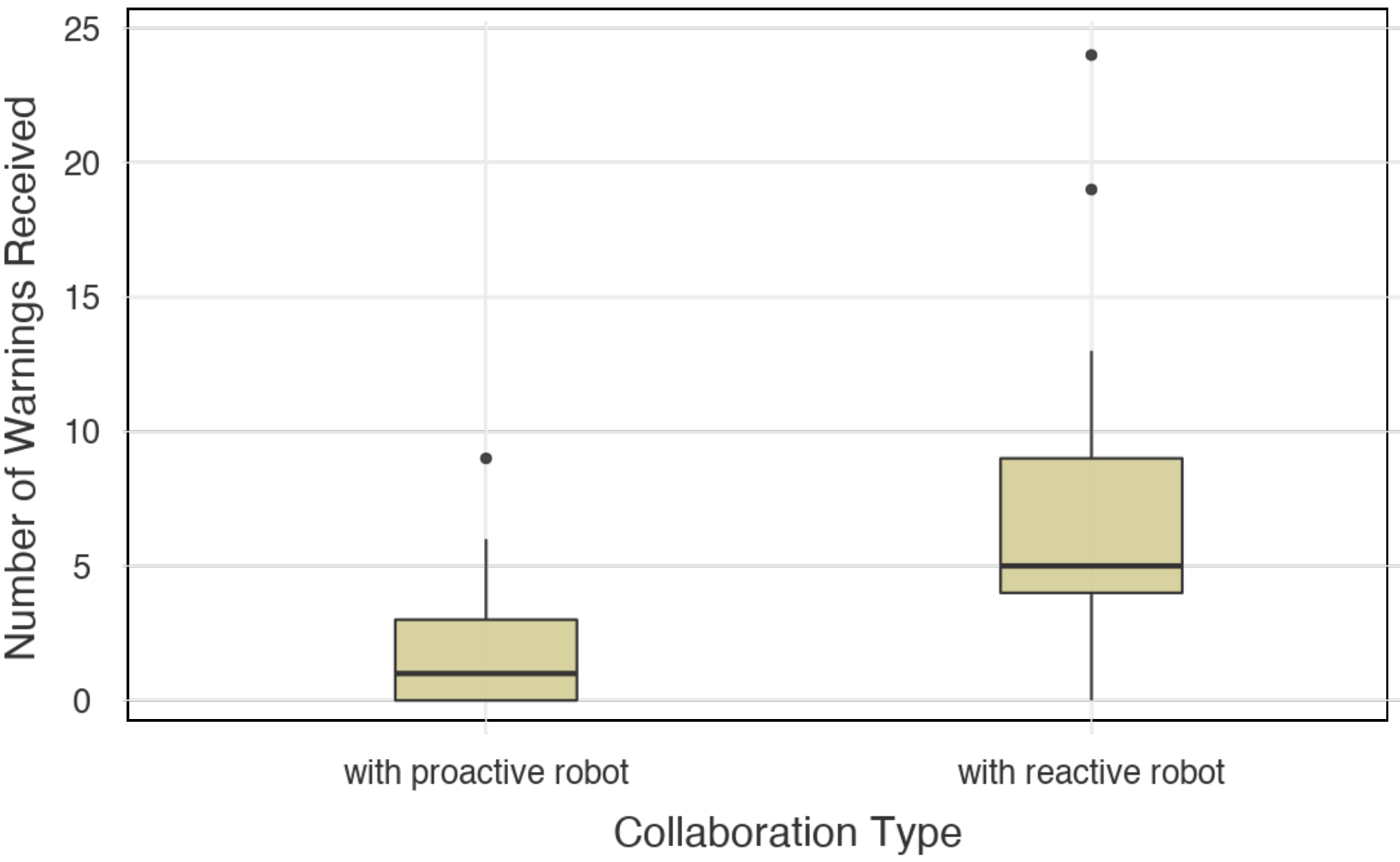} }}
		%\end{minipage}
		%\begin{minipage}[c]{0.4\columnwidth}
		\hspace{0.02\columnwidth}
		\centering
		\subfloat[Rewards gathered by the robots \label{fig:exp1_rewards}]{{\includegraphics[width=0.47\columnwidth]{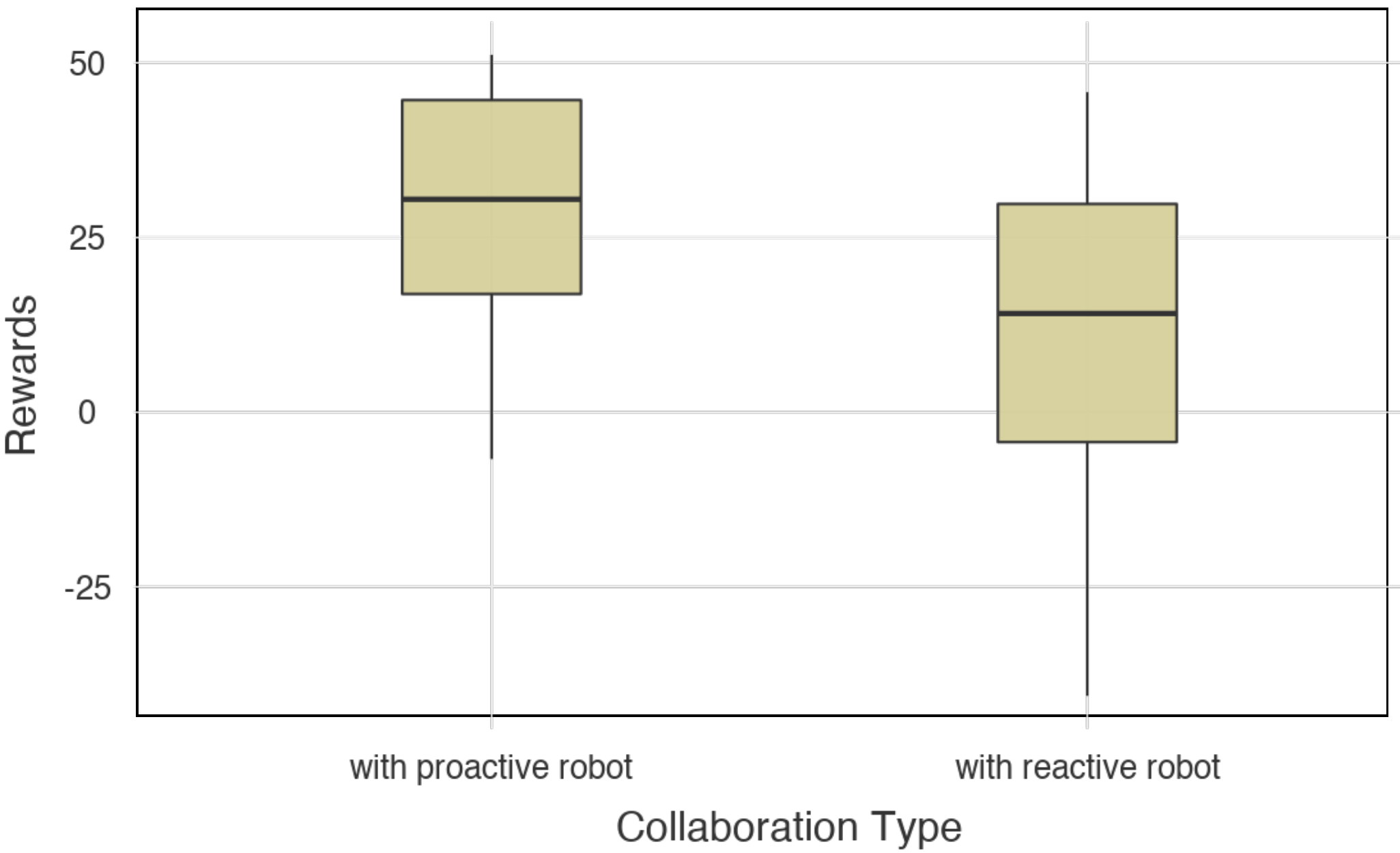} }}%
	\end{minipage}
	\begin{minipage}[c]{\columnwidth}
		\centering
		\subfloat[Mean values and the ANOVA results \label{fig:exp1_warnRewardsTable}]{{\includegraphics[width=\columnwidth]{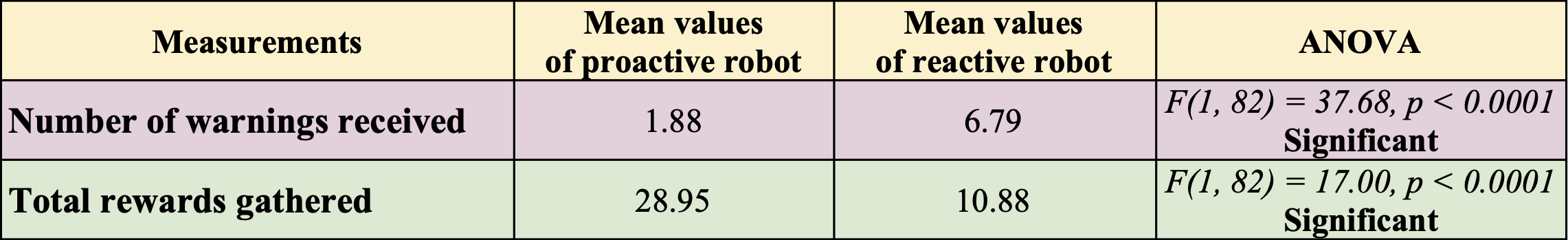} }}%
	\end{minipage}
	\caption{(a), (b) Box and whisker plots of the number of warnings the proactive robot and the reactive robot have received from a human and the rewards they gather during a task, averaged over 14 participants on 42 tasks for each of the cobots in total. (c) Mean values of the two cobots and the ANOVA results for the comparison of their performance.}
	\label{fig:exp1_warningsRewards}
\end{figure}

As discussed in Section~\ref{ssec:exp1_measures}, the naturalness reflects a fluent communication in that the handovers and turn-taking need to be interpreted correctly by both of the collaborators. Figure~\ref{fig:exp1_warnings} suggests that the proactive robot could keep the warnings close to zero, suggesting a higher accuracy in estimating a human's unanticipated behaviors and need for help. In Figure~\ref{fig:exp1_warnRewardsTable}, the ANOVA results point out the significantly higher number of warnings the reactive robot received, which is $3.6$ times of the proactive robot's with $p < 0.0001$. The participants also evaluated if a collaboration with a cobot felt comparatively natural to them, i.e., more human-like. The participants stated that the reactive robot's interference distracted them significantly more than the proactive case (with $p=0.023$ in Table~\ref{tbl:exp1_subjective_results}a). This is largely due to the significantly increased unexpected interferences from the reactive robot (with $p=0.012$ as shown in Table~\ref{tbl:exp1_subjective_results}b). The ``expectation'' here is an ambiguous term that might differ from one person to another; however, it is inherently discussed in the literature that an efficient collaboration is achieved when the partners reach a joint intention. Thus, the expectations of the partners are often toward understanding each other and obtaining a joint action on a task \citep{Bauer2008}.

Finally, Figure~\ref{fig:exp1_rewards} shows the rewards gathered by the cobot, which is the combination of task success and the number of warnings. As expected, the proactive robot has received $2.6$ times more rewards than the reactive robot ($p < 0.0001$). With that and the analysis above, we conclude that our A-POMDP cobot model (the proactive robot) leads to a more efficient and natural collaboration when compared to the same cobot that does not handle a human's unanticipated behaviors, which supports \textbf{Hypothesis~\ref{hypo_3}}.

% trust
Hypothesis~\ref{hypo_4} is evaluated through the subjective statements of the participants in Table~\ref{tbl:exp1_subjective_results}.
Firstly, since the collaboration with both of the cobots achieved very high success rates, the participants rated their trust in both of the cobots very high (the proactive robot with the mean rating of $4.43$, the reactive robot with $4.05$ out of $5.00$ in Table~\ref{tbl:exp1_subjective_results}c). The participants still thought that the proactive robot is significantly more trustworthy than the reactive one (with $p=0.041$). This is in line with and can be explained through the other subjective statements. The participants think that the proactive robot took over the task with significantly more accurate timing, i.e., when they needed assistance, than the reactive robot (with $p=0.002$). This is also supported by the statements of ``The robot acted as I expected'' and ``The robot was able to adapt to my assistance needs'' that are both rated significantly higher for the proactive robot (with $p=0.012$ and $p=0.004$, respectively, in Table~\ref{tbl:exp1_subjective_results}b). Similarly, another consistent analysis is made for the negative statement, ``I did not want the robot to take over the task'', which is rated significantly higher for the reactive robot that took over more frequently ($p=0.030$ in Table~\ref{tbl:exp1_subjective_results}c). All in all, these analyses point out a better collaboration experience with the proactive robot, resulting in a higher level of trust.

% Positive teammate traits
In general, better anticipation of a person's assistance needs, respecting her preferences, and greater trust suggest a higher acceptance for the proactive robot. In addition, the proactive robot contributed to an increased performance of its partner, and led to more efficient task completion. We thus conclude that the proactive robot has more positive teammate traits than the reactive one. The participants also indirectly support this by stating that they would prefer to work with the proactive robot on this kind of demanding tasks significantly more than the reactive robot, even though both of the cobots are rated high ($\mu_{proactive}=4.50$ out of $5.00$, with $p=0.004$ as shown in Table~\ref{tbl:exp1_subjective_results}c). More positive teammate traits and a higher trust may already indicate a higher perceived collaboration for the participants. As a supporting statement, the participants think that the reactive robot was significantly more competitive in its behaviors, whereas the ratings for the proactive robot were below average (with $p<0.001$ in Table~\ref{tbl:exp1_subjective_results}c). However, since the partners share a mutual goal, competition is not encouraged for better team performance. Finally, the participants affirm that they feel more comfortable with the proactive robot ($p=0.003$) and they are more pleased collaborating with it ($p=0.011$), all pointing out a higher perceived collaboration for the proactive robot.

%Merge two below for adaptation
Many of the positive traits of the proactive robot mentioned is a result of better human adaptation skills, which is hard to directly observe and evaluate in HRI in general, especially when it requires reasoning about the hidden human states in a static environment. Thus, the ground truth information, whether a cobot has correctly anticipated and adapted to a human is not explicit and can only be known to the interacted human. In Table~\ref{tbl:exp1_subjective_results}b, we show that the participants think the proactive robot was able to adapt significantly better to their assistance needs (with $p=0.004$) whereas the reactive robot behaves more repetitively rather than responding to their changing behaviors ($p<0.001$). Finally, at the end of the experiments we asked the participants two direct technical comparison questions where the robots are renamed anonymously (see comparison category in Table~\ref{tbl:exp1_subjStatements}). For the first question, the vast majority of the participants,i.e., $71.4\%$, picked the reactive robot as the cobot that is following preset rules instead of responding to their changing needs and preferences. For the second one, $78.6\%$ of the subjects picked the proactive robot as the one that was learning and adapting better to the participant's assistance needs. From these statements we conclude that our A-POMDP model adapting to a human's unanticipated behaviors shows better adaptation skills and it has a higher perceived collaboration, trust, and positive teammate traits, than a cobot model that does not handle such behaviors, which supports \textbf{Hypothesis~\ref{hypo_4}}.

With this experiment, we show the negative impact of the unanticipated human behaviors on the fluency of a collaboration, which are mostly overlooked in HRC studies. Despite the diverse backgrounds of the participants, their statements show great consistency, suggesting that the unexpected cobot interference occurs mostly due to the wrong anticipation of and adaptation to the current behavior of a person, which then results in a significantly less efficient and natural collaboration. Since our robot decision model is a POMDP, it is not learning from the history of interaction but it was able to reach to conclusions like ``the human may still be assessing the subtask (after a long wait) as she has mostly succeeded so far'' or ``this long wait may indicate tiredness after a decrease in her performance. Such conclusions are reached due to the probabilistic distributions over our state machine design with multiple anticipation stages (see Figure~\ref{fig:apomdp}). However, we are aware that the same intrinsic parameters of our A-POMDP design may not respond reliably to all human types. Hence, the next section discusses our long-term adaptation mechanism. With this experiment and analysis, we validate our cobot's extended short-term adaptation skills on a real setup, which is in line with our simulation experiments in \citep{Gorur2018}.

\subsection{Evaluation of Long-Term Adaptation and Complete Framework}
\label{ssec:eval_longTerm}

In this section we evaluate the extended long-term human adaptation and the integrated performance of FABRIC, i.e., the full system with the \textit{adaptive policy selection} component in Figure~\ref{fig:framework}.
In this experiment, our goal is to support the hypotheses:
\begin{hypo}\label{hypo_5}
	Our collaboration setup and the task induce changes in long-term human characteristics, such as their expertise and collaborativeness.
\end{hypo}
\begin{hypo}\label{hypo_6}
	Our framework with A-POMDP models and the ABPS mechanism (the full FABRIC) provides a personalized, fast, and reliable adaptation to both short- and long-term changing human behaviors and characteristics, while it is perceived to have high collaboration skills, positive teammate traits, and trust.
\end{hypo}

As mentioned in Section~\ref{ssec:training}, the base model used in creating the policy library is the proactive robot model from the experiments in Section~\ref{ssec:eval_shortTerm}. This has already led to more natural and efficient collaboration in a challenging work environment. However, we have observed dynamic characteristics and preferences in the participants during the experiments. Selecting different policies would provide even better adaptation toward more personalized collaboration. In this experiment, our first goal is to show that the ABPS mechanism is able to provide this adaptation fast and reliably, improving a cobot's collaboration performance. Also, we want to prove the effectiveness and applicability of our complete system in a real-world scenario.

\subsubsection{Objective and Subjective Measures}
\label{ssec:exp2_measures}
\hfill \break

We evaluate the long-term adaptation capability of the system by highlighting the long-term differences in human behaviors and the cobot's ability to detect and respond to such changes. We use the same objective measures as in Section~\ref{ssec:exp1_measures}, which are the overall task success rate, the number of warnings received from a participant, a human's success rate, a human's contribution to the overall success and the task efficiency, but we also analyze their changes in time over the course of a collaboration for the long-term effects. In addition, we calculate the regret for a selected policy which denotes the distance of a total discounted reward collected in a task from the maximum utility, i.e., a discounted reward that can be obtained by the best policy for the current human type.
The subjective measures are obtained from the questionnaires that are given to the participants (in Table~\ref{tbl:exp2_subjStatements}). We use some of the statements from the previous experiments with minor additions to analyze the participant statements over time. This gives more insights on the dynamics of the participants' changing stamina, motivation, and perceived difficulty of the tasks, and how they perceive the cobot's collaboration skills over time.

\begin{table}[!htbp]
	\caption{Subjective statements asked during the experiments.}
	\centering
	\includegraphics[width=\columnwidth]{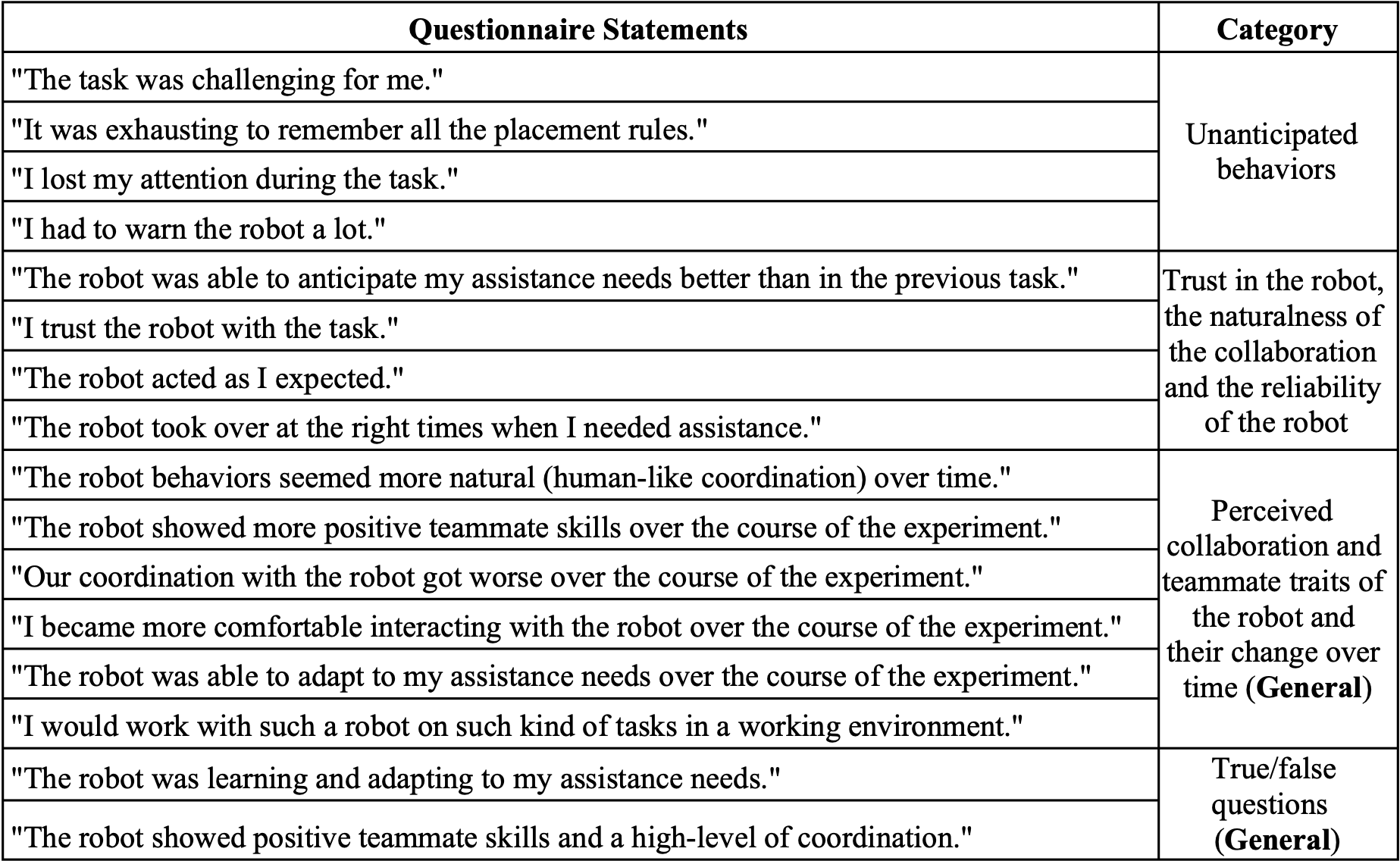}
	\label{tbl:exp2_subjStatements}
\end{table}

\subsubsection{Experiment Protocol}
\label{ssec:exp2_protocol}
\hfill \break

We invited 11 people, who again had no previous interaction with a robot, and are from different backgrounds and demographics (ages between 18 and 35). As mentioned, our purpose is to show the effectiveness and reliability of our complete framework, along with its long-term adaptation capability. We follow mostly the same protocol as in Section~\ref{ssec:exp1_protocol}). The only difference is that we do not run and compare different cobot decision models as in the previous experiment but only the complete framework. However, the ABPS mechanism may select a different decision model for a task; therefore, the participants are asked to compare the cobot performance between the tasks, where possible, without knowing that its strategy may change. Each participant works on 8 tasks in a row and fills out a survey evaluating the performance of each task right after it is completed. 
Then, at the end of the experiment they answer another survey for the general statements (in Table~\ref{tbl:exp2_subjStatements}). We use task type-5, demonstrating the Stroop effect \citep{stroop_studies_1935} in Figure~\ref{fig:taskRules}, throughout the experiment (for all 8 tasks) to ensure the task conditions remain constant. This task invokes the learning effect and the participants could gain noticeable expertise within an experiment, which normally requires more practice. At the beginning of the experiments the participants did not know the tasks have the Stroop effect, and it took them several tasks to notice and master it. In addition, each experiment takes approximately 1.5 hours to complete; hence, we expect to observe accumulated tiredness or a decrease in motivation.

\subsubsection{Results and Discussions}
\label{ssec:exp2_results}
\begin{figure}[!t]
	%\hspace{0.01\columnwidth}
	\begin{minipage}[c]{\columnwidth}
		\centering
		\subfloat[ \label{fig:exp2_coLoad}]{{\includegraphics[width=0.45\columnwidth]{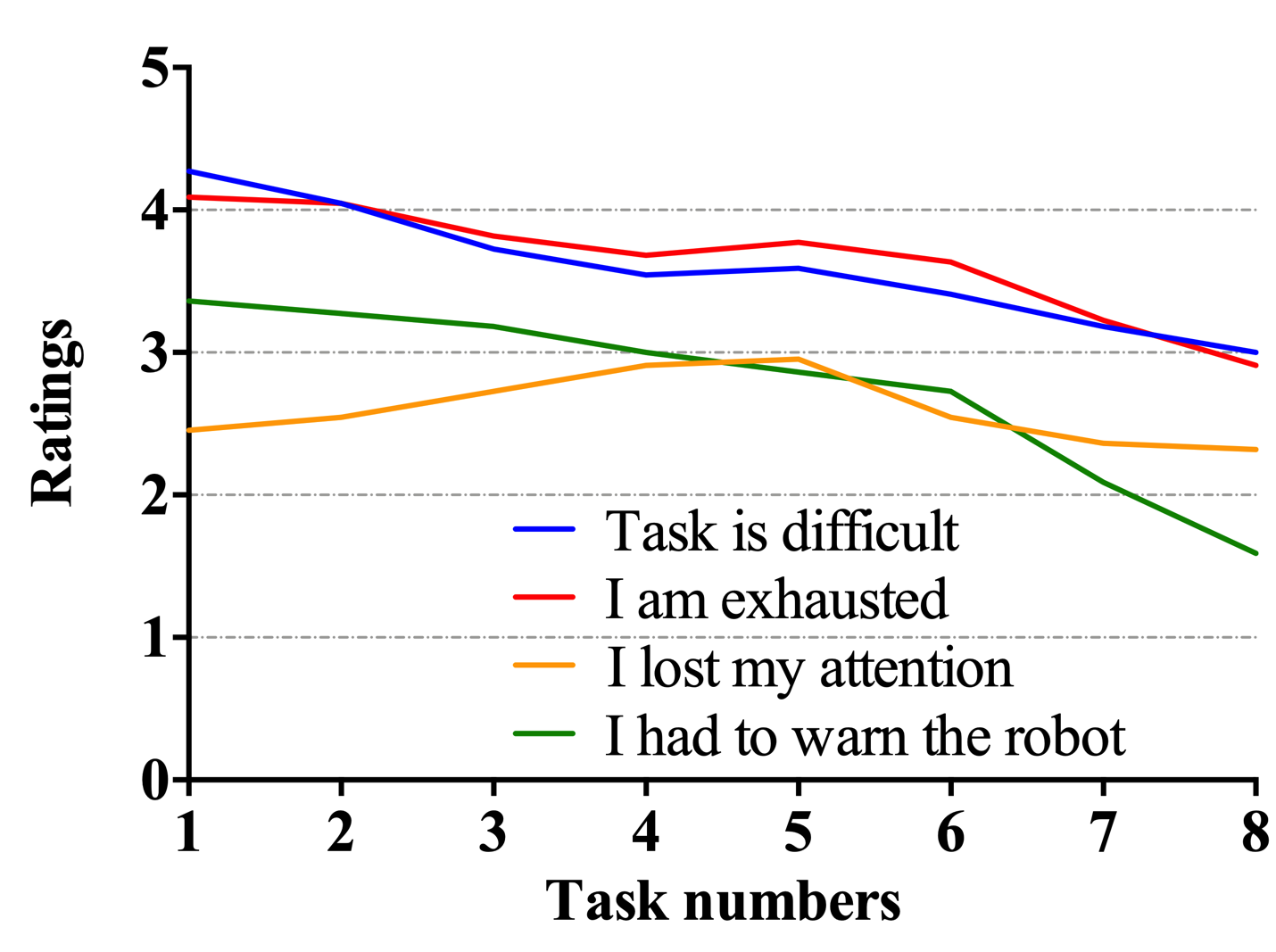} }}
	%\end{minipage}
	%\begin{minipage}[c]{\columnwidth}
		\hspace{0.05\columnwidth}
		\centering
		\subfloat[ \label{fig:exp2_humanSuccess}]{{\includegraphics[width=0.45\columnwidth]{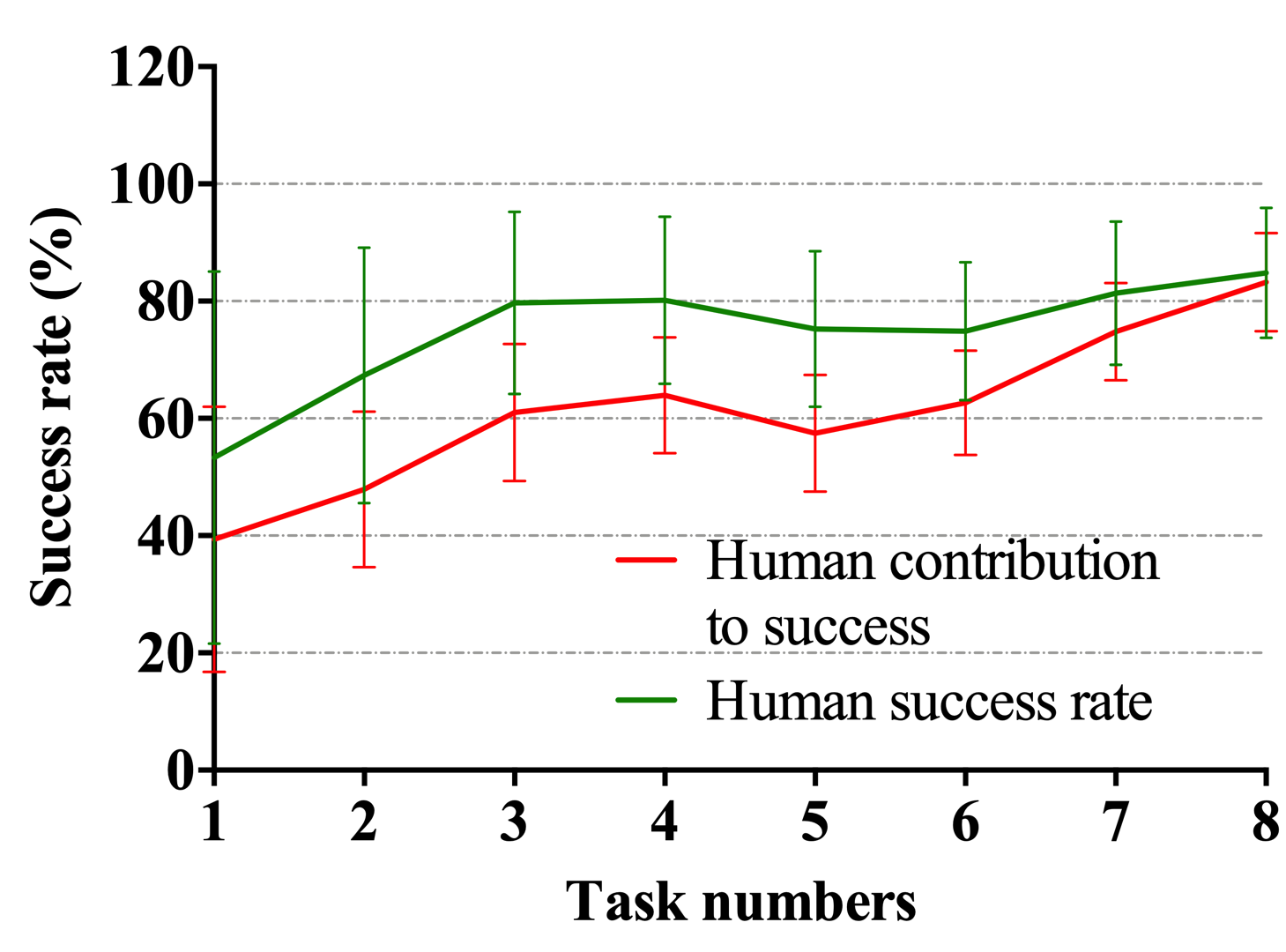} }}%
	\end{minipage}
	\caption{a) The subjective ratings of the participants on their cognitive load and trust in the robot over time. b) The participant's moving averaged success rates and their contribution to success over time.}
	\label{fig:exp2_humanChanges}
\end{figure}

\begin{figure}[!t]
	%\hspace{0.01\columnwidth}
	\begin{minipage}{\columnwidth}
		\centering
		\subfloat[ \label{fig:exp2_anticipateTime}]{{\includegraphics[width=0.46\columnwidth]{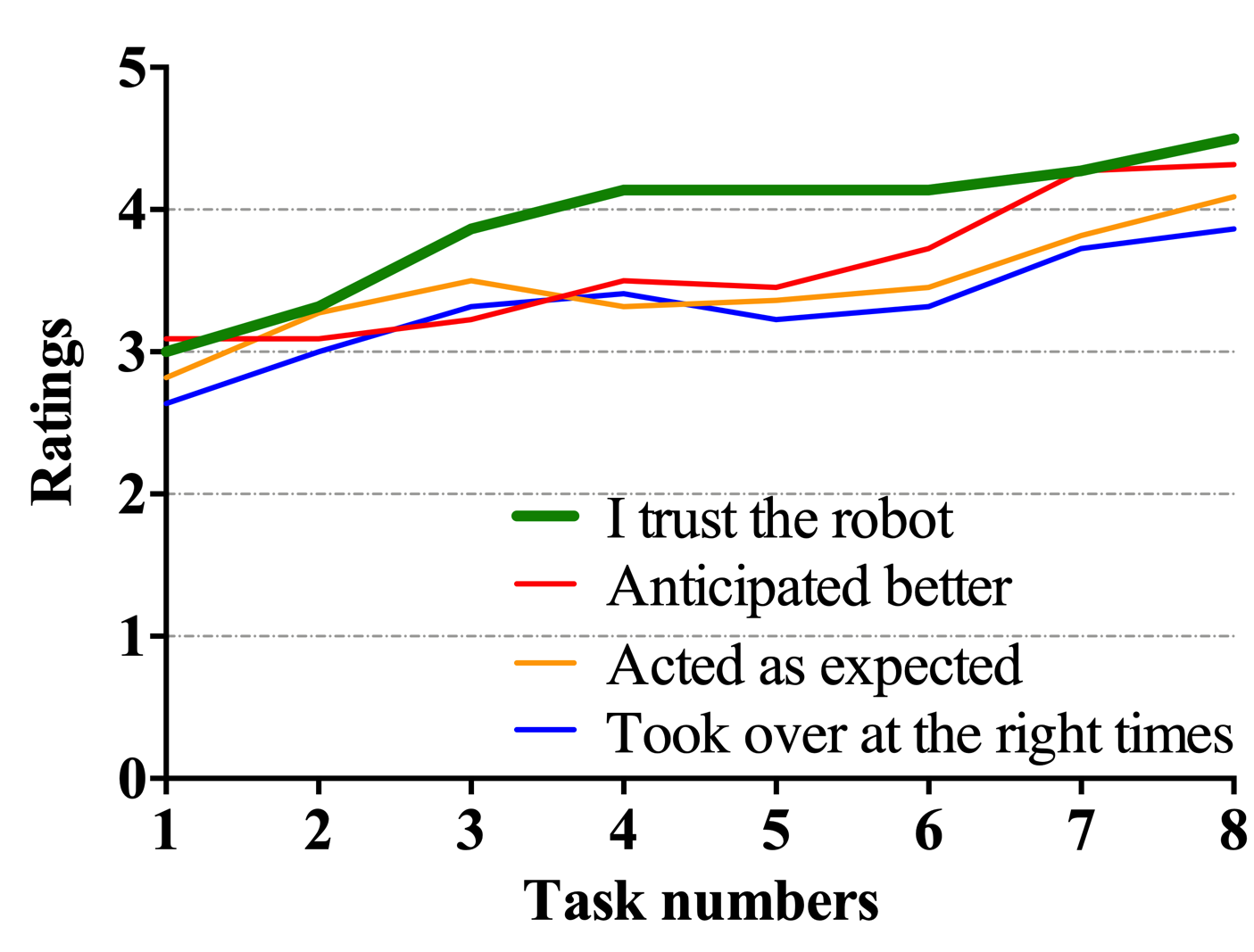} }}
	%\end{minipage}
	%\begin{minipage}[c]{\columnwidth}
	\hspace{0.02\columnwidth}
		\centering
		\subfloat[ \label{fig:exp2_anticipateBox}]{{\includegraphics[width=0.48\columnwidth]{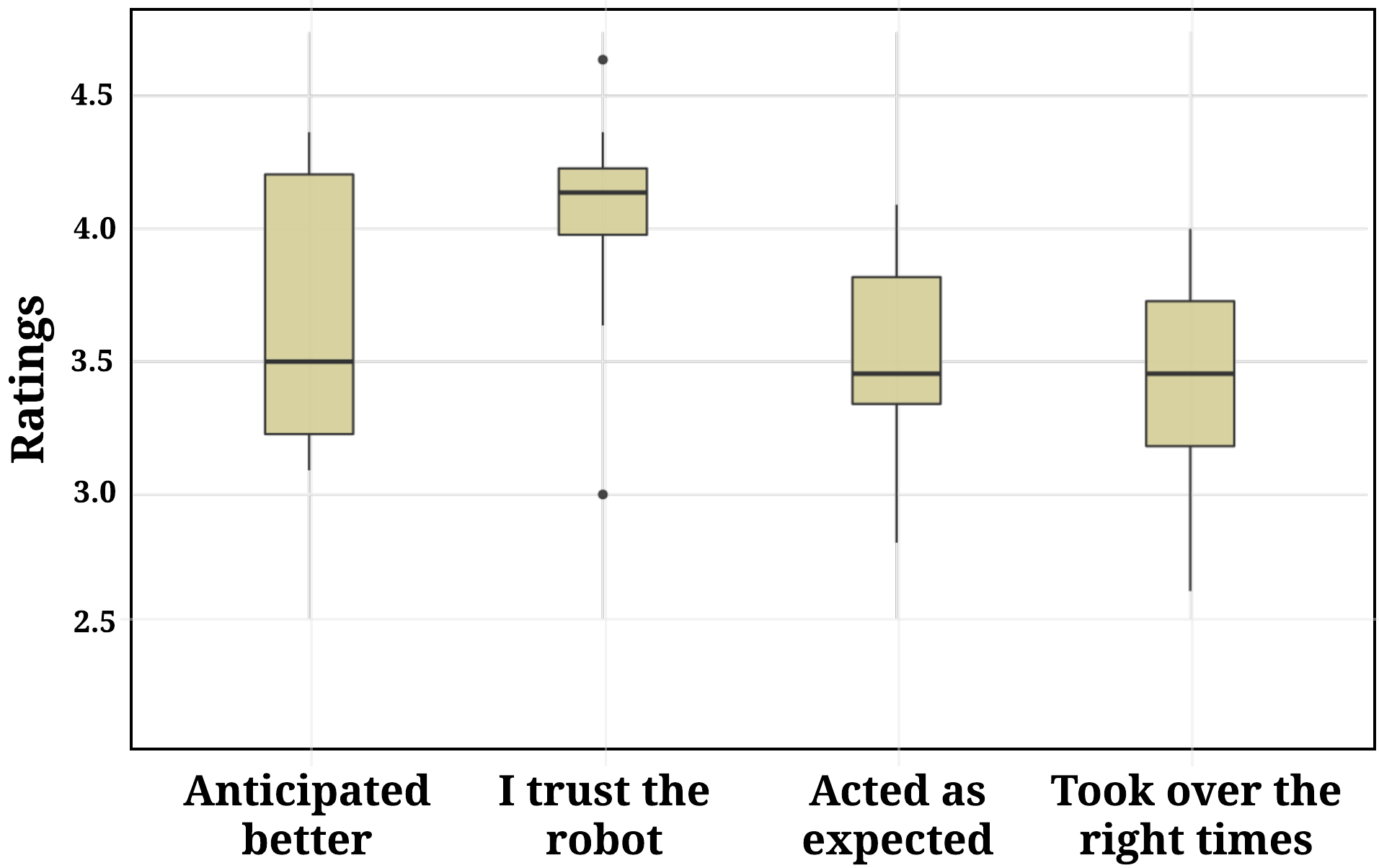} }}%
	\end{minipage}
	\caption{The subjective evaluations of the participants to the statements of: ``The robot was able to anticipate my needs and behaviors better than in the previous task.'', ``I trust the robot.'', ``The robot acted as I expected.'' and ``the robot took over at the right times when I needed assistance.''. a) shows the change of the participant ratings over the task assignments, b) is the box and whisker plots of the overall ratings.}
	\label{fig:exp2_robot_anticipation}
\end{figure}

\hfill \break
We first analyze the subjective evaluations of the participants on their perceived task difficulty, attention, exhaustion, and collaborativeness. Even though the difficulty and the length of a task always remain the same, the participants have perceived the tasks as being easier and less exhausting over time. This is in line with our measurements from Figure~\ref{fig:exp2_humanSuccess} visualizing the increasing trend of the human success rates and the successful contribution of the human in a task. We deduce that the participants gained more expertise and got used to the task and the environment, which indicates the practice-effect that also affects their perceived difficulty of and the exhaustion in a task. The collaborativeness of the participants has also changed during the experiments. Figure~\ref{fig:exp2_coLoad} demonstrates that the participants warned the cobot less over time whereas Figure~\ref{fig:exp2_anticipateTime} shows that their trust in the cobot increased, both indicating that they become more collaborative in time. That said, their expertise, collaborativeness, stamina, and motivation (i.e., through their handling of more subtasks) do change over time and that a cobot should adapt to them. We conclude that our assumption about the change of human characteristics is valid and that we could invoke and observe this during our experiments, which supports Hypothesis~\ref{hypo_5}.

\begin{figure}[!t]
	%\hspace{0.01\columnwidth}
	\begin{minipage}[c]{\columnwidth}
		\centering
		\subfloat[ \label{fig:exp2_reward}]{{\includegraphics[width=0.48\columnwidth]{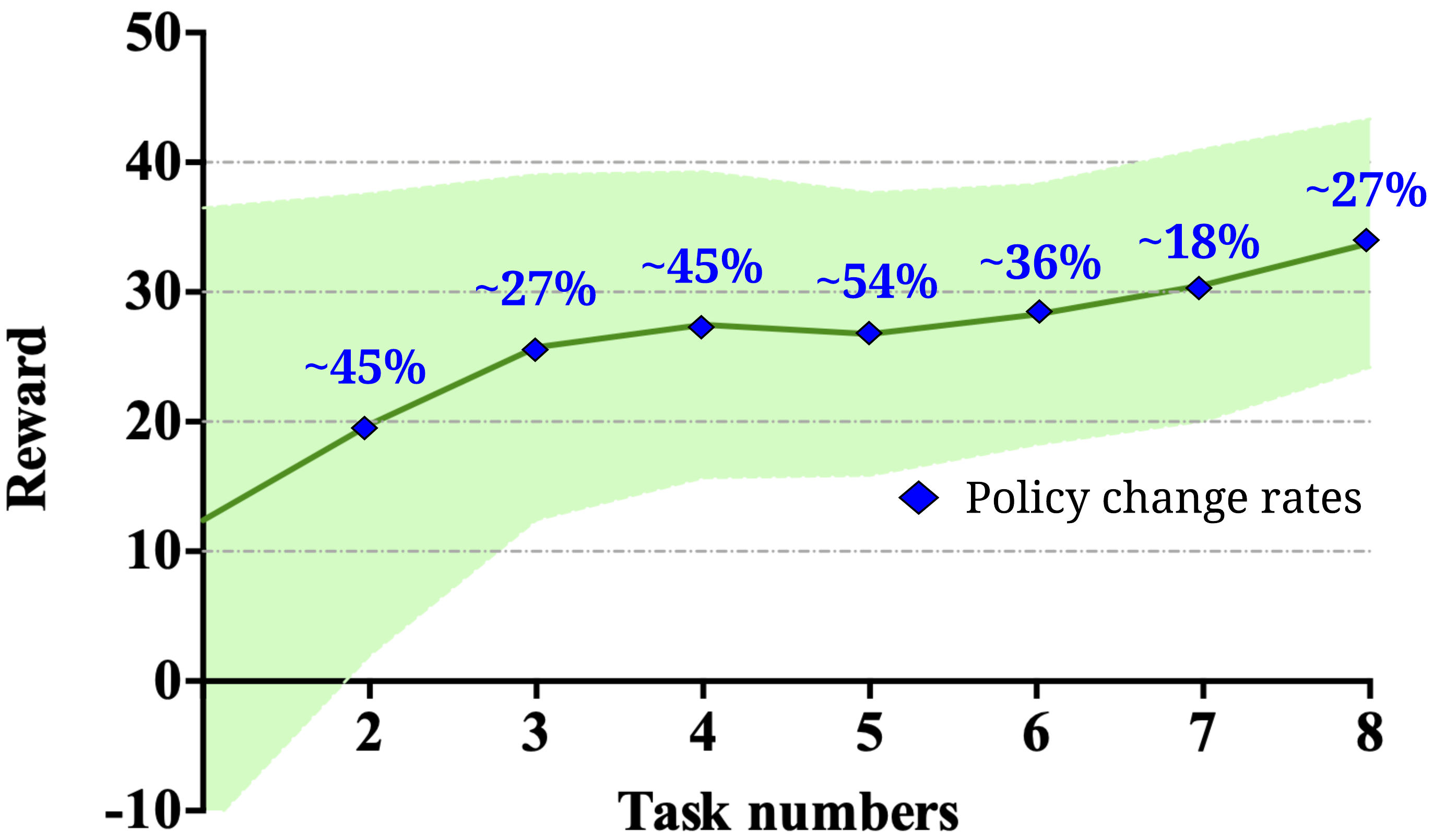} }}
		\centering
		\subfloat[ \label{fig:exp2_regret}]{{\includegraphics[width=0.48\columnwidth]{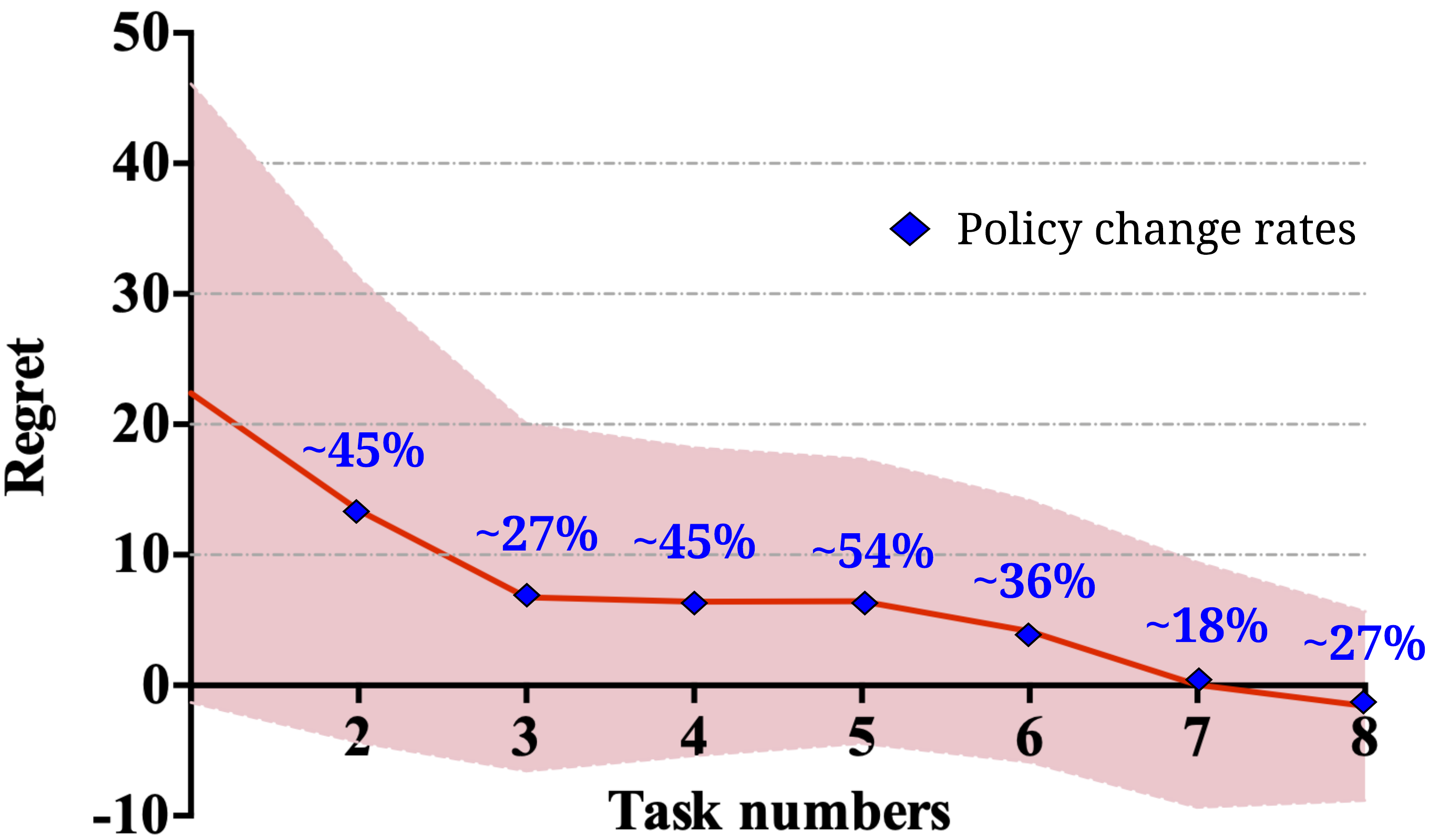} }}%
	\end{minipage}
	\begin{minipage}[c]{\columnwidth}
		\centering
		\subfloat[ \label{fig:exp2_taskSuccess}]{{\includegraphics[width=0.48\columnwidth]{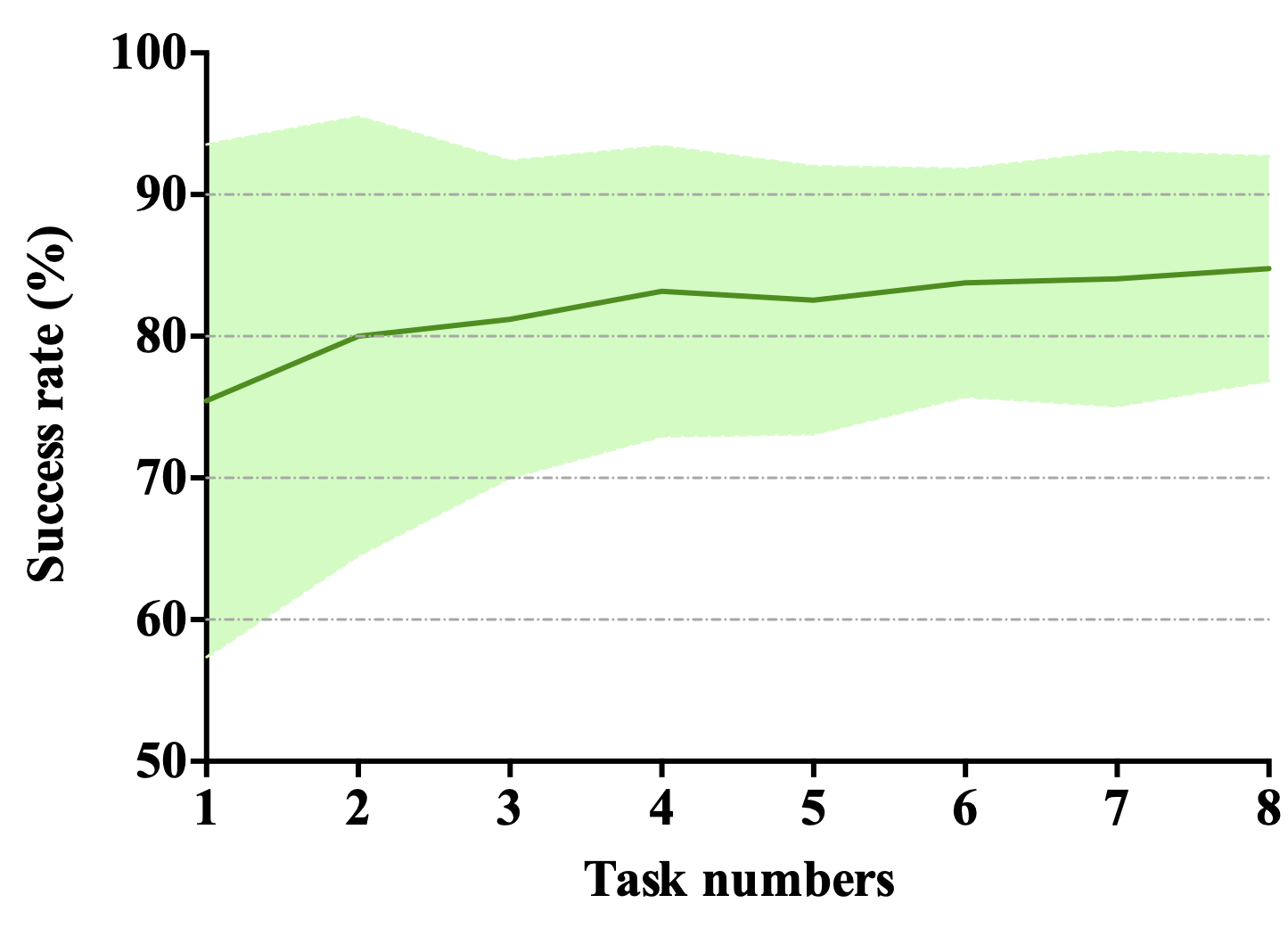} }}%
		\centering
		\subfloat[ \label{fig:exp2_warning}]{{\includegraphics[width=0.48\columnwidth]{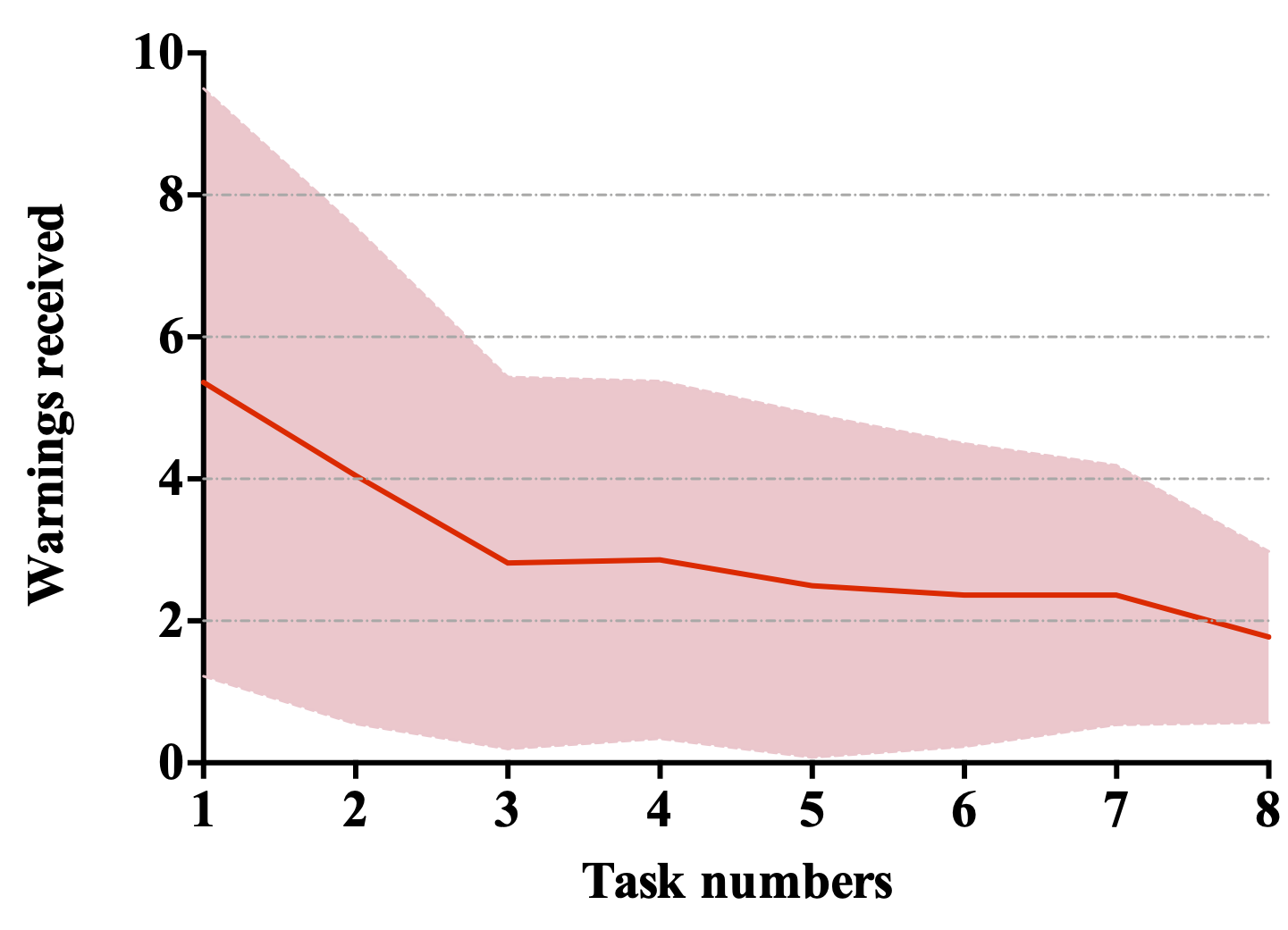} }}
	\end{minipage}
	\begin{minipage}[c]{\columnwidth}
		\centering
		\subfloat[ \label{fig:exp2_efficiency}]{{\includegraphics[width=0.48\columnwidth]{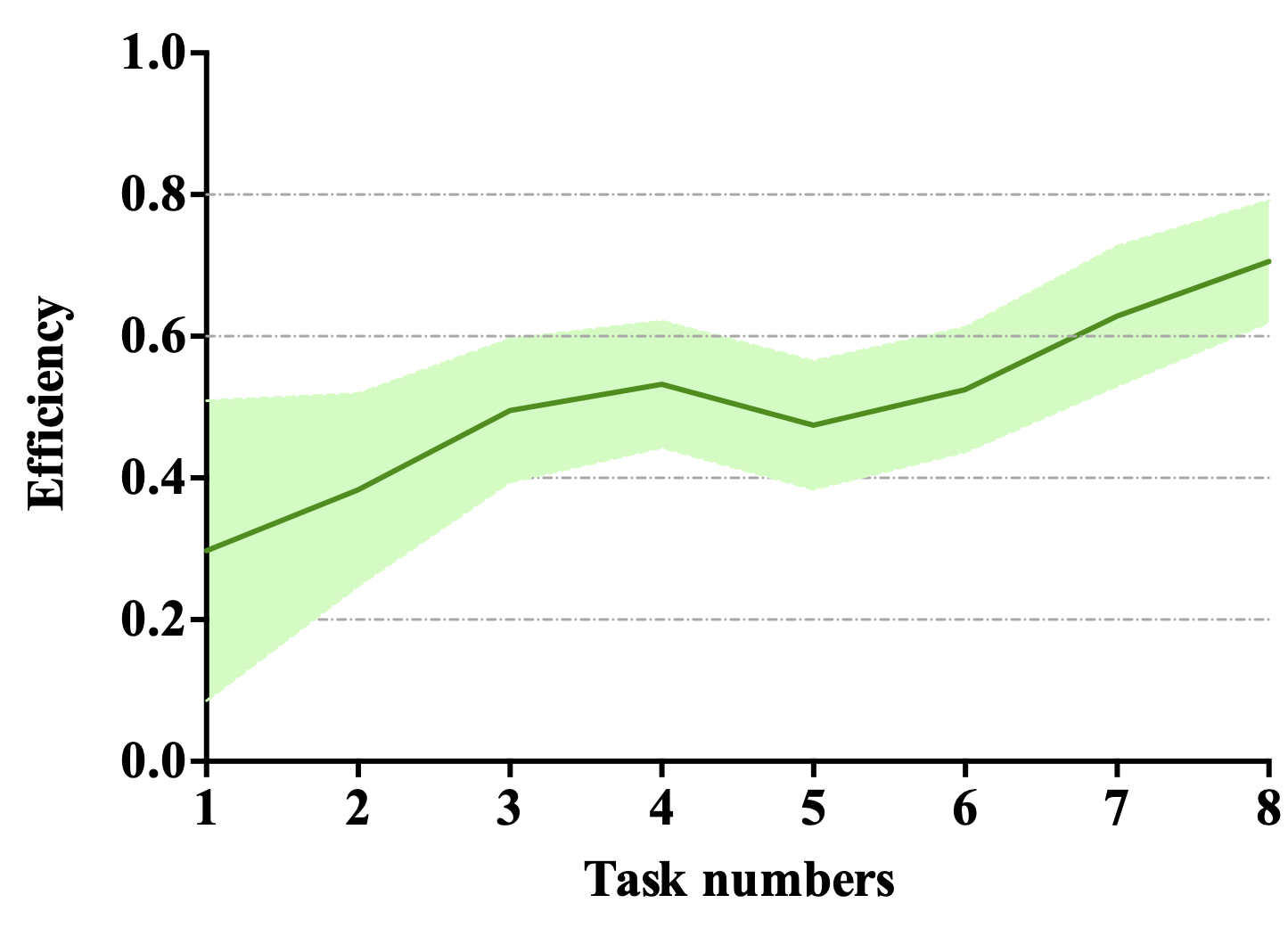} }}%
		\centering
		\subfloat[ \label{fig:exp2_timeTook}]{{\includegraphics[width=0.48\columnwidth]{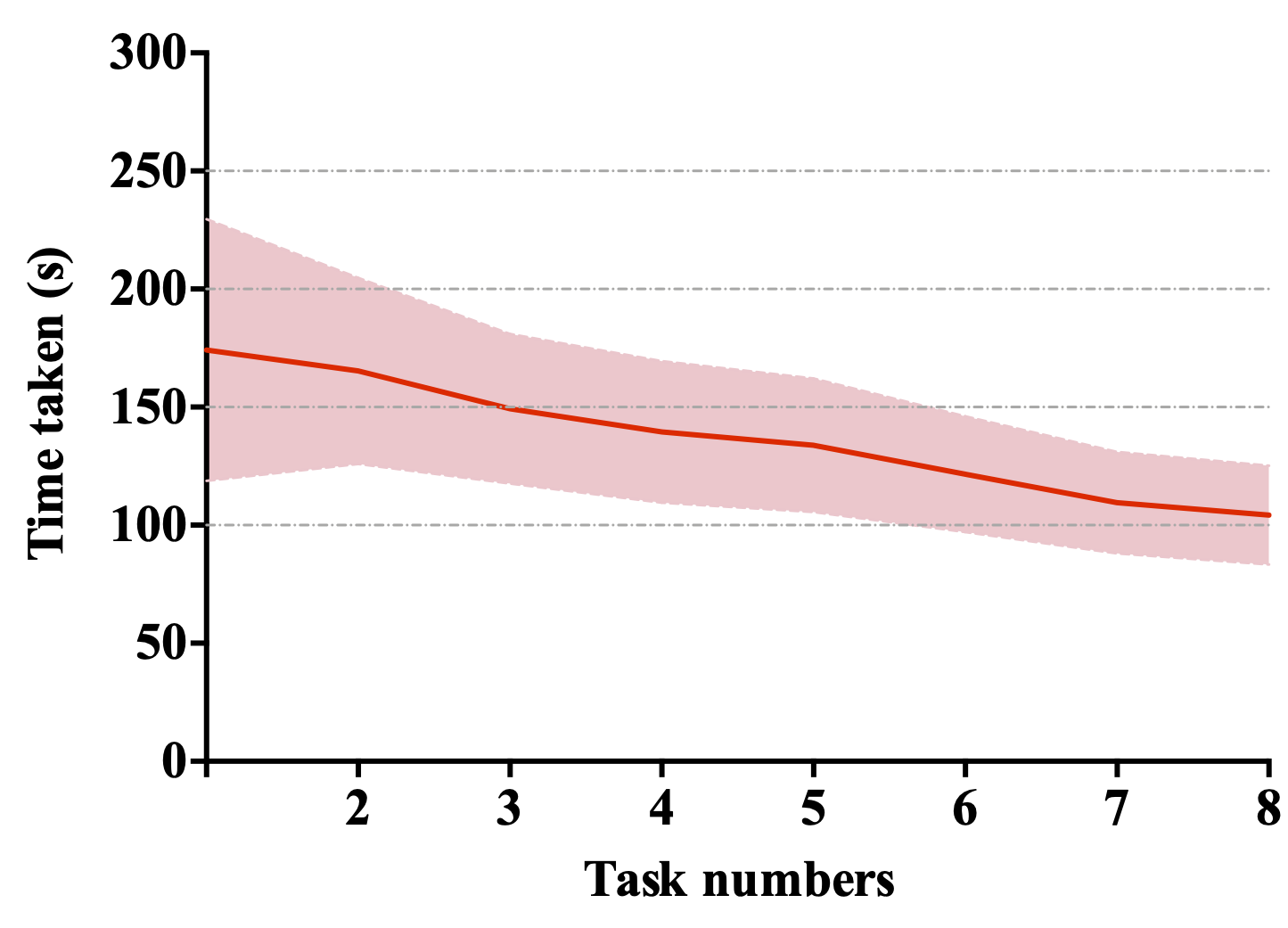} }}
	\end{minipage}
	\caption{The first two plots give the performance of ABPS: a) moving average total discounted reward over task assignments collected by the cobot; b) moving average regret over task assignments with error bars denoting the standard deviation, collected by the cobot. The policy change rates on these two figures denote the average rate of the cobot selecting a different policy before the task starts, whereas the reward and the regret values on the data points are collected at the end of that task. The others show the overall performance of the cobot averaged over the task assignments: c) moving average of the task success rate; d) number of warnings received by the cobot; e) moving average of the task efficiency as in Section~\ref{ssec:exp1_measures}; f) moving average of the task durations in seconds.}
	\label{fig:exp2_robotPerformance}
\end{figure}

We expect the ABPS mechanism to select a different policy when, for instance, a participant is more motivated to complete a task herself due to her increased expertise. In that case, the cobot should give more space to a human to finalize a task. For example, an A-POMDP policy needs to be selected that less favors a state transition from ``human is not struggling'' to, e.g., ``human may not be capable'' (in Figure~\ref{fig:apomdp}). As another example, some of the participants built up more trust in the cobot resulting in leaving a task more often to the cobot. Such drastic and stochastic behavioral changes are difficult to model in a single decision-making strategy. In Figure~\ref{fig:exp2_robotPerformance}, we demonstrate how ABPS has responded to such changes. Figure~\ref{fig:exp2_reward} gives the total discounted rewards the cobot has collected over time. In the same figure, we also show the average rate of ABPS selecting a different policy for the new task to start (i.e., policy change rates as data points).
For instance, before the second task starts, ABPS has picked another policy in $~45\%$ of the experiments, according to its current estimate of that human type. With that, at the end of the second task the collected rewards on average have been almost doubled compared to the first task. As another example, the number of warnings (in Figure~\ref{fig:exp2_warning}) increased at the fourth task with the increasing human contribution (i.e., expertise) as in Figure~\ref{fig:exp2_humanSuccess}. As a reaction, ABPS has picked another strategy at the fifth task in $~54\%$ of the experiments, mostly leaving the task to the participants. This has successfully dropped the number of warnings by $~20\%$; however, the participants have reached a less success rate on average (in Figure~\ref{fig:exp2_humanSuccess} and Figure~\ref{fig:exp2_taskSuccess}). Then, the cobot has picked another policy at the sixth task, which has balanced the collaboration better and resulted in $~14\%$ increase in the task efficiency (in Figure~\ref{fig:exp2_efficiency}). This shows the significant positive effect of the ABPS mechanism.

Considering the plots in Figure~\ref{fig:exp2_robotPerformance}, by the fourth task, ABPS has found a policy that is nearly optimal for the interacted humans, when averaged over all of the experiments. In general, we can say that a mutual steady-state has almost been reached after the third task where both the human and the cobot have more or less satisfactory coordination (see the stabilizing curves in the plots). This is also stated by the participants through their almost stabilized ratings to, for example, ``I trust the robot'' in Figure~\ref{fig:exp2_anticipateBox} and their contribution to success in Figure~\ref{fig:exp2_humanSuccess} after the third task. We can see the same effect also on the task efficiency in Figure~\ref{fig:exp2_efficiency}.
We note again the efficiency drop at the fifth task; however, this is also compensated at the next task by the cobot. The constant increase in the efficiency after the sixth task can be explained by the increasing human expertise (in Figure~\ref{fig:exp2_humanSuccess}) and the cobot's successful adaptation to it. This adaptation is mutual, where both the human and the cobot have finally achieved higher levels of coordination. The cobot first selected a strategy (exploration) that has left almost the entire task to the participants at the fifth task. But eventually it could find strategies that mostly leaves the tasks to the participants and takes over the tasks only if the human failed consistently or idled for longer. The latter was often observed after the participants discovered when the cobot usually takes over and so they were intentionally waiting for the cobot to pick up when they do not remember the rules, which is another example of the mutual adaptation.

Finally, in Figure~\ref{fig:exp2_regret}, we show the performance of ABPS also through the change of the regret values over time. Since the actual human types are unknown to the cobot and to us, for the regret calculation we approximate it by taking the average of the maximum discounted rewards collected by the policies in the policy library during the training phase run on the simulation environment (see Section~\ref{ssec:training}). Hence, we observe negative values of regret at the last task in Figure~\ref{fig:exp2_regret} since the rewards collected at this task were more than this assumed maximum utility value. Overall, the moving average regret has reached an almost steady-state after the third task. 
Then, thanks to the increasing collaborative success with the human expertise and the cobot's adaptation to it with a different policy that received less number of warnings, the regret values have drastically decreased after the fifth task. All in all, such human characteristic changes have been observed between the first and the third tasks and between the fifth and the seventh tasks, which shows that ABPS has gradually adapted to them despite their constant change throughout the experiments. We add that the response of ABPS is considerably fast but it could be faster with a more accurate human type estimation.
Also, we show that the system is reliable as it has continuously contributed positively to the overall task efficiency and human satisfaction, i.e., naturalness, as shown in Figure~\ref{fig:exp2_robot_anticipation}.

\begin{table}[!t]
	\caption{Average ratings of the participant answers to the general statements asked at the end of each experiment.}
	\centering
	\includegraphics[width=\columnwidth]{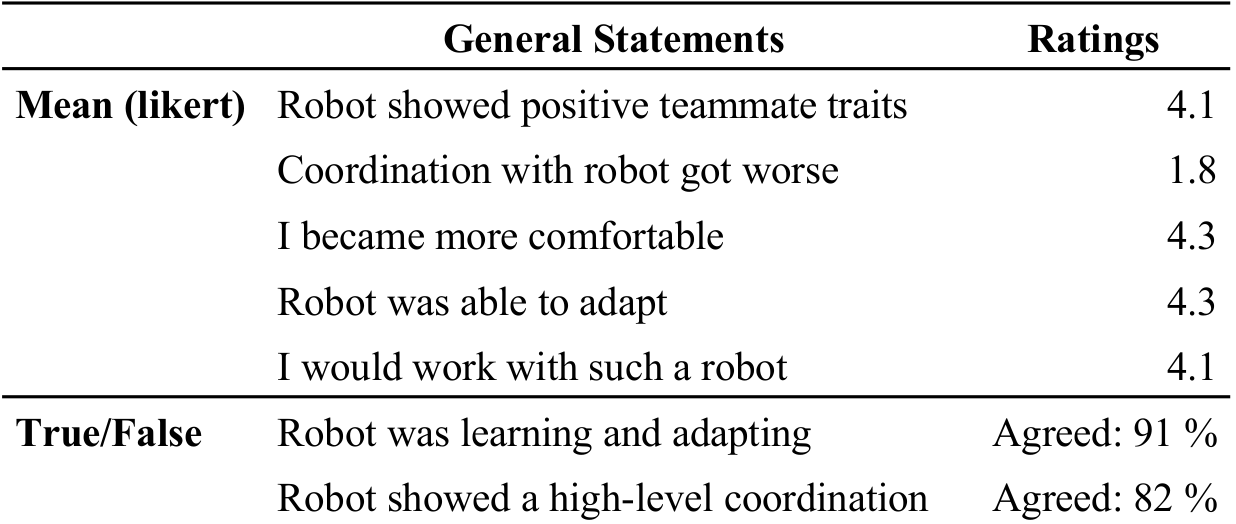}
	\label{tbl:exp2_generalStatements}
\end{table}

In addition to the performance of ABPS, we also evaluate the collaboration skills and teammate traits of our cobot perceived by the participants, and their trust in the cobot. In general, the better anticipation of a person's assistance needs, the correct timing in assisting, the expected and reliable cobot responses, and an increasing trust suggest a higher acceptance for the cobot over the course of the experiments. We see in Figure~\ref{fig:exp2_anticipateTime} that, on average, the participants were satisfied with the performance of the cobot throughout the experiment with the constantly increasing ratings of the given statements, reaching up to the levels of $75\%$ of the maximum rating. In particular, the trust level has reached over $90\%$ of the maximum rating by the end of the experiments.
Our cobot could provide positive teammate traits through anticipating and respecting its partner's needs and preferences, contributing to the increased performance of its partner and leading to more efficient task completion. Supporting this, the participants evaluated that the cobot showed positive teammate traits, it was able to adapt to the changing human needs and preferences, they became more comfortable with the cobot, and they would work with such a cobot with the mean ratings over $4$ out of $5$ as given in Table \ref{tbl:exp2_generalStatements}. In addition, $91\%$ of the participants agreed that the cobot was learning and adapting, and $82\%$ of them confirmed that the cobot has high coordination skills. All these objective measures and the positive subjective statements of the participants support \textbf{Hypothesis~\ref{hypo_6}}.

\section{Conclusion}

In this paper, we focus on an extended human adaptation of collaborative robots (cobots). We propose solutions for the following research challenges: How to anticipate and adapt to unanticipated human behaviors in short-term; how to handle a diversity of long-term human characteristics for a personalized collaboration; how to develop accurate human models that simulate a great diversity of human behaviors in work environments; and how to design a real collaboration experiment that does not assume turn-taking, does not constrain human intention space, and invokes unanticipated human behaviors to properly evaluate our robot's adaptation goals. With the purpose of obtaining an integrated system, we devise our novel lightweight autonomous framework, called FABRIC, that hierarchically integrates our approaches to the challenges above.

To design and evaluate our framework, we propose a pipeline that trains and runs rigorous tests in simulation, then improves and deploys the solution in the real-world for user studies. We first design a novel simulated human model and a 3D factory environment that samples such dynamic human behaviors. We are aware of the possible biases in the experiments which could be introduced by the simulated humans. However, the abstracted states in our design are shown to be observed in real humans through the calibration experiments. On the other hand, we also show that the simulation provides even more diversity than the user studies thanks to its scalable behavior sampling. This and the calibration experiments have proven the necessity of our human simulations in training and for rigorous tests. We then transfer the simulation results into the real-world through our novel experiment setup and our collaboration task that together induce cognitive load on the human participants. We show that we could observe a variety of human responses and preferences, including human behaviors that lead to mistakes and changing human expertise.

For the short-term adaptation of FABRIC, our approach is our novel Anticipatory-POMDP (A-POMDP) model design adapting to a human's changing \textit{intent, attention, tiredness, and capability}, to better estimate if the human needs help and whether the cobot should intervene. Our first round of user studies have shown that handling such human variability increases the overall efficiency and the naturalness of an HRC, also leading to more positive teammate traits and higher trust as perceived by the participants. For the long-term adaptation, we introduce our novel adaptive Bayesian policy selection, ABPS, that runs on top of several A-POMDPs with distinct intrinsic parameters. Toward a personalized collaboration, ABPS selects a model according to an estimate of a human's workplace characteristics, we call types, such as her levels of expertise, stamina, attention, and collaboration preferences. We conduct another user study that deploys our complete framework, FABRIC. At first, we have shown that ABPS provides fast and reliable policy selection in adapting to unknown and changing human types. Then, the objective and subjective results have demonstrated that FABRIC is able to reliably operate in our dynamic environment that does not follow a turn-taking collaboration by fluently coordinating with humans thanks to its extended human adaptation. FABRIC has provided significant improvements in a cobot's human adaptation toward a more natural and efficient collaboration with high perceived teammate skills and trust. We believe that such an adaptation will positively contribute to the acceptance and the long-term use of cobots.

\subsection{Limitations and Future Works}
Our human type estimation is not accurate enough for some cases, such as, when estimating whether a human is tired or a beginner after she idles longer. We are aware that it is a nontrivial task to differentiate such hidden human states. Nonetheless, we believe that our assumption of a limited human action space is also a limiting factor for this. Our simulated humans are also modeled with the same assumption; hence, some simulation runs have generated quite similar behavior patterns for certain human types in the long-term. In this study, since our goal was to highlight the importance of handling the unanticipated human behaviors, the cobot has responded similarly to such cases by offering more assistance. Hence, we have not observed a significant impact of this inaccuracy. However, during the user studies we found that an accurate estimation and more tailored handling of such behaviors would result in even more efficient collaboration. In the future, we believe that more data should be collected from larger scale long-term user studies to model the long-term traits of humans and to generate more accurate human simulations.

During our long-term experiments, some participants have gained enough experience over time and perfectly achieved the tasks alone. In such cases, they started to question the necessity of the cobot. To better analyze our cobot's contribution when the human characteristics change, we need to repeat our experiments that compare the human-alone performance with the cobot collaboration, also for the long-term.
Two participants also added that calling for cobot assistance only when they want to would yield more efficient collaboration. For that, we may experiment on a command and control system to compare it with FABRIC's performance in the long-term. This can also help us analyze the additional cognitive load such a command system may put on the human operators. Additionally, further validation of the contribution of ABPS to the collaboration is needed through a more comprehensive comparison of it with the best policy in hindsight running alone. Finally, we can further improve our system's reliable adaptation and response time by developing a policy change mechanism even during a task.
The improved system could be deployed on a larger scale industrial setup to prove its applicability with the real human workers for a broader impact.

\begin{acks}
This work is a revised and expanded version of our papers
``Social Cobots: Anticipatory Decision-Making for Collaborative Robots Incorporating Unexpected Human Behaviors'' published in the Proceedings of the 11th ACM/IEEE International Conference on Human–Robot
Interaction (HRI'18) and ``Anticipatory Bayesian Policy Selection for Online Adaptation of Collaborative Robots to Unknown Human Types'' published in the Proceedings of the 18th International Conference on Autonomous Agents and Multi-Agent Systems (AAMAS'19). We also thank Elia Kargruber, Minh Nghiem Vi, and Güner Dilsad Er for their contributions.

\section*{Funding}
This work was supported in part by the German Federal Ministry of
Education and Research (BMBF) under Grant 01IS16045.

\end{acks}

\bibliographystyle{SageH}
\bibliography{diss_journal.bib}
% \begin{thebibliography}{99}
% \bibitem[Kopka and Daly(2003)]{R1}
% Kopka~H and Daly~PW (2003) \textit{A Guide to \LaTeX}, 4th~edn.
% Addison-Wesley.
%
% \bibitem[Lamport(1994)]{R2}
% Lamport~L (1994) \textit{\LaTeX: a Document Preparation System},
% 2nd~edn. Addison-Wesley.
%
% \bibitem[Mittelbach and Goossens(2004)]{R3}
% Mittelbach~F and Goossens~M (2004) \textit{The \LaTeX\ Companion},
% 2nd~edn. Addison-Wesley.
%
% \end{thebibliography}
\newpage
\section*{Supplementary Materials}
\pagenumbering{gobble}

\subsection*{Developing, Training, and Testing our Framework}
\begin{figure*}
	\centering
	\caption[Evaluation Pipeline]{Our pipeline for integration, deployment and the evaluation of an anticipatory collaborative robot with extended human adaptation (FABRIC). The layers reflect how the components in our FABRIC framework (in Figure~\ref{fig:framework}) is integrated and deployed first in simulation then on a real setup, following our pipeline in Figure~\ref{fig:pipeline}.}
	\includegraphics[width=\textwidth]{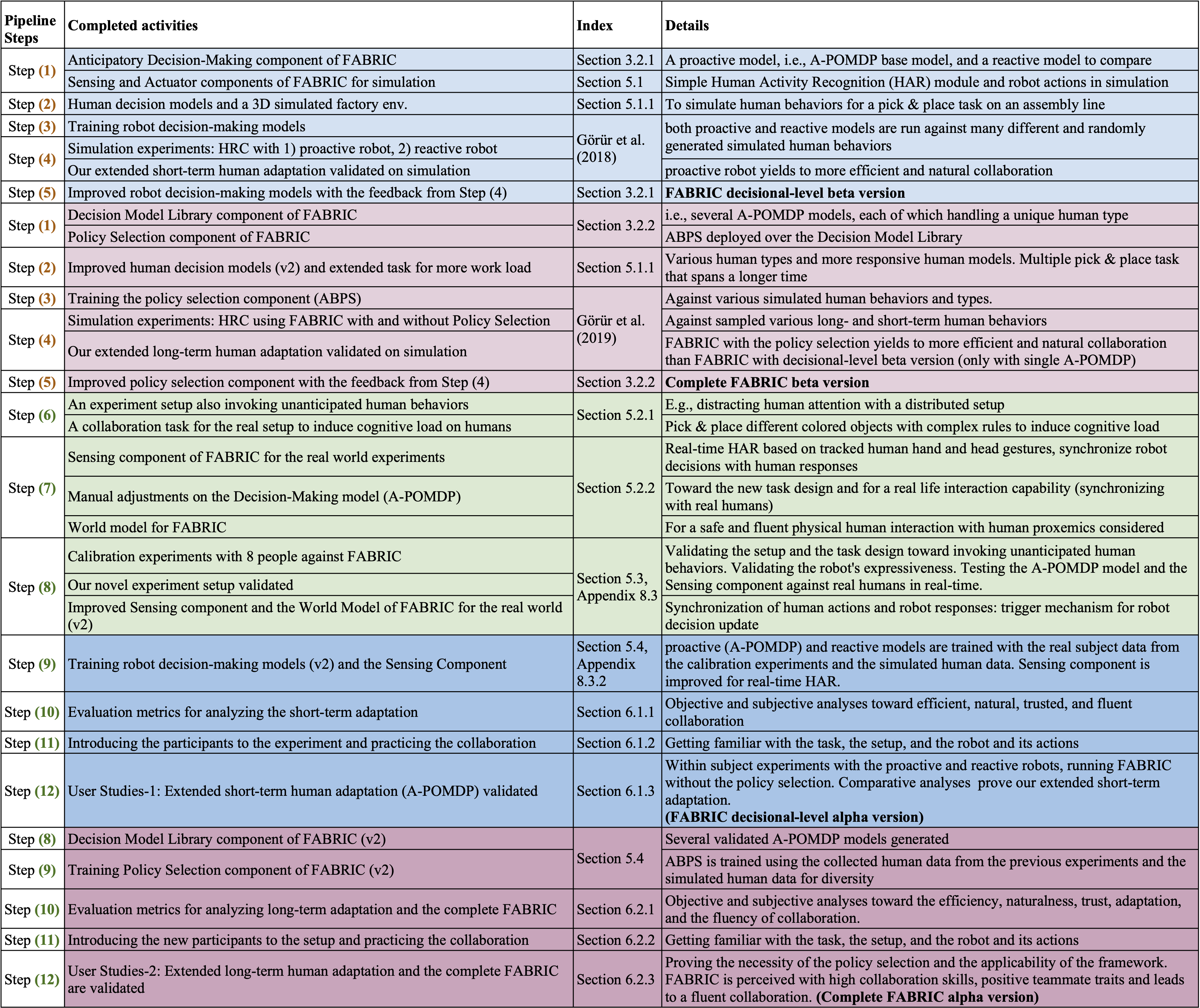}
	\label{fig:pipelineFollow}
\end{figure*}

Here we provide an overview of the development, training and testing phases of our framework by following the pipeline steps. Figure~\ref{fig:pipelineFollow} summarizes all activities with their chronological order and also indexes where these steps are described in this paper. 

\subsection*{Validating the Simulated Human Models}
\label{ssec:appendix_humanValidate}
%NOTE: This para below goes to the journal

In this section, we validate the reliability of the simulated human models by analyzing and comparing their generated human observations with the real observed behaviors from the humans. Our goal is to show that our novel simulated MDP human models generate reliable human behaviors. Nevertheless, we note that not all of the behaviors generated from this modeling scheme are necessarily accurate representations of the real human behaviors since they are sampled with a random factor to increase the variety of the behaviors.
In fact, we believe that such a single human decision model that reflects the greater diversity of human behaviors may not even be possible. In the literature, there exist some abstracted categorizations of such behaviors that reflect some of the possible human characteristics. This is inherited also in our model design, where we abstract the human states in a work environment into the certain intention and behavior sets, which are inspired by the literature. In particular, we are indifferent to the real motivations behind the states the humans are in, which can have infinitely many possibilities (see Figure~\ref{fig:human_model}). Hence, the generated human behaviors from our simulated models are nothing but the sampled reflections of these abstracted states, and the diversity is reached through a random walk between them.

Our approach to validate the human models is to compare the generated observations from the simulated models and the real human observations collected during the evaluation of short-term adaptation in Section~\ref{ssec:eval_shortTerm} (i.e., from 14 participants). As we discuss before, we expect the simulated models to generate much more diversity of human behaviors and characteristics than the ones we have observed from only 14 people during the user studies. It is also possible that some of the observations from the generated models do not reflect a real scenario; however, the models we have used throughout the thesis are run and tested on our 3D simulated environment several times to make sure that they do not always exert unreliable human behaviors, e.g., a human is tired at the very beginning, a human constantly fails and does not let the robot take over. This is configured by manually tuning the decision models.
To prove that the models are able to reflect real-life scenarios, we calculate the likelihood of an observation set being generated from our simulated human models. Each task starts with a task assignment and ends with either a global success or a global fail for both the simulated human models and real humans. Also, the action sets are confined within the work environment; hence, both of the real and the simulated humans generate an observation of interest (abstracted observations for the robot) from the same observation space (in Section~\ref{ssec:obsVector}). Because of that, the generated observation sets differ from each other in terms of the sequence of the executed human actions.

\begin{figure*}[!t]
	%\hspace{0.01\columnwidth}
	\begin{minipage}[c]{\columnwidth}
		\centering
		\subfloat[ \label{fig:simuLikelihoods}]{{\includegraphics[width=\columnwidth]{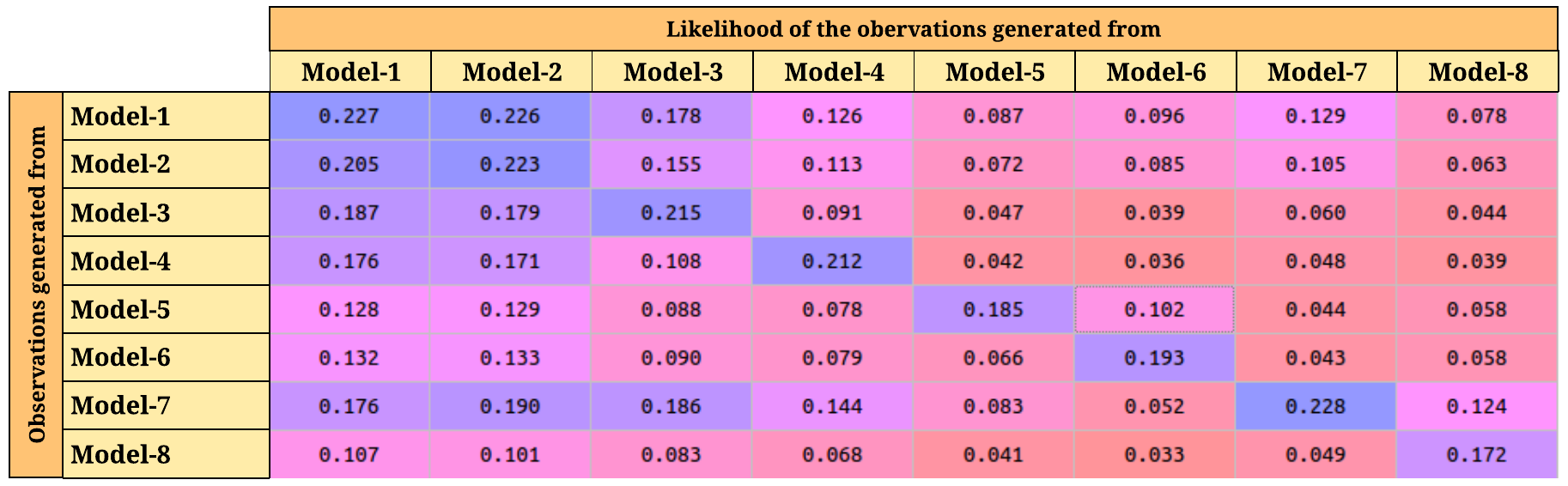} }}
	\end{minipage}
	\begin{minipage}[c]{\columnwidth}
		\centering
		\subfloat[ \label{fig:partLikelihoods}]{{\includegraphics[width=\columnwidth]{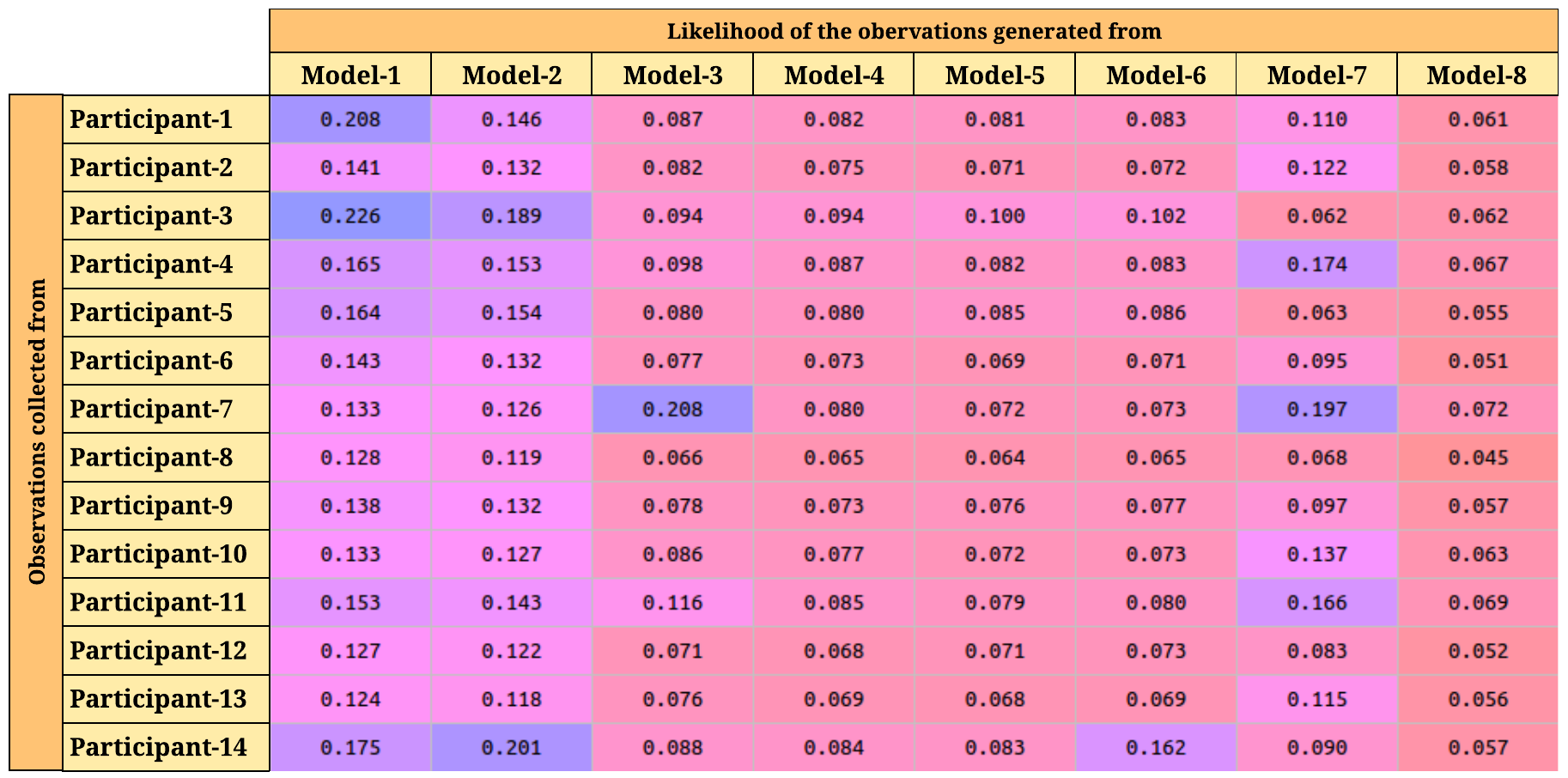} }}%
	\end{minipage}
	\caption{The likelihoods of: a) the simulated observations generated from the same simulated models; b) the participant observations generated from the simulated models. The darker blue the cells get, the higher the probability is.}
	\label{fig:likelihoods}
\end{figure*}

We first take each of the human actions observed in a task one by one sequentially (that is, with the frequency of observation update as in Section~\ref{ssec:obsUpdate}), and calculate the probability of the action generated by the human states in our model design. This gives us a belief distribution on the current human state. For simplicity, we call it ``action belief''. Afterward, we calculate the current belief starting from the initial belief distribution of a simulated human model, and using the action taken and the state transition probabilities of the human model. This gives us an estimate of the actual belief state that the human model would be in if the generated action was taken from it, we call it ``current belief''.
The multiplication of the action belief and the current belief gives us the likelihood of that action generated by the model. We keep multiplying the likelihood values for each observed action in an observation set until a task ends (i.e., until the end of the set). Then, we average the likelihood values obtained from each task to obtain one value for comparison. We repeat this for all of the eight simulated human models we have created and used in our previous experiments, and for all of the real observations obtained from 14 participants. We give the resulting likelihood values of the simulated observations being generated from the same simulation models in Figure~\ref{fig:simuLikelihoods}, whereas in Figure~\ref{fig:partLikelihoods}, we show the resulting likelihood values of the real observations emitted by the participants being generated from the simulated human models. The main reason behind the calculation of the likelihoods of the simulated observations is to have a ground truth, i.e., to know what is the best likelihood values that can actually be obtained given the randomness and all possible state transitions in the simulated humans.

Figure~\ref{fig:simuLikelihoods} shows the distribution of these values over each model. The largest likelihoods are all the diagonal values, which states that the generated observations from each of the models resemble the best likelihoods as expected. That means despite the random walk, the generated actions still reflect the intrinsic characteristics of their models. Yet the values are very small, ranging between $0.17$ and $0.23$. This shows the randomness of our MDP state sampling.
Our belief is that this would give enough diversity even within the same model itself, which reflects the real human behaviors better and which is also our goal to observe a broader range of human dynamics. In Figure~\ref{fig:partLikelihoods}, we show the same calculation, this time for the participant observations. For 14 participants, we observe that the maximum likelihood values are also in the range of $0.14$ to $0.23$, where each participant shows a high resemblance to at least one of the models. This indicates that the probability of a real human observation being generated from one of our simulated models is as likely as they are actually generated from that model. Hence, the observations show a great similarity, supporting the reliability of our models generating realistic human behaviors. It is not possible to run variance analysis as the number of the human observations is very small compared to the simulated ones. Finally, as it is visualized in Figure~\ref{fig:partLikelihoods}, the participants mostly show a resemblance only with three or four out of eight simulated models. This supports our idea that the user studies in lab environments are less likely to provide enough diversity of human behaviors for training and testing collaborative robots. Our simulated models, specifically, Model-4,-5 and -8 in our case, contribute greatly to that diversity.

\newpage
\subsection*{More on the Calibration Experiments}
\subsubsection*{Evaluating Cognitive Load of the Tasks}
\label{ssec:eval_CogLoad}

\begin{table}[!htbp]
	\caption{Subjective statements asked to evalute the cogntive load}
	\centering
	\includegraphics[width=\columnwidth]{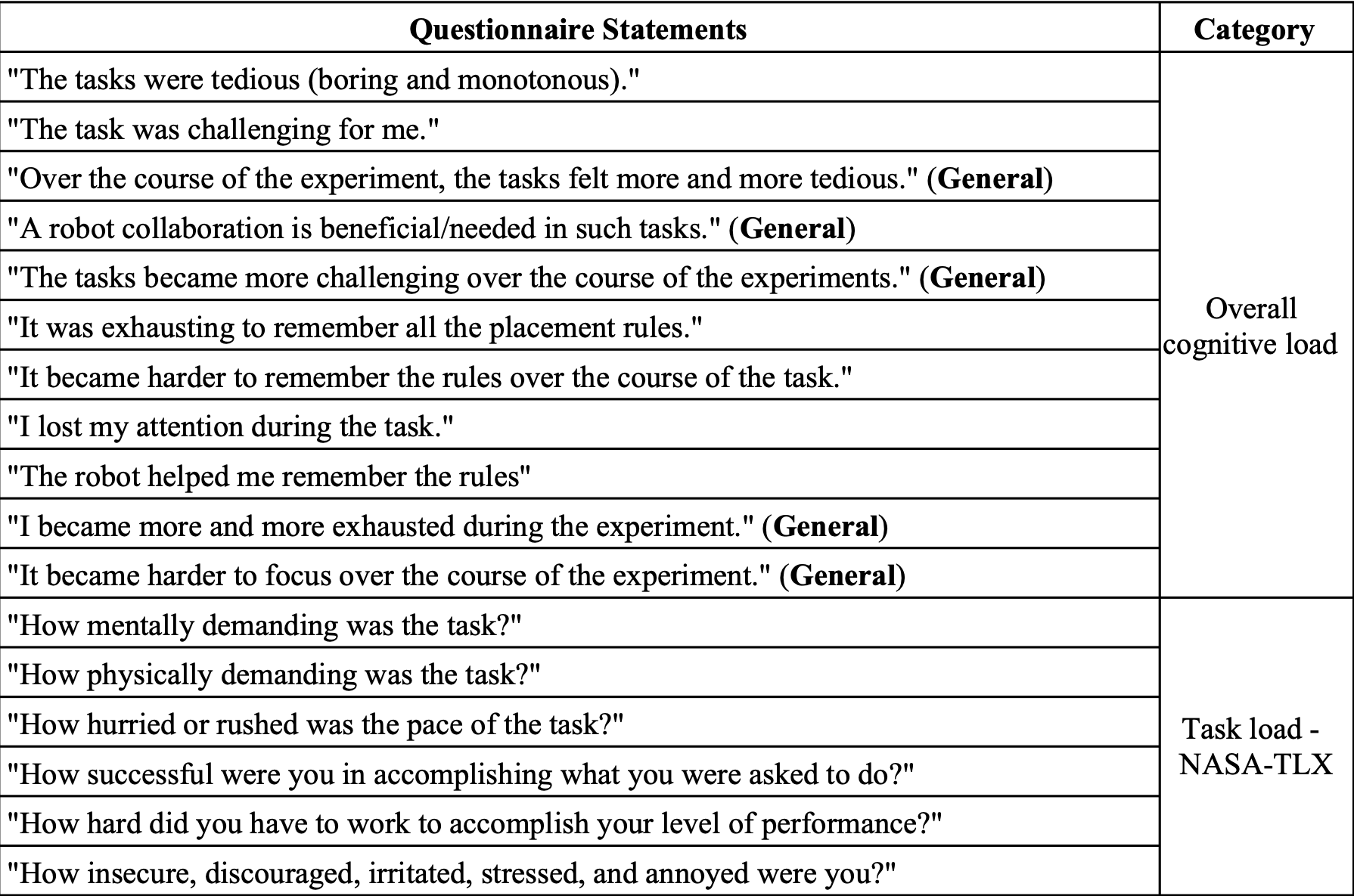}
	\label{tbl:exp0_statements}
\end{table}

Our collaboration setup aims to place a cognitive load on the humans and the degree of cognitive load may differ between the task types. As a result of the cognitive load and long working hours, unanticipated human behaviors should be invoked and observed during a task.
To achieve this, we ran the calibration experiments as part of a master's thesis study at Technische Universit\"at Berlin \citep{Kargruber2019}. Here we summarize the experiments and the results. We let 8 participants and our cobot collaborate over an extended period (approximately 1.5 hours each) executing our tasks designed for the experiments (see in Figure~\ref{fig:taskRules}). Our goal is to compare the types and choose suitable ones for the final experiments. Our criteria are: 1) The task is cognitively demanding enough to observe unanticipated human behaviors and 2) the task is easy and motivating enough to keep the human engaged in the collaboration. Hence, we measure the rewards the cobot receives, the number of cobot interferences to count how many times the cobot has taken over the task, and the number of warning gestures a participant made during a task.

The subjective measures are collected by means of questionnaire responses that the participants complete either after a task is completed or at the end of the experiment. We use a 5-step Likert scale to evaluate each of the statements in the questionnaire (in Table~\ref{tbl:exp0_statements}). Additionally, the NASA-TLX measures are used to rate the task load induced over the participants on a 20 point scale \citep{hart_development_1988}. It measures the load in 6 dimensions, namely \textit{mental demand}, \textit{physical demand}, \textit{temporal demand}, \textit{performance}, \textit{effort} and \textit{frustration}.
In Table~\ref{tbl:exp0_statements}, the statements marked with the type \textbf{General} target the experiment in general, and so they are asked only once after the whole experiment is completed. The other statements are task-specific and they are repeatedly asked after each task completion. Also, the experiments are conducted as a within-subject design to better compare the effect of each task type.

\begin{figure}[!htbp]
 \centering
 	\centering
 	\includegraphics[width=0.85\columnwidth]{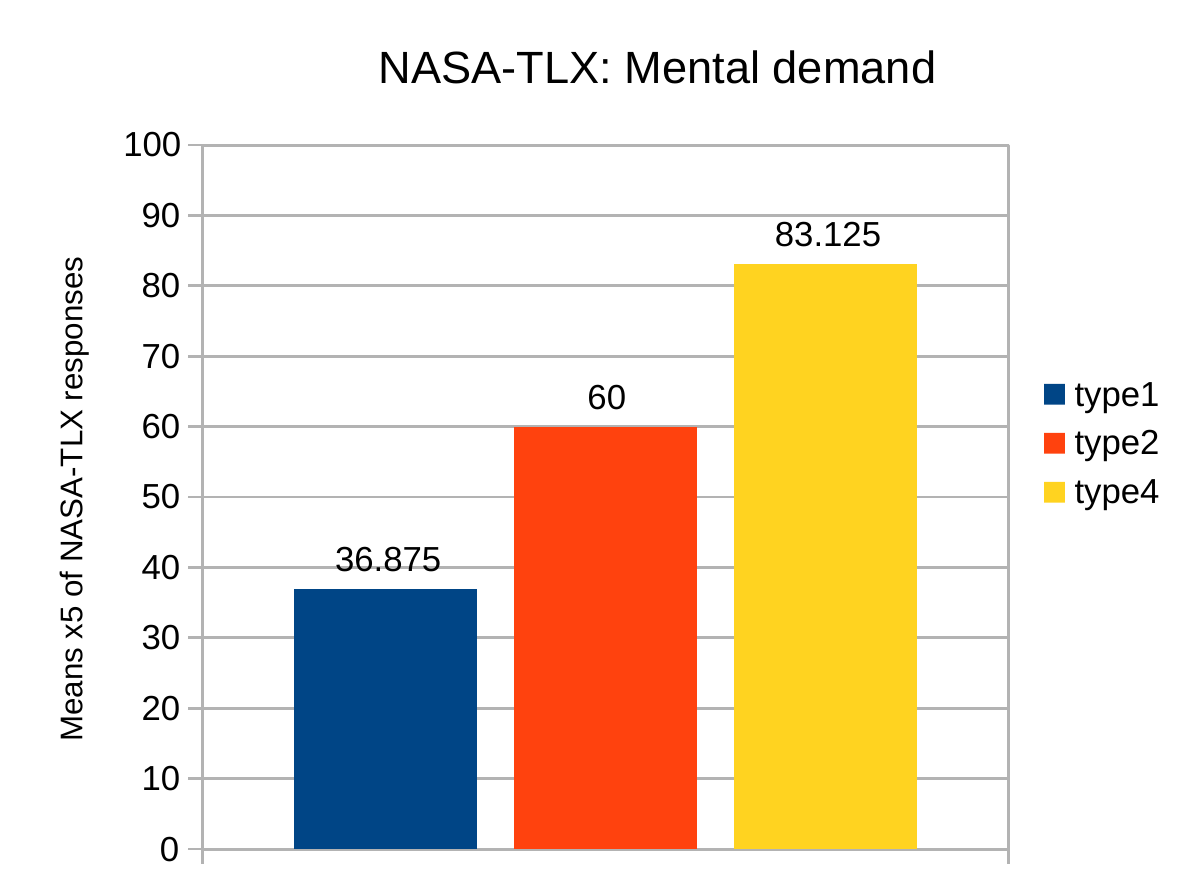}%
 	\label{fig:exp0_NASA_mental}
 %\hspace{0.01\columnwidth}
 \caption{NASA-TLX scores}
\end{figure}

During the experiments, we noticed that the task type-3 and type-5 (in Figure~\ref{fig:taskRules}c and e, respectively) are particularly difficult for humans, which has led the participants to always leave the tasks to the cobot. Therefore, we drop them from this experiment that cover shorter periods of collaboration on a same task. Each participant collaborates with the cobot on each of the 3 task types we would like to examine (type-1, type-2 and type-4 as given in Figure~\ref{fig:taskRules}) three times in a changing order. This ensures that the inspected effects are caused by the difference of the cognitive loads induced by the task types rather than by a possibly heterogeneous participant group composition or the practice-effect the participants may gain throughout the experiment.

\begin{figure}[!t]
	\begin{minipage}{\columnwidth}
		\centering
		\subfloat[``The task was challenging for me'' \label{fig:exp0_challenge}]{{\includegraphics[width=0.42\columnwidth]{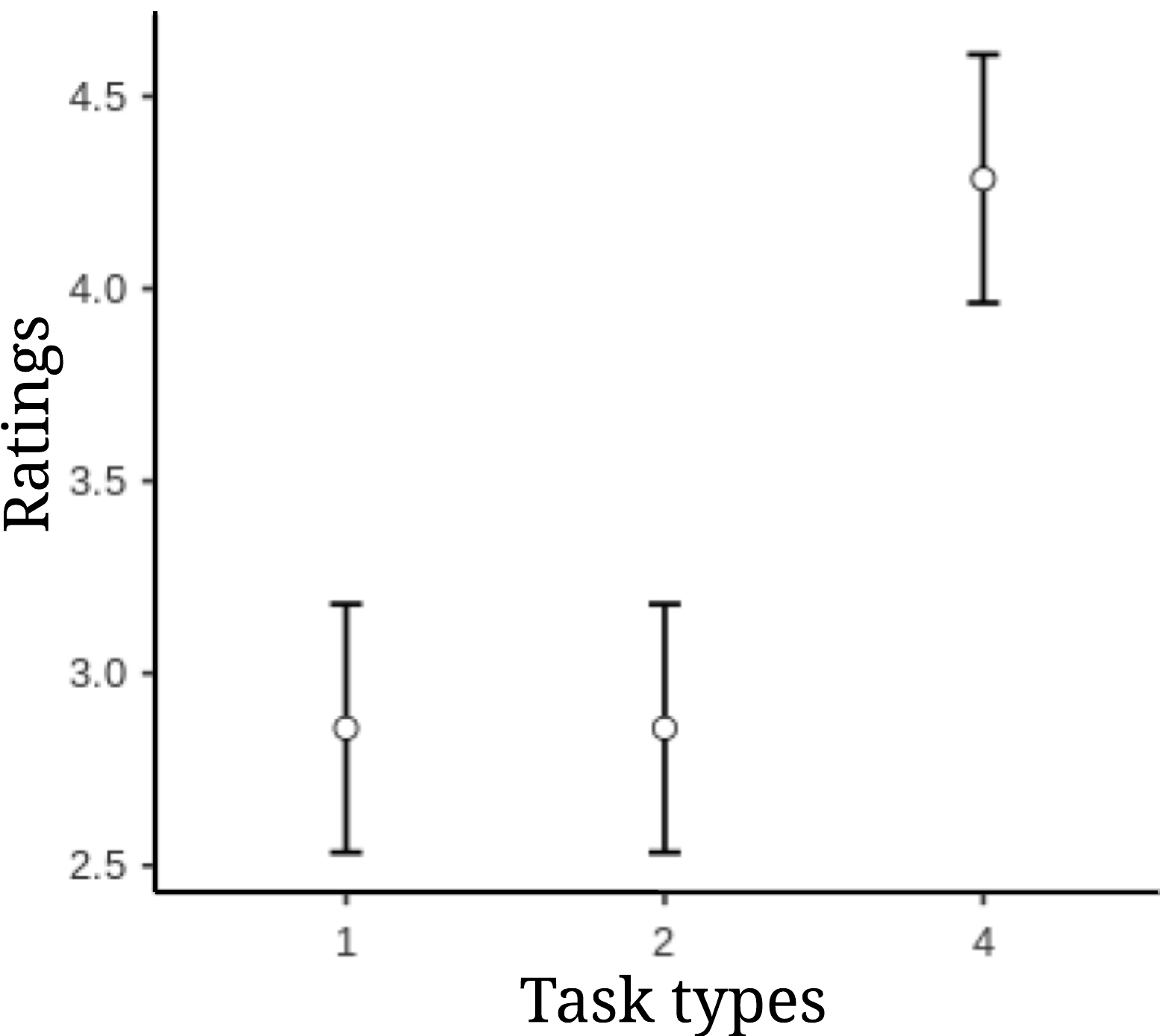} }}%
	%\end{minipage}
	%\hspace{0.01\columnwidth}
	%\begin{minipage}{0.30\columnwidth}
		\hspace{0.07\columnwidth}
		\centering
		\subfloat[``It was exhausting to remember all the rules.''\label{fig:exp0_exhaust}]{{\includegraphics[width=0.42\columnwidth]{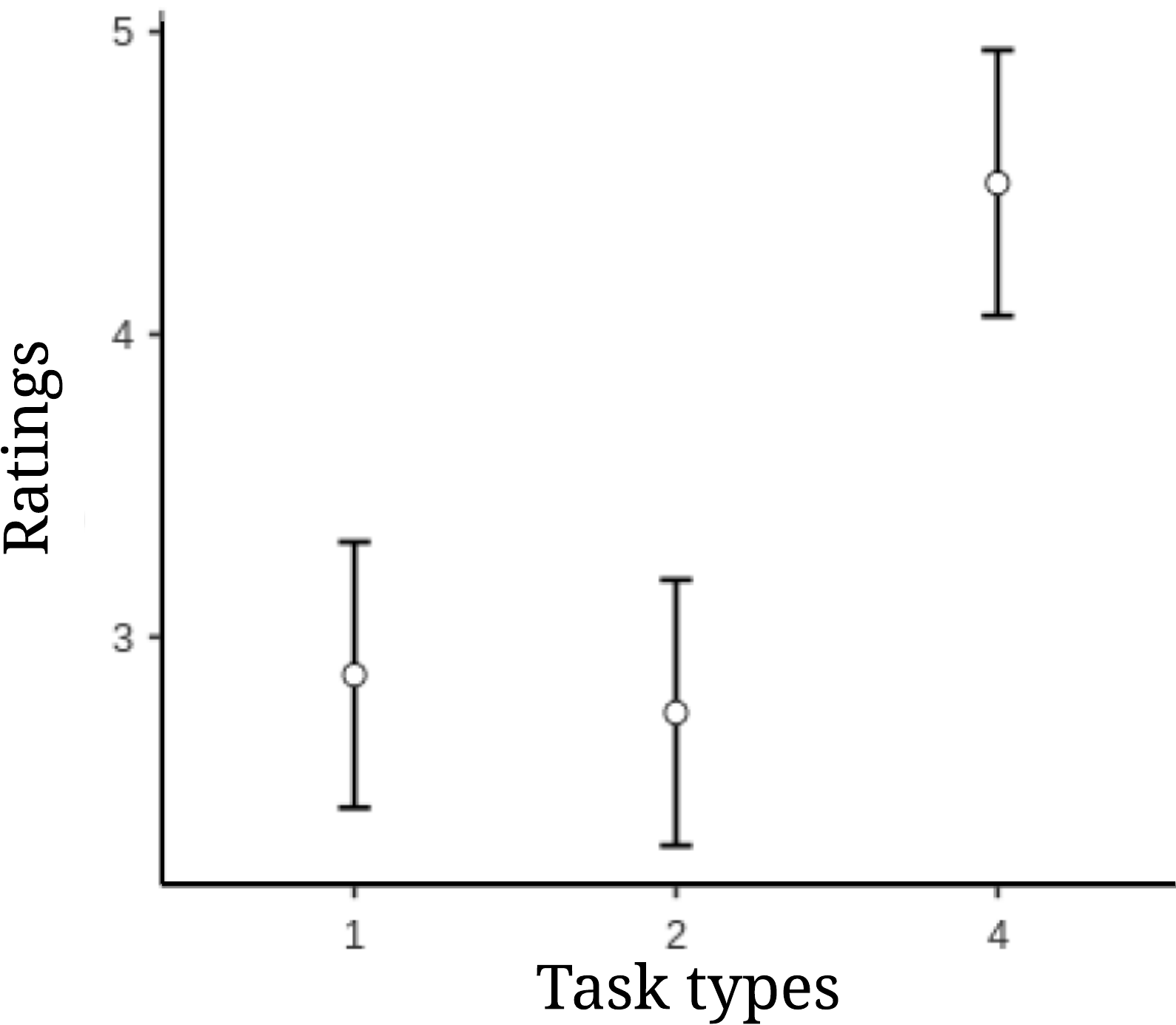} }}%
	\end{minipage}
	%\hspace{0.01\columnwidth}
	\begin{minipage}{\columnwidth}
		\centering
		\subfloat[``I lost my attention during the task.'' \label{fig:exp0_attention}]{{\includegraphics[width=0.42\columnwidth]{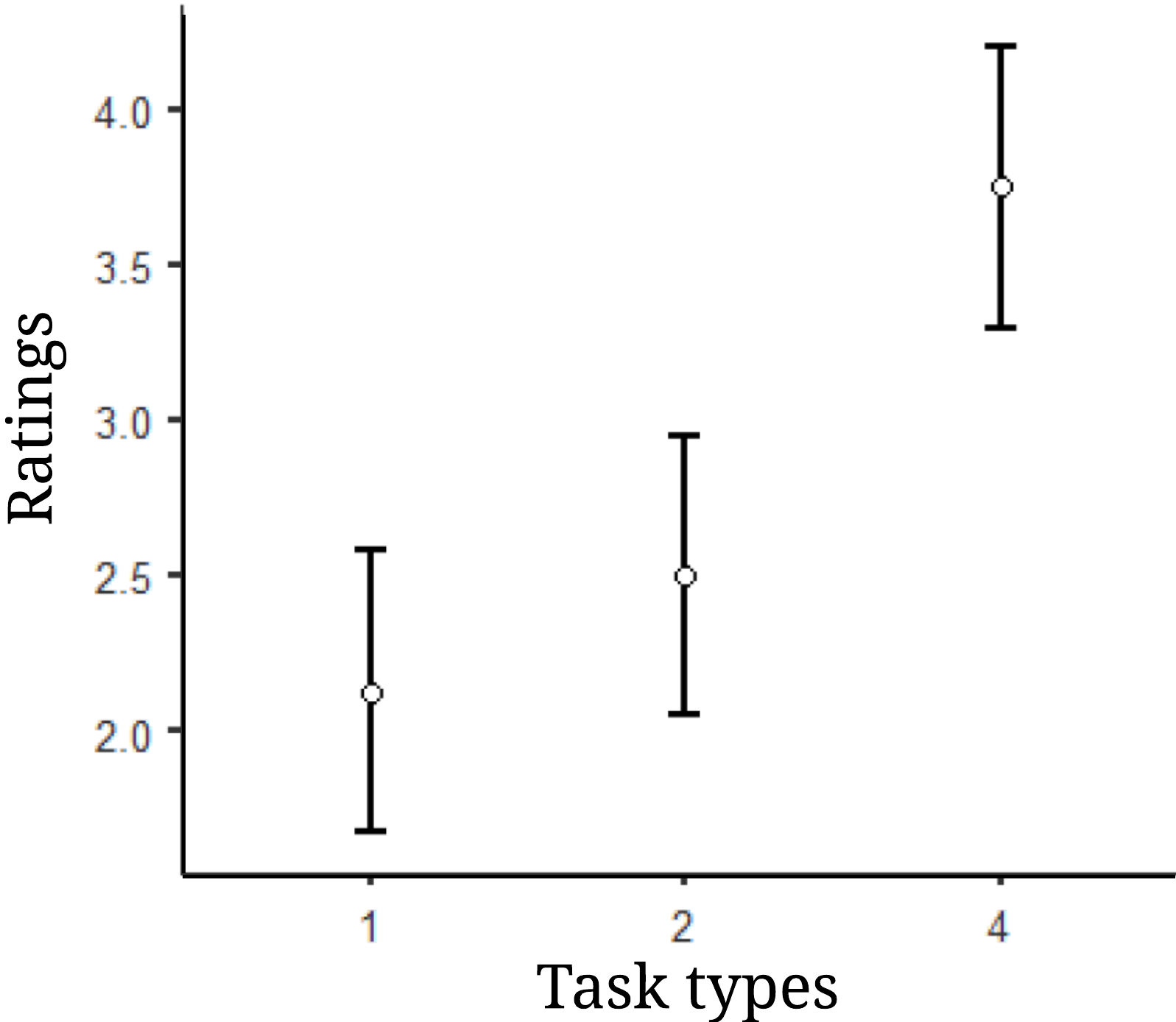} }}%
	%\end{minipage}
	%\begin{minipage}{0.30\columnwidth}
		%\includegraphics[width=\columnwidth,height=4cm]{img/meth_taskrules/RuleDisplay_type3_mixed_complex.png}%
		\hspace{0.07\columnwidth}
		\centering
		\subfloat[``It became harder to remember the rules over the course of the task.'' \label{fig:exp0_remember}]{{\includegraphics[width=0.42\columnwidth]{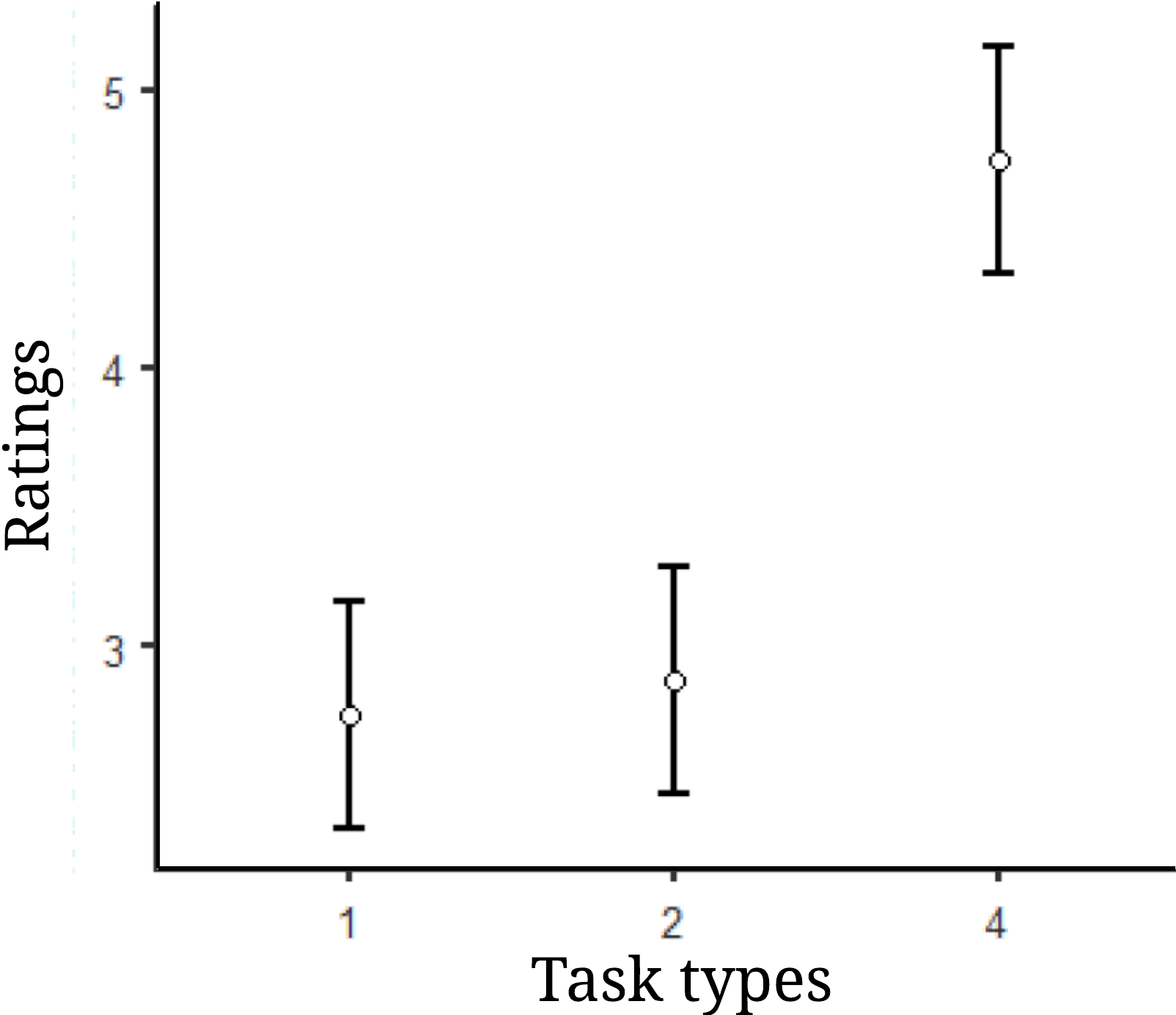} }}%
	\end{minipage}
	\caption{Mean plot of the participant ratings in 5-step Likert scale to the statements for the perceived task difficulty and the cognitive load for different task types.}
	\label{fig:exp0_difficulty}
\end{figure}

We first use a training round to introduce our industrial scenario of a human and the cobot collaborating on an assembly line. We also add that the tasks are initially assigned to the participants, leading to higher efficiency and a score if they are done by them, and the cobot's job is to assist whenever needed. We demonstrate to the participants how to grasp and place the objects and how to interact with the cobot to avoid practice-effect. We also remind them that whenever an object is detected in one of the containers, the participants should wait for the audio feedback indicating that the placement is processed. In total, 8 participants interacted with the cobot using the same A-POMDP decision model (i.e., the base model described in Section~\ref{ssec:training}). Starting with NASA-TLX rates, since in our experiments the tasks have been designed to be cognitively demanding, our focus is on the mental demand dimension that is depicted in Figure~\ref{fig:exp0_NASA_mental}. We discard the physical load in our analysis as the participants already stated that they did not feel this. The task type-4 has been reported to require the highest mental workload with a mean score of $83.125$ out of $100$, while the task type-2 had a score of $60$, and for the type-1, it is $36.875$. 

\begin{figure}[!htbp]
 	\centering
 	\begin{minipage}[c]{0.7\columnwidth}
 		\centering
 		\subfloat[Collected rewards \label{fig:exp0_rewards}]{{\includegraphics[width=\columnwidth]{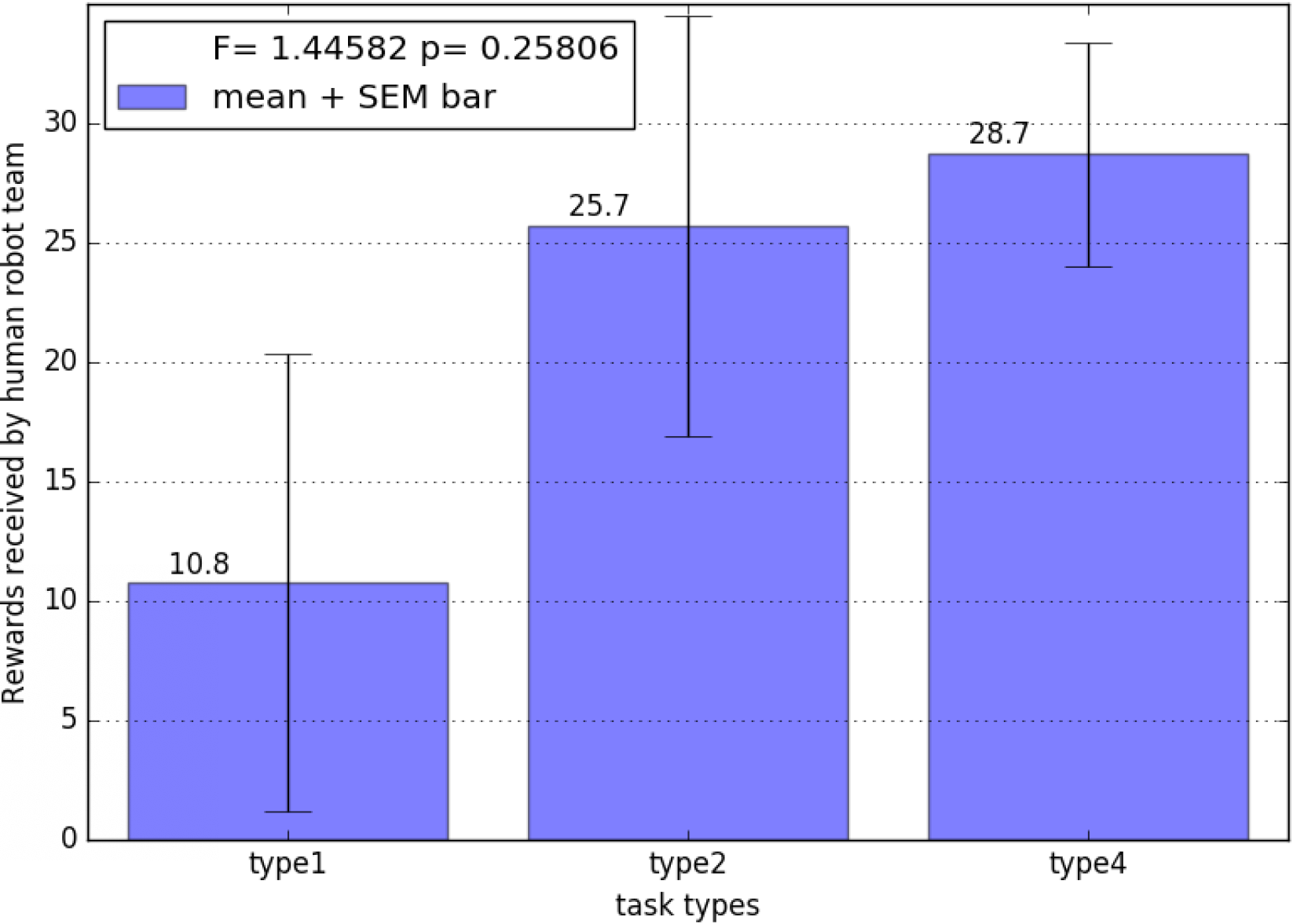} }}%
 	\end{minipage}
 	\centering
 	\begin{minipage}[c]{0.7\columnwidth}
 		%\includegraphics[width=\columnwidth,height=4cm]{img/meth_taskrules/RuleDisplay_type3_mixed_complex.png}%
 		%\hspace{0.05\columnwidth}
 		\centering
 		\subfloat[The amount of robot interference \label{fig:exp0_interference}]{{\includegraphics[width=\columnwidth]{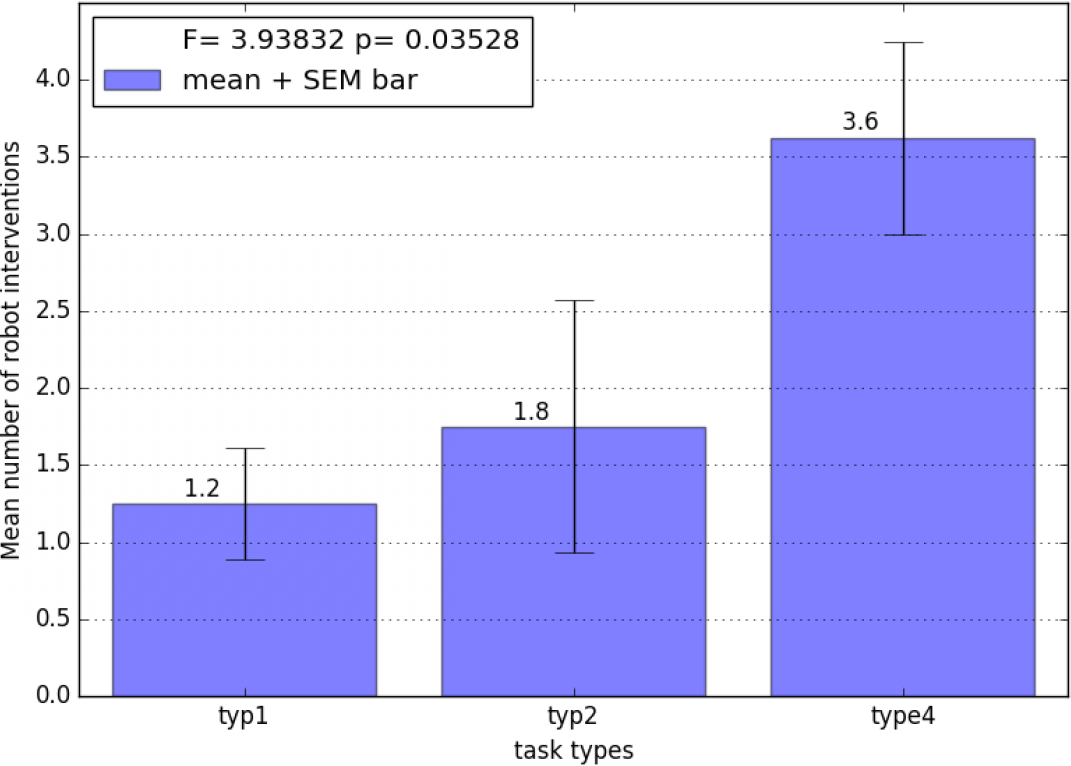} }}%
 	\end{minipage}
 	\centering
 	\begin{minipage}[c]{0.7\columnwidth}
 		\centering
 		\subfloat[The amount of human warnings \label{fig:exp0_warning}]{{\includegraphics[width=\columnwidth]{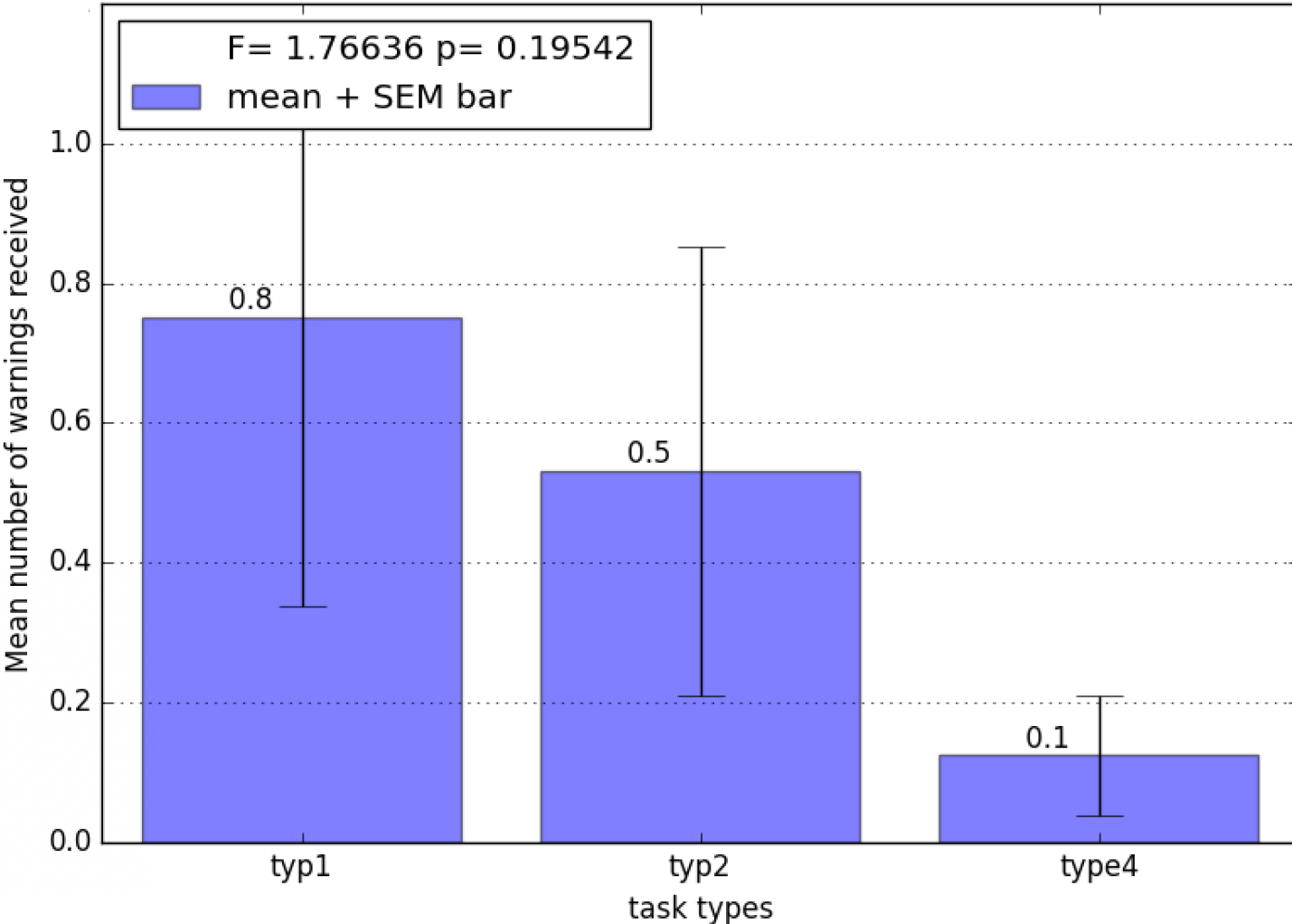} }}%
 	\end{minipage}
 	\caption{Quantitative measures averaged over the participants for each task type}
 	\label{fig:exp0_quantitative}
 \end{figure}

For the survey responses on the task difficulty, one-way ANOVA results show that the three task types significantly differ from each other concerning how challenging they were perceived by the participants ($p=0.007$, in Figure~\ref{fig:exp0_challenge}). Additionally, an $\eta^2$ of 0.42 indicates a large effect size. The post-hoc Tukey-HSD reveals that the mean difficulty for task type-4, $4.29$ out of $5$, is significantly larger than the others ($P_{tukey}=0.015$), which is in line with the NASA-TLX test. Similarly, task type-4 was significantly more exhaustive and caused more distraction, as perceived by the participants (in Figure~\ref{fig:exp0_difficulty}). As the task rules are perceived to be difficult, the participants needed to look at the task monitor (in Figure~\ref{fig:realsetup}) several times during a task to track its current state. This is also recognized by the cobot as distracted since the attention is removed from the work environment. Finally, the participants agree that they became increasingly tired during the experiment for all task types in general ($mean=3.875$). This indicates that even though a task type is kept the same, it is perceived to be more demanding over time. This perceived difficulty was particularly strong with type-4 (mean = $4.75$).
From the analysis above, we deduce that type-4 is significantly challenging and induces a cognitive load. In addition, it is well suited for the cases in which the perceived cognitive load on the participants increases over time.
 
  \begin{figure}[!htbp]
 	%\hspace{0.01\columnwidth}
 	\begin{minipage}{\columnwidth}
 		\centering
 		\subfloat[``The robot helped me to remember the rules'' \label{fig:exp0_help}]{{\includegraphics[width=0.45\columnwidth]{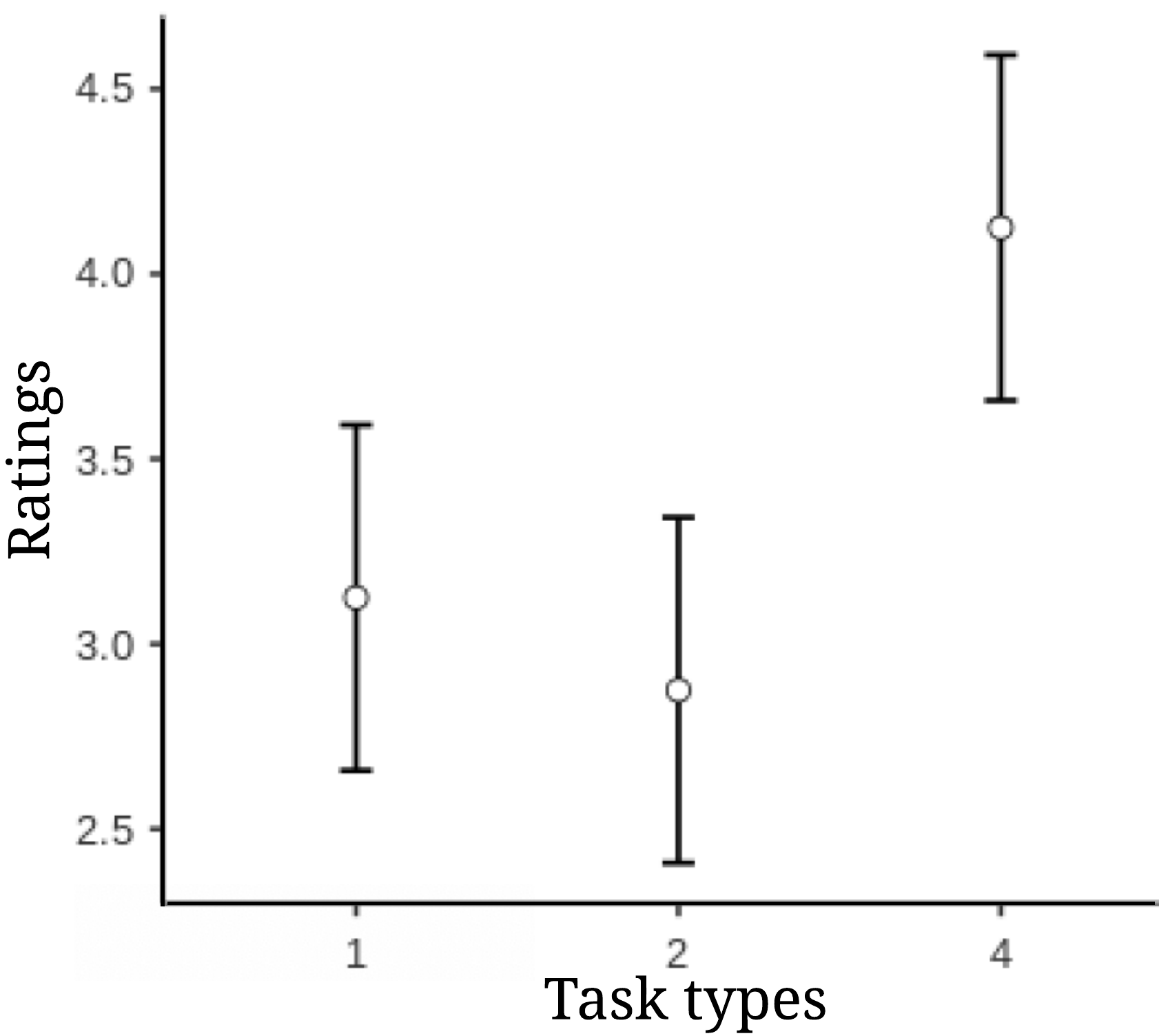} }}%
 	%\end{minipage}
 	%\begin{minipage}{0.30\columnwidth}
 		%\includegraphics[width=\columnwidth,height=4cm]{img/meth_taskrules/RuleDisplay_type3_mixed_complex.png}%
 		\hspace{0.05\columnwidth}
 		\subfloat[``I would have scored better without the robot'' \label{fig:exp0_needRobot}]{{\includegraphics[width=0.45\columnwidth]{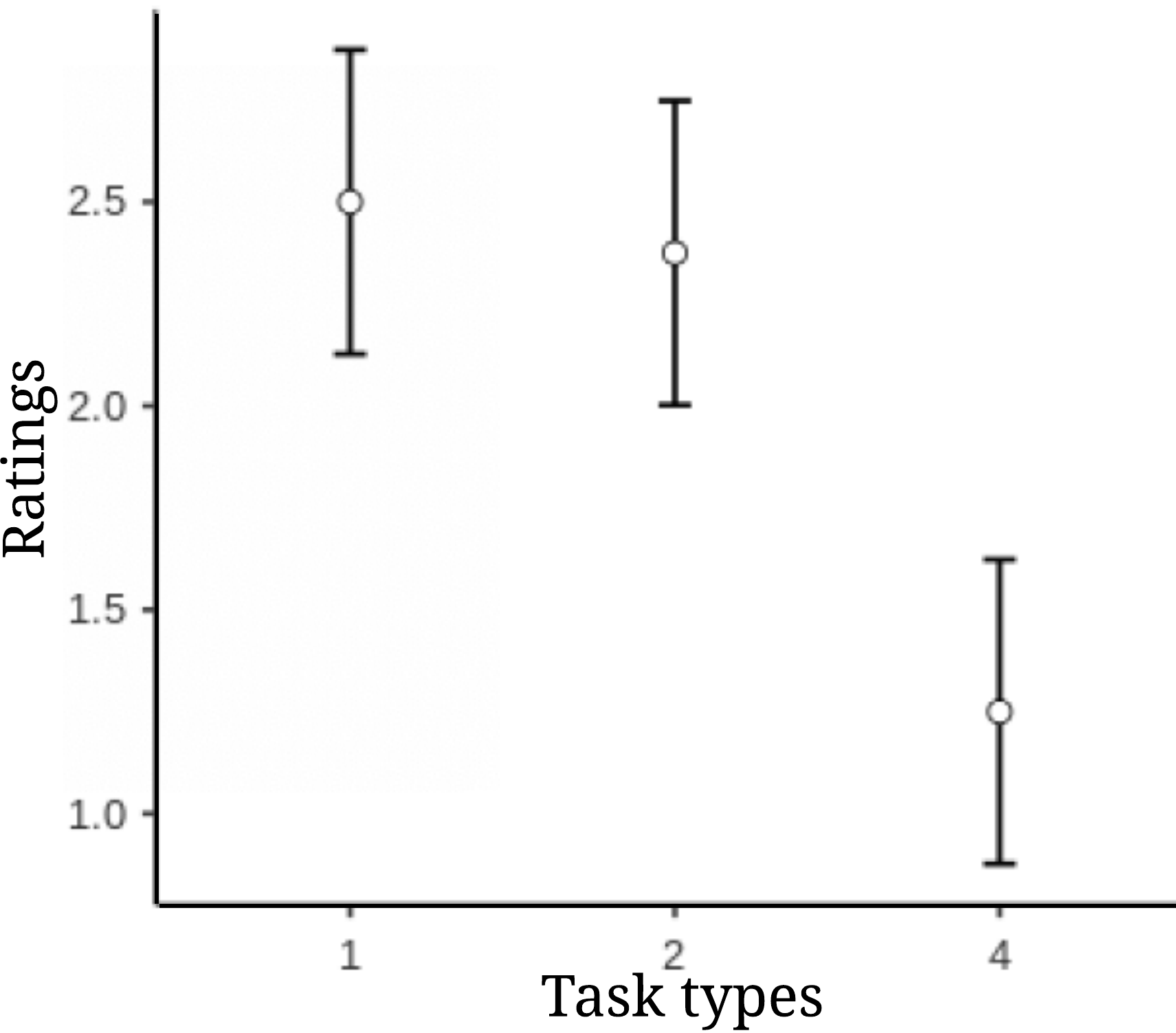} }}%
 	\end{minipage}
 	\caption{Mean plot of the participant ratings in 5-step Likert scale to the statements related to how they perceive the robot's collaboration during different task types.}
 	\label{fig:exp0_robotpercept}
 \end{figure}

Since we require collaboration, it should not be too overwhelming for the participants, or demotivating to leave the task completely to the cobot. For this purpose, we check the success rates, cobot's interference, and the participants' opinions about the cobot and the reliability of its reactions in a task. First of all,
Figure~\ref{fig:exp0_rewards} compares the received rewards ($mean_{type1}=10.8, mean_{type2}=25.7, mean_{type3}=28.7$) with
no significant difference between the rewards the cobot received after each task type ($p=0.258$). This indicates that, even though type-4 is stated to be the most difficult, the success rates are still higher. This is mainly due to significantly large contributions of the cobot in this task as shown in Figure~\ref{fig:exp0_interference}. A cobot interference describes a successful take over of the cobot, which is a successful placement of an object without receiving a warning from a participant. In other words, the participants also approve the cobot's assistance offer. Most cobot interferences occurred during task type-4 ($mean=3.6$), which is significantly higher than task type-2 ($mean=1.8$) and type-1 ($mean=1.2$) with $p=0.035$. However, the warning levels were low as given in Figure~\ref{fig:exp0_warning}. Finally, the participants also agree that the cobot, in general, helped them to remember the rules in task type-4 ($mean=4.13$), and in type-4 they would have scored significantly worse without the cobot's help than in tasks type-1, $p=0.027$ and type-2, $ p=0.045$ as shown in Figure~\ref{fig:exp0_needRobot}. As a result, it can be stated that task type-4 has led to a better coordination where the cobot could effectively support its human collaborator.

In summary, we conclude that the setup is effective in invoking unanticipated human behaviors. This is clear in task type-4 through the participant responses on not remembering the rules, their lost attention, increasing exhaustion over the course of the experiment, and from very high NASA-TLX scores. In addition, the cobot has also successfully estimated these behaviors by correctly offering its assistance.
These analyses show that the experiment setup is able to place a cognitive load on the humans, and the higher the cognitive load is the more unanticipated human behaviors, followed by more human errors. As mentioned, we choose task type-4 as the main collaboration task to evaluate our cobot's short-term adaptation capabilities.
For the evaluation of long-term adaptation, we select to work with task type-5, which has the same configuration as type-4 along with an additional Stroop effect (in Figure~\ref{fig:taskRules}). Even though it was very challenging, type-5 was very suitable for a longer collaboration as it leads to a noticeable change in human characteristics, like expertise through the learning effect (i.e., learning the Stroop effect) as discussed in Section~\ref{ssec:exp2_results} of the main paper.

\subsubsection*{Durations of Human Actions}
\label{ssec:eval_CogLoad}
The human action update frequency in Figure~\ref{fig:decisionTrigger} depends on the collaboration setup and the task. Hence, we analyze the collected observations from the calibration experiments for the average duration of human actions.
As Table \ref{tbl:durations_of_interaction_measured} indicates, it takes between 3-5 seconds for a human to grasp an object and place it in one of the containers and another 1-2 seconds to return from the container to the conveyor belt, which is the longest human action on average. We, therefore, set the same action timeout to be 3 seconds (see Figure~\ref{fig:decisionTrigger}), which is the average time needed for an action to be completed. This also makes sure that during the longest action, the observation update informs the decision-making block at least twice, stating that the human is progressing with the task. As soon as the human action changes, it is processed by triggering a response on the cobot decision-making. This timeout has been extensively tested while interacting with the environment and has shown to deliver a good balance between a timely reaction to new observations and reliable responses during long-lasting and continuous interaction.

\begin{table}[!htbp]
	\centering
	\caption{Average durations of human actions while interacting with the cobot}
  \resizebox{\columnwidth}{!}{
	\begin{tabular}{l|c}
		\textbf{Human Action} & \textbf{Duration of the action [seconds]}   \\
		\hline
		grasping and placing                            & 3-5             \\\hline
		%returning from container to conveyor            & 1-2             \\\hline
		idle times between grasp attempts               & 1-3            \\\hline
		warning the robot										& 3-4              \\\hline
		%grasping, placing and returning to conveyor     & 4-7              \\
	\end{tabular}
  }
	\label{tbl:durations_of_interaction_measured}
\end{table}
\pagenumbering{gobble}

\balance

\end{document}